\documentclass{ieeeaccess}
\usepackage{cite}
\usepackage{amsmath,amssymb,amsfonts} 
\usepackage{graphicx}
  
\usepackage{hyperref}
 \usepackage{subcaption} 
\usepackage{setspace}
\usepackage{amsmath}
\usepackage{adjustbox}
 \usepackage{xcolor}
\usepackage{textcomp}
\def\BibTeX{{\rm B\kern-.05em{\sc i\kern-.025em b}\kern-.08em
    T\kern-.1667em\lower.7ex\hbox{E}\kern-.125emX}}
\begin{document}
 \doi{https://doi.org/10.1109/ACCESS.2024.3472859}

\title{Recursive deep learning framework for forecasting the decadal world economic outlook}

\author{\uppercase{Tianyi Wang}\authorrefmark{1},
\uppercase{Rodney Beard \authorrefmark{3}, John Hawkins \authorrefmark{3}, and Rohitash Chandra}.\authorrefmark{1} 
\IEEEmembership{SM, IEEE}}
\address[1]{Transitional Artificial Intelligence Research Group, School of Mathematics and Statistics, UNSW Sydney, Australia } 
\address[3]{Centre for Artificial Intelligence and Innovation, Pingala Institute, Sydney, Australia }


\corresp{Corresponding author: R. Chandra (e-mail: rohitash.chandra@unsw.edu.au)}

\begin{abstract}

The gross domestic product (GDP) is the most widely used indicator in macroeconomics and the main tool for measuring a country's economic output. Due to the diversity and complexity of the world economy, a wide range of models have been used, but there are challenges in making decadal GDP forecasts given unexpected changes such as emergence of catastrophic world events including pandemics and wars. Deep learning models are well suited for modelling  temporal sequences and  time series forecasting.  
In this paper, we develop a deep learning framework to forecast the GDP growth rate of the world economy over a decade. We use the \textit{Penn World Table} as the data source featuring  13 countries prior to the COVID-19 pandemic, such as Australia, China, India, and the United States.  We present a  recursive deep learning framework to predict the GDP growth rate in the next ten years. We test prominent deep learning models and compare their results with traditional econometric  models for selected developed and developing countries. Our decadal forecasts reveal that   that most of the developed countries would experience economic growth slowdown, stagnation and even recession within five years (2020-2024). Furthermore, our model forecasts show that only China, France, and India would experience stable  GDP growth. 

\end{abstract}


\titlepgskip=-15pt

\maketitle

\section{Introduction}

Economists and policymakers rely on a wide range of macroeconomic indicators to guide decisions and social policies that impact the economies of the world \cite{jansen2018combining}. Among macroeconomic indicators, the gross domestic product (GDP) is the most widely used. The GDP is a measure of the market value of all the final goods and services produced in a specific time period in a  country \cite{callen2012gross}. The GDP is seen as a powerful statistical indicator of national development and progress \cite{lepenies2016power} and is often used as  a means to measure the economic health of a country. The GDP is also correlated with the employment rate \cite{akkemik2007response,vladuvsic2020macroeconomic} which can also give an indication of trade and investment opportunities \cite{bekaert2007global}. \textit{Expectations} are a key element of macroeconomic theories that refers to forecasts and viewpoints of experts about economic indicators  such as sales, income, taxes,  and future prices \cite{evans2012learning,casey2020macroeconomic}. Macroeconomic forecasts can also be expected to have an influence on how individuals anticipate the economy will evolve. It is precise because GDP is closely related to other macroeconomic indicators such as stock prices and the unemployment rate \cite{giannone2008nowcasting} that it is regarded as the ``core" metric to measure the development and growth of a country. Accurate forecasting of GDP  is critical for economic policy in order  to minimise errors in decision-making \cite{lyhagen2015beating}. Furthermore, GDP forecasts routinely tend to fail to capture drastic economic  trends \cite{Ewing2005}, which are likely to be more critical for effective policy responses. International economic and financial organizations typically publish forecast reports   for governments and corporations. The International Monetary Fund (IMF)  publishes quarterly reports about their views on the world's economic development  \cite{imf2022report} that include views on GDP that can influence the domestic and foreign policy of a country \cite{vreeland2003imf,garboden2020sources}.

GDP forecasting often requires the use of exogenous variables (features)    such as oil and stock prices that can be collected more frequently than quarterly GDP data \cite{nicholson2017varx}. GDP forecasting has necessitated the use of domain specific modelling techniques that merge data across timescales, including \textit{bridge equations} or \textit{dynamic factor models} \cite{marcellino2010factor,Feldkircher2015,Garnitz2019,Ollivaud2016}.
These models are generally linear equations composed over a series of lagged and differences of exogenous and autoregressive variables \cite{montgomery2021introduction}. They fit within the broader family of traditional time series models defined by the autoregressive integrated moving average (ARIMA) \cite{box2015time}  and vector autoregressive (VAR) methodologies \cite{sims1980macroeconomics}. The weakness of these approaches is the restriction to linear relationships, which has been a necessary compromise in the absence of large data sets. However, these approaches have consistently failed to identify crucial changes in the economy \cite{Ewing2005,Ollivaud2016}. It remains an open research question to determine the best forecasting methodology for a given economic purpose. Machine learning models have the ability to better capture nonlinearity and adapt to noisy data, and hence have drawn the attention of econometricians \cite{athey2018impact,athey2019machine}.

 Deep learning provides a set of machine learning models   that have demonstrated to overcome  some of the limitations of statistical and econometric models \cite{yoon2021forecasting, Deng2018ArtificialII}. These techniques have achieved widespread adoption in areas such as image recognition,  recommender systems, time series forecasting, and autonomous driving   \cite{lecun2015deep}. Recurrent Neural Networks (RNNs) are deep learning models designed to process arbitrarily long sequences of data \cite{Elman1990} which has made them prominent for natural  language processing (NLP)\cite{chandra2021covid,sun2017review,chandra2021biden}, bioinformatics \cite{tcbb:Hawkins+Boden:2005}, and time series forecasting \cite{chandra2021evaluation}. The Long Short-Term Memory (LSTM) network is an advanced  RNN designed to identify long-term dependencies in sequences \cite{hochreiter1997long}. Since the development of LSTM networks, further advancements have been made.  The Bi-directional LSTM (BD-LSTM) was developed for exploiting information across an entire sequence in order to improve NLP models \cite{graves2005bidirectional}; however, it also can be used for time series forecasting by processing the preceding sequence in both directions \cite{said2021predicting,chandra2021evaluation}. The Encoder-Decoder LSTM (ED-LSTM) is another variant  that learns to encode a variable-length sequence into a fixed-length vector representation and to decode a given fixed-length vector \cite{cho2014learning}, widely used in pattern recognition   \cite{park2018sequence} and time series forecasting \cite{chandra2021evaluation}. Convolutional neural networks (CNNs) \cite{alzubaidi2021review} are deep learning models developed specifically for image processing which have shown very promising results in time series prediction \cite{chandra2021evaluation}, such as stock price prediction \cite{chen2016financial}.  We note that although they have  potential, CNNs have not been extensively applied in macroeconomic forecasting \cite{maehashi2020macroeconomic}.

Machine learning models have been very promising in forecasting competitions, typically through the involvement of sophisticated ensemble learning methods \cite{makridakis2018statistical,gilliland2020value}.  In addition, econometricians have identified important roles for machine learning models in GDP forecasting \cite{bianchi2022belief,goulet2022machine}, with
 deep learning models for forecasting  the GDP growth rate using a multivariate approach \cite{sa2020prediction,zhang2022china,sokolov2016economic}.  Zhang et al. \cite{zhang2022china} combined LSTM and hidden Markov models (HMMs) to use consumer price index (CPI)\cite{bryan1993consumer} to forecast China's GDP before the COVID-19 pandemic. Based on  import and export data, Sokolov-Mladenovic et al. \cite{sokolov2016economic} forecasted the GDP of European Union countries using simple neural networks and extreme learning machine models. Mourougane \cite{mourougane2006forecasting} made a monthly forecast of Canada's GDP based on hard indicators related to quantifiable indicators (industrial production, unemployment, etc.) and soft indicators that are less tangible community characteristics and values \cite{petrova2015using} (confidence index, exchange rate, etc.). Longo et al. \cite{longo2022neural} used the ensemble method based on the LSTM to forecast the GDP of the United States and provided insights about the contribution of different features (economic indicators) during COVID-19. The economic indicators   included 141 variables taken from the Federal Reserve Economic Data (FRED) database. 

 Long-term (decadal prediction) is important for policymaking, since many countries plan their economy and investments for decades ahead of time. An example is China, which had planned and invested in Africa since the 1990s \cite{kidane2013china}. The International Monetary Fund  and the World Bank have a decadal ahead forecast which is taken as an input in policymaking, considering factors such as climate change, population growth/decline and energy supply \cite{vreeland2003imf,paloni2012imf}.  
  Time series forecasting  employs two major strategies, which include \textit{direct} and \textit{recursive} strategies \cite{clements1996multi,klein1971essay,marcellino2006comparison}; although direct strategy can be used for multi-step time series prediction,  the recursive strategy is more useful when the prediction horizon spans longer time. Hence, it would be more appropriate to use a mix of direct and recursive strategies in decadal forecasting. Moreover, there is a gap in decadal forecasting methods using deep learning with multivariate analysis of economic indicators. We would need to forecast the economic indicators using a recursive deep learning strategy, in order to forecast the decadal GDP growth rate.  We find that most of the studies in the literature focused on  developed countries for GDP forecasting, and only a few studies    forecast developing countries. This could be because developed countries have better access and complete data about economic indicators, which are more helpful for forecasting. It is a challenge to develop deep learning models given limited and missing data about economic indicators. There is a growing interest in utilizing  appropriate deep learning models for GDP forecasting when compared to traditional econometric models. 
 
 In this paper, we  use  deep learning models  that include LSTM-based models and CNNs for decadal GDP growth rate forecasting of major world economies. We focus on large economies around the world, including developing and developed countries. We present a recursive  deep learning  framework  where the direct strategy is used for model development and the recursive strategy is used for the decadal forest. We use  data from the past few decades to forecast the future decade.  We also investigate the data partitioning strategy that is best for model training and development. We further compare the performance of deep learning  models 
with traditional time series forecasting models (ARIMA and VAR) for different countries. Our data  includes periods of smooth development, rapid development, and periods of financial crisis in the training set, in order to better prepare the test set and forecast data. We first use a direct strategy to evaluate the respective deep learning models and then use the best model for the recursive strategy, where we estimate the economic indicators   to forecast the decadal GDP growth rate. Our data features GDP along with economic indicators prior to 2019 that excludes the COVID-19 pandemic, and we forecast the  decadal world economic outlook for 2020 - 2030.

The rest of the paper is organised as follows. Section 2  provides a review of the literature and Section 3 presents  the methodology where we evaluate deep learning models and present a recursive deep learning framework for the world economic outlook. In Section 4, we present  the results and  Section 5 summarizes the results and discusses promising future developments. Section 6 concludes the paper by highlighting major results.

\section{Related Work}

\subsection{Econometric models for time series forecasting}

The ARIMA  model is a  prominent statistical model  for time series analysis and   forecasting macroeconomic trends. A number of studies in the literature have utilized it for economic forecasting, such as the prediction of Singapore's quarterly GDP based on monthly external trade\cite{abeysinghe1998forecasting} and the  Irish Consumer Price Index (CPI)   \cite{meyler1998forecasting}, and the decadal Egyptian GDP \cite{abonazel2019forecasting}.   Sims \cite{sims1980macroeconomics} introduced  the VAR model   in 1980 for macroeconomic modelling and forecasting in order to deal with endogeneity issues that plagued macroeconomic forecasting models. Freeman and Williams \cite{freeman1989vector} used a VAR model to analyze indicator variables that relate to policy and the economy,  and reported that the VAR model was better at capturing policy endogeneity when compared  to structural equation models \cite{hox1998introduction}. A decade later, Robertson et al. \cite{robertson1999vector}  used the VAR model to predict United States GDP using six economic indicators and found  that imposing imprecise prior constraints on VAR led to more accurate predictions.  Salisu et al. \cite{salisu2021effect} analyzed how the oil uncertainty stock affected the GDP of 33 countries  using a global VAR. Iorio et al. \cite{iorio2022comparison} compared France and Germany's unemployment and GDP growth rate using a VAR model. ARIMA and VAR models remain widely applied in a range of econometrics and finance applications \cite{hryhorkiv2020forecasting,d2018macroeconomic}.
 
 \textcolor{black}{Other approaches to time series prediction include machine learning models such as support vector regression (SVR).} M{\"u}ller et al.\cite{muller1997predicting} used  SVR  for time series forecasting using benchmark   problems. Lau et al.\cite{lau2008local} implemented SVR  for Sunspot time series forecasting with better results than the radial basis function network in relatively long-term prediction. In the financial time series forecasting domain, Lu et al.\cite{lu2009financial} combined independent component analysis   with the SVR model and improved the accuracy further. Although a  wide range of applications of time series forecasting using SVR has been presented \cite{sapankevych2009time,amiri2009svm}, our study focus on deep learning models. 
 
Researchers in the past have combined  statistical  models with  deep learning models for improving their performance. Tseng et al. \cite{tseng2002combining} modelled the production value of Taiwan's machinery industry using seasonal time series data of soft drink factories by combining seasonal ARIMA and  multilayer perception \cite{li2012brief} that achieved better results than standalone models. Choi  \cite{choi2018stock} presented an ARIMA-LSTM  model for stock price correlation coefficient prediction, where the ARIMA model processed linear correlation in the data and passed the residual to the LSTM model. Ouhame et al.\cite{ouhame2019multivariate} proposed a VAR-LSTM hybrid model where  the assumed linearity and the  LSTM model catered for nonlinear data.

\subsection{Deep learning models for time series forecasting}

Deep learning methods learn from raw data to understand structures and patterns for prediction  \cite{goodfellow2016deep,patterson2017deep}. Deep learning has been applied across multiple domains such as speech recognition, visual object recognition, genomics \cite{jelodar2020deep,cadieu2014deep,eraslan2019deep}, and time series forecasting \cite{chandra2021evaluation}. 

Among deep learning models, the LSTM model is most prominent for time series forecasting and has a wide range of applications, including  macroeconomics, stock price, and   COVID-19 infection.  Karevan et al. \cite{karevan2020transductive} forecasted  weather for cities in Belgium and the Netherlands using a transductive LSTM  where  the cost function was altered to differently weigh the data for improving the accuracy of the prediction. Smalter et al. \cite{smalter2017macroeconomic} used an LSTM model to forecast macroeconomic indicators and extended the model into the ED-LSTM architecture. They used these models to predict the unemployment rate and compared them with a VAR model, where the LSTM and  ED-LSTM were better than the traditional models. Siami-Namin et al. \cite{siami2019performance} used stock market data from 1985 to 2018 to compare ARIMA, LSTM and BD-LSTM models. Chandra et al. \cite{chandra2021evaluation} presented a deep learning framework for predicting novel COVID-19 cases in India, where the univariate  ED-LSTM  model provided  the best performance and provided  forecasts two months in advance.

 The CNN model  can also be used for time series-related problems. Selvin et al. \cite{selvin2017stock} employed a CNN model to predict the stock prices of companies such as Infosys, Tata,  and Cipla and reported that CNN  was better than the LSTM model. Piao et al. \cite{piao2019housing} predicted housing price using a CNN model with  variable selection. Livieris et al. \cite{livieris2020cnn} used  CNN-LSTM hybrid model to predict the prices and price trends for gold markets. 

\section{Methodology}

\subsection{Data}

 The Group of Seven (G7) is an inter-governmental political forum of seven developed countries consisting of  Canada, France, Germany, Italy, Japan, the United Kingdom, and the United States. These countries exert significant influence over global geo-political affairs that impact the world economy \cite{vu2020sources,jorgenson2013emergence}. On the other hand, BRICS refers to the five major emerging economies and developing countries, including Brazil, Russia, India, China and South Africa.  
 These developing countries occupy an indispensable position in the future development of the world \cite{o2001building}. 
 
 We select thirteen countries by combining Australia with the  G7 and BRICS. This set includes developed and developing countries and covers a significant proportion of the global economy. We use the Penn World Table (PWT) version 10.0 \footnote{\url{https://www.rug.nl/ggdc/productivity/pwt/?lang=en}{PWT 10.0}} dataset \cite{feenstra2015next} to extract a time series dataset of economic indicators for each of these countries. The PWT 10.0 database contains information on relative levels of income, output, input and productivity. It covers 183 countries between 1950 and 2019. We extracted data from 1980 to 2019 in order to maximize the number of countries with complete coverage. The exception is Russia, for which there was data only for the period from 1991-2019.

 We choose a subset of the available variables from PWT (economic indicators) to use in our models for predicting the GDP growth rate. This includes GDP, net population, consumer price index
(CPI), employment rate, the share of labor compensation in GDP at current national prices, exchange rate,  GDP per-capita at current purchasing power parity  (CGDPo), and price levels that refer to  the perspectives of macroeconomics (production, expenditure, and trade), finance, and human resources \footnote{url{https://data-planet.libguides.com/PennWorldTables}}. Our model will predict the GDP rate using a predefined  window of the previous GDP rates along with economic indicators.  We also    supplement the PWT data by a price index that can show whether inflation is present, in order to help us better predict economic development and economic crises \cite{zhang2022china,sokolov2016economic,mourougane2006forecasting,longo2022neural}. We transform data into a year-on-year ratio in order to predict the GDP growth rate.

The year-on-year ratio $R$ formula is:
\begin{equation}
R =\frac{Data_n - Data_\text{n-1}}{Data_\text{n-1}}\\
\end{equation}
where $n$ represents the year.

We transform all of the variables(features) in the PWT data using a year-on-year ratio. We apply a min-max scaling operation to remove the inter-variable scale differences and ensure all values reside between zero and one.  

 \subsubsection{Data Processing}

In the deep learning models, in order to develop multivariate and multistep-ahead prediction models, we need to reconstruct the original time series data with input-output pairs for  training the model. This is done typically using a sliding-window approach, and hence also known as windowing. Suppose our original time series data is given by \[[x_1, x_2,\ldots, x_N; y_1, y_2,\ldots, y_N]\] 
where $N$ is the length of data. Next, we introduce the definition of time window \(T\). The reconstructed data \(D\) consists of multiple data windows \[[T_1, T_2,\ldots, T_{N-m}]\]. 

Each data window can be regarded as a sample (instance) of the data used for model training. We need to determine the input step size $m$ and output step size $n$ first. In a time window, we have $m$-step input \(X\) to represent the $m$-th exogenous series in \(T\) and n-step output \(Y\), as the n-th target series of \(T\). 
Therefore, the first time window is as follows.

\begin{equation}
\begin{aligned}
& X_1 = [x_1,x_2,\ldots,x_m]\\
& Y_1 = [y_{m+1},y_{m+2},\ldots,y_{m+n}]\\ 
\end{aligned}
\end{equation}

Next, using the same method, we can reconstruct the remaining data until \(x_N\) is included in the latest time window, so we can write its universal formula below.

\begin{equation}
\begin{aligned}
& X_i = [x_i,x_{i+1},\ldots,x_{i+m}]\\
& Y_i = [y_{i+m+1},y_{i+m+2},\ldots,y_{i+m+n}]\\ 
\end{aligned}
\end{equation}

\subsection{Econometric models}

\subsubsection{ARIMA model}

The ARIMA model by  Box and Jenkins  \cite{box2015time} has been prominent  for nearly half a century in time series forecasting. It is a combination of three components, autoregressive model \((AR)\), integrated average \((I)\) and moving average model \((MA)\). ARIMA models have three parameters to represent the three parts in this model respectively, written as \(ARIMA(p,d,q)\). \(AR(p)\) represents the past value   used for prediction,  and \(p\) can be determined by the partial autocorrelation function. \(I(d)\) refers to the times of differences to ensure the data is stable. We use the  Augmented Dickey-Fuller (ADF) test  to help us to find \(d\). \(MA(q)\) expresses the current data and errors in the past $q$ values, which  can be   found by analyzing the autocorrelation function. We define the ARIMA model  by 
\begin{equation}
\begin{aligned}
w_t&=\Delta^d y_t\\
w_t&=\phi_1 w_{t-1}+\ldots+\phi_p w_{t-p}\\
   &\quad+\varepsilon_t \\
   &\quad-\theta_1 \varepsilon_{t-1}-\ldots-\theta_q \varepsilon_{t-q}
\end{aligned}
\end{equation}
where $\phi$ and $\theta$ are the coefficients of values $w$ and errors $\varepsilon$, respectively.

\subsubsection{VAR model}

The VAR model   has been traditionally used to explain the relationship with exogenous variables (features) in the data \cite{sims1980macroeconomics}. The model combines the lagged endogenous variables with the exogenous variables to explain the endogenous variable (target) which extends the regression model to multivariate  time series.

\begin{equation}
\begin{aligned}
Y_t=BX_t+A_1 Y_{t-1}+\ldots+A_p Y_{t-p}+u_t 
\end{aligned}
\end{equation}
where $X$ is the exogenous variable, $Y$ is the endogenous variable, $u$ is the error term, $A$ and $B$ are the coefficients. 

\begin{equation}
AIC=-2 \ln (L)+2k
\end{equation}


The Akaike Information Criterion (AIC) \cite{bozdogan1987model} is typically used in the VAR model for determining model hyperparameters, such as  determining the best lag, where $L$ is the maximum likelihood, and $k$ is the number of parameters.

\subsection{Deep learning  models}
\subsubsection{LSTM model}

\begin{figure}[htbp!]
 \centering
 \includegraphics[width=0.3\textwidth]{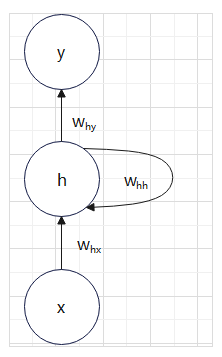}
 \caption{RNN structure showing information from input $x$ to output $y$ via the state and hidden layers $h$.}
 \label{fig:RNNframe}
\end{figure}

RNNs feature a recurrent (context) layer $h$ that acts as a memory for preserving state information for output $y$ during the processing of a sequence of data, $x = (x_1,...,x_t)$. The hidden layer $h_t$  stores the previous hidden result to update the weight $w_{h h}$ to process the $x_t$ in next time step and output $y_t$, as  shown in Figure \ref{fig:RNNframe} and given by 
\begin{equation}
 \begin{aligned}
 &\mathrm{y}_t=f_1\left(h_t w_{h y}+b_y\right) \\
 &\mathrm{h}_t=f_2 \left(x_t w_{x h}+h_{t-1} \mathrm{w}_{h h}+\mathrm{b_h}\right)
 \end{aligned}
\end{equation}
where $w$ is the weight in different layers, $f_1$ and $f_2$ are the activation function, and $b$ is the bias. 

\begin{figure}[htbp!]
    \centering
    \includegraphics[width=0.45\textwidth]{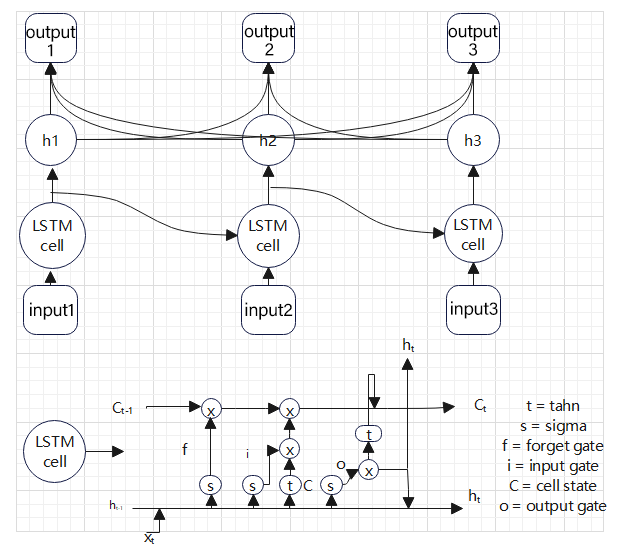}
    \caption{LSTM network with memory cell for handling long-term dependencies.}
    \label{fig:LSTM_frame}
  \end{figure}

Conventional RNNs faced limitations in training due to vanishing  gradients in long sequences; hence,  Hochreiter and Schmidhuber  \cite{hochreiter1997long} designed  an LSTM network which introduced memory cells and gates that enable them to have better performance in processing the  long-term dependencies in sequences.  Figure \ref{fig:LSTM_frame}) presents the LSTM network where the context layer is leveraged  by memory cells. 

\begin{figure}[htbp!]
    \centering
    \includegraphics[width=0.45\textwidth]{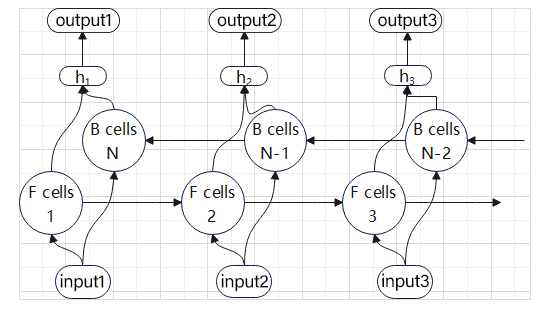}
    \caption{BD-LSTM model showing the  forward direction LSTM cells \(F\ cells\)  and backward direction LSTM cells \(B\ cells\).}
    \label{fig:BD-LSTM_frame}
  \end{figure}

  
 BD-LSTM is a variant with two types of LSTM cells that process the input sequence data in two directions, from the start to the end, and from the end backwards. This bi-directionality allows the model to build up two separate hidden state representations of the input sequence as shown in Figure \ref{fig:BD-LSTM_frame}.
 
\begin{figure}[h]
    \centering
    \includegraphics[width=0.45\textwidth]{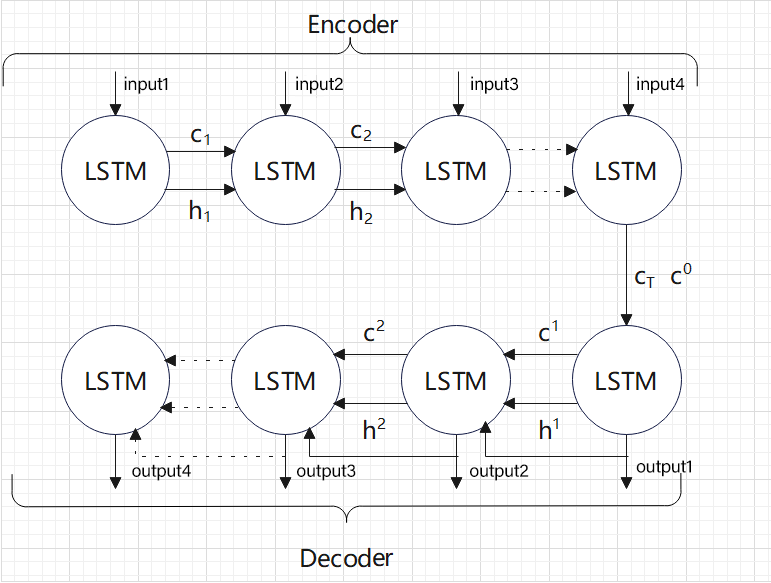}
    \caption{ED-LSTM architecture showing the Encoder  and the Decoder modules. }
    \label{fig:ED-LSTM_frame}
  \end{figure}

The    encoder-decoder architecture (ED-LSTM) proposed by Sutskever et al. \cite{sutskever2014sequence} features two types of  cells to process the data, the encoder processes the input series data $x_1,...,x_t$ with $T$ length, then transfer the summary of past series data to the cell state $c_t$, in the encoder system. There are $T$ LSTM cells to recursive $T$ time until getting the summaries of all the series data and transfer the result into cell state $c_T$, $c_T$ is also the initial input to the decoder system (i.e., $c_T$ = $c_0^{\prime}$). This helps the LSTM cells in the decoder system to output $y_1,...,y_t^{\prime}$, and each time of update, the previous output would be used as input in the current step as shown in Figure \ref{fig:ED-LSTM_frame}.
   
  The  ED-LSTM estimates the conditional probability of the output sequence $y_1,...,y_t^{\prime}$ given the input sequence $x_1,...,x_t$ (i.e.,$p\left(y_1, \ldots, y_t^{\prime} \mid x_1, \ldots, x_t\right)$).
  
  \subsubsection{CNN}
  
 LeCun introduced CNN \cite{lecun1995convolutional} with inspiration from research on the visual cortex of the mammalian biological brain. CNN consists of an input, convolution, pooling, fully connected and output layer. The convolutional  and   pooling layers can be placed multiple times as a combination depending on  the application. The  architectural setting requirement for different applications has enabled the design of specific CNN architectures such as VGG-Net\cite{majib2021vgg} and Alex-Net\cite{chen2022alexnet} for complex computer vision tasks. The convolutional layer extracts features of the input through the convolution operation, that are then filtered by the pooling layer. Finally, information is passed to the fully connected layer, which is typically a simple neural network. Although the CNN has been prominent for computer vision and image processing tasks;    it also has been applied for time series prediction \cite{selvin2017stock,piao2019housing, chandra2021evaluation}, and hence we use it as a model for comparison. 

\subsection{Direct and recursive deep learning framework for decadal world economy outlook}

\begin{figure*}[htbp!]
    \centering
    \includegraphics[width=1\textwidth]{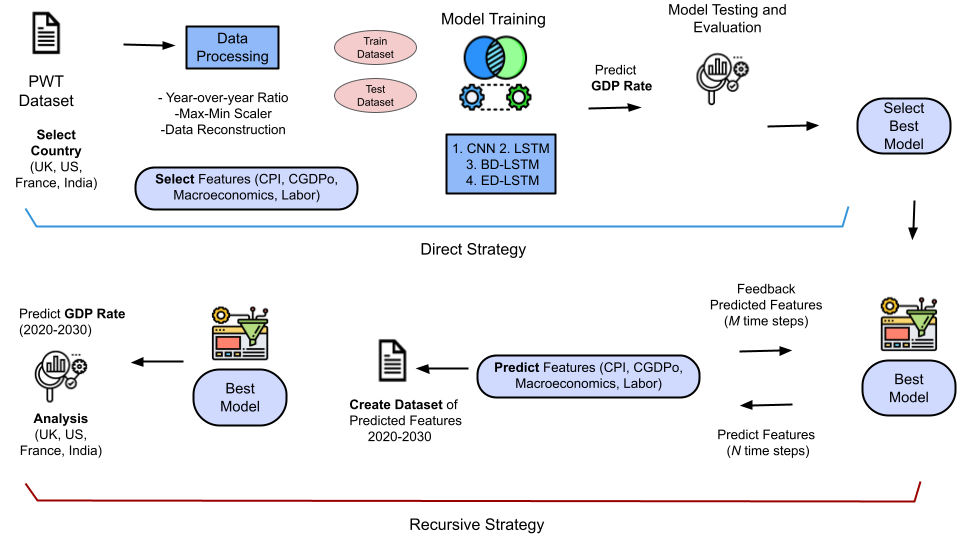}

    \caption{Deep learning-based framework for predicting decadal GDP growth rate using the direct and the recursive strategy. The model only predicts the GDP growth rate in the direct strategy using data from 1980-2010, and model testing is done for data from 2011-2019. In the recursive strategy, the model first predicts the features (economic indicators) for the decade ahead (2020-2030), which are then used to predict the decadal GDP growth rate (2020-2030). }
    \label{fig:framework}
  \end{figure*}

We apply two strategies for forecasting, which include direct and recursive, as illustrated in Figure \ref{fig:framework}. We use the direct strategy for model development and model evaluation for multistep time series prediction. We then use the  recursive strategy for decadal forecasting based on the model developed using the direct strategy. Therefore, our framework is direct and recursive, as it uses the prediction as input to the model to make future (decadal) predictions.

 We note that in the direct strategy, the model only predicts the GDP growth rate, whereas, in the recursive strategy, the model first predicts the features (economic indicators) which are then used to predict the decadal GDP growth rate. The advantage of the recursive strategy is that we can optimize the length of time step required for the deep learning model within a reasonable range, and not have the longer prediction window limit our data partitioning. We can use the optimal short-range forecast model to pursue prediction accuracy, and then use a recursive strategy to achieve our long-term forecast goal. The PWT training data set covers the period 1980-2010, Table 1 indicates that the test data set covers the period 2011-2019. We provide GDP growth rate prediction for the period 2020-2030 which lies outside the same data  with a view for later validation and for comparison with other forecasting models.

In the direct strategy, we use a fixed window of historical economic indicators from the PWT database.
We process the data before training the respective econometric and deep learning models. Our data processing features four steps including i. transform data into year-over-year ratio, ii. min-max scaler processing, iii. data reconstruction using windowing, and iv. data shuffling for creating training and test sets. We convert the data to year-on-year ratios and normalize them to remove the effects of variable ranges.  We reconstruct the data to train the respective deep learning models including the CNN, LSTM, ED-LSTM and BD-LSTM. We shuffle the data to create a training and test rest while preserving temporal dependence for a short time window, i.e. 4 years in the past to forecast 3 years in the future. We use 5 input features (economic indicators) to predict the GDP rate. We treat different countries as different datasets. After the data has been processed, we train the model and then compare the accuracy and average ranking of the models on the test data set.   We further compare the performance of deep learning models on the test data set with econometric models (ARIMA and VAR). We then select the model with the best forecasting accuracy from these models as the model for the recursive strategy. 

In the recursive strategy (Figure \ref{fig:framework}), we use the best model and data partitioning strategy evaluated in the direct strategy phase of model training. Our model features the selected economic indicators as input for a timeframe ($M$ time steps) with  multiple  output units ($N$ time steps) for prediction (multi-to-multi architecture). We apply a recursive procedure where the predictions are fed back into the model as input, since $N$ does not cover the decade and we need to predict all the economic indicators for the entire decade. We then obtain a dataset of economic indicators (2020-2030) and we used this to predict the decade-ahead GDP growth-rate for each of the selected countries and provide independent analysis.

\subsubsection{Training and test data}

We use a shuffling strategy for processing the dataset, where we take 1980-2010 period as the training data  and the 2011-2019 period as the test data set. In the shuffling strategy, we implement data reconstruction and apply windowing and then apply shuffling to randomly order the data windows to avoid any bias. In the case of  Russia, its train data set begins in 1991 but still ends in 2010. The Russian test data set share the same period with other countries as shown in Table 1. Before importing the data into the deep learning model, we need to transform the training set and test set to a structure suitable for time series forecasting.   

\begin{table}[htbp!]
    \caption{the period of train and test data set}
    \label{tab:Data_time_range}
\centering 
\begin{tabular}{ c c c } 
\hline
\hline
 & Train Data set & Test Data set  \\ 
\hline
Russia & 1991-2010 & 2011-2019 \\
\hline
Others & 1980-2010 & 2011-2019 \\ 
\hline
\hline
\end{tabular}

\end{table}
 
\begin{align}
\label{shuffled_data_structure}
  X_i &= [x_i,x_{i+1},\ldots,x_{i+m}]\\ \nonumber
 Y_i &= [y_{i+m+1},y_{i+m+2},\ldots,y_{i+m+n}]\\  \nonumber
  T_i &= [X_i;Y_i]\\  \nonumber
 D_{shuffled} &= [T_3,T_8,T_5\ldots,T_2]
\end{align} 

We use a data structure of 5 inputs as multivariate features considering 4-time steps (years)   for 3-step ahead prediction  (i.e. 3 prediction horizons).  We shuffle the  data set which considers  time windows in the reconstructed data set as given in Equation (\ref{shuffled_data_structure}). We note that the order and length inside the time window does not change, i.e.  our data window features the past 4 years of data with 5 features  to predict the next 3 years which does not change when shuffled.  Note that although we use this data processing setting to demonstrate our framework in Figure \ref{fig:framework},  other values can be used to create the training and the test dataset, i.e. the time steps (years) and the number of features, number of steps-ahead for prediction.

\section{Results}
\subsection{Technical set-up}

Our experiments first use the direct strategy to evaluate the model, using the data from 13 countries in the PWT dataset   and select the best model  according to prediction accuracy. Afterwards, we use the best-performing model to forecast the GDP growth rate for the next ten years, where we need a recursive strategy to predict  features (economic indicators) in order  to predict decadal GDP growth rate.
 
 We ran trial experiments and reviewed the literature for the hyperparameters in the respective deep learning models that include the maximum training epochs, number of hidden neurons, and learning rate parameters (alpha and beta values) for the optimization algorithm (Adam optimizer \cite{kingma2014adam}).   
 We use a maximum of 500 epochs with  a batch size of 64 in the training data for the respective deep learning models, and rectifier linear activation units (Relu) in their hidden and output layers. We use adaptive moment estimation (Adam) optimizer \cite{kingma2014adam} for training  the respective deep learning models (LSTM, BD-LSTM, ED-LSTM, and CNN). The LSTM-based deep learning models feature five input units that use 4 recursive time steps (years) and provide 3-step-ahead (year) predictions using 3 output units. 




In the VAR model, after the stationarity and co-integration test, we use the AIC as the error measure to help us determine the number of lags in the time series data.    Table \ref{tab:param_detail} gives the detail of the hyperparameter of ARIMA and VAR. We use root mean squared error (RMSE) as the main prediction accuracy measure:

\begin{equation}
\begin{split}
RMSE =\sqrt{\frac{1}{N} \sum_{i=1}^N\left(y_i-\hat{y}_i\right)^2}
\end{split}
\end{equation}

where $N$ represents the amount of samples, $y_i$ is the real value and $\hat{y}_i$ is the predicted value.

\begin{table}[htbp!]
    \caption{The details for the parameters of the VAR and ARIMA models.}
    \label{tab:param_detail}
\centering 
\begin{tabular}{ c c c } 
\hline
\hline
Country & ARIMA & VAR lags  \\ 
\hline
Japan & (4, 2, 3) & 3   \\ 
\hline
Germany & (2, 0, 2) & 1   \\ 
\hline
America & (2, 0, 2) & 1 \\ 
\hline
British & (5, 0, 3) & 1  \\ 
\hline
Canada & (1, 2, 3) & 1  \\ 
\hline
France & (0, 1, 1) & 1  \\ 
\hline
Italy & (0, 1, 1) & 1  \\ 
\hline
Australia & (0, 2, 2) & 4  \\ 
\hline
China & (1, 0, 0) & 1  \\ 
\hline
India & (0, 3, 3) & 1  \\ 
\hline
South Africa & (2, 0, 3) & 1  \\ 
\hline
Brazil & (4, 2, 2) & 1  \\ 
\hline
Russia & (0, 1, 1) & 1  \\ 
\hline
\hline
\end{tabular}

\end{table} 

We report the mean of RMSE, and 95$\%$ confidence intervals, for 30 independent experiment runs (with random model initialization of weights and biases)  for the train/test datasets. We also report the RMSE at each prediction step, given by the year (prediction horizon) for the respective deep learning models. We present results in two groups, i.e. based on developed and developing countries. Figures \ref{fig:Australia-RMSE} to  \ref{fig:United_States_RMSE} show the performance across 3-step prediction in 8 developed countries, which are Australia, Canada, France, Germany, Japan, United Kingdom and the United States.   Figures \ref{fig:Brazil_RMSE} to   \ref{fig:South_Africa_RMSE} represent the performance in the 5 developing countries; i.e. Brazil, China, India, Russia and South Africa.

\subsection{Results: Developed Countries}

Developed countries have a more stable economic environment due to their sound financial and social security systems. The inflation, employment rate and other major economic indicators do not vary substantially year-over-year. As for the GDP, the economic structure of developed countries is generally stable and gradually changing, so GDP does not have large fluctuations. As the most important group of countries in the world economy, they rarely encounter economic difficulties, except for financial crises \cite{edey2009global} and pandemics such as COVID-19 \cite{izzeldin2021impact}. 

 \begin{figure}[htbp!]
     \centering
     \begin{subfigure}[h]{0.45\textwidth}
         \centering
         \includegraphics[width=\textwidth]{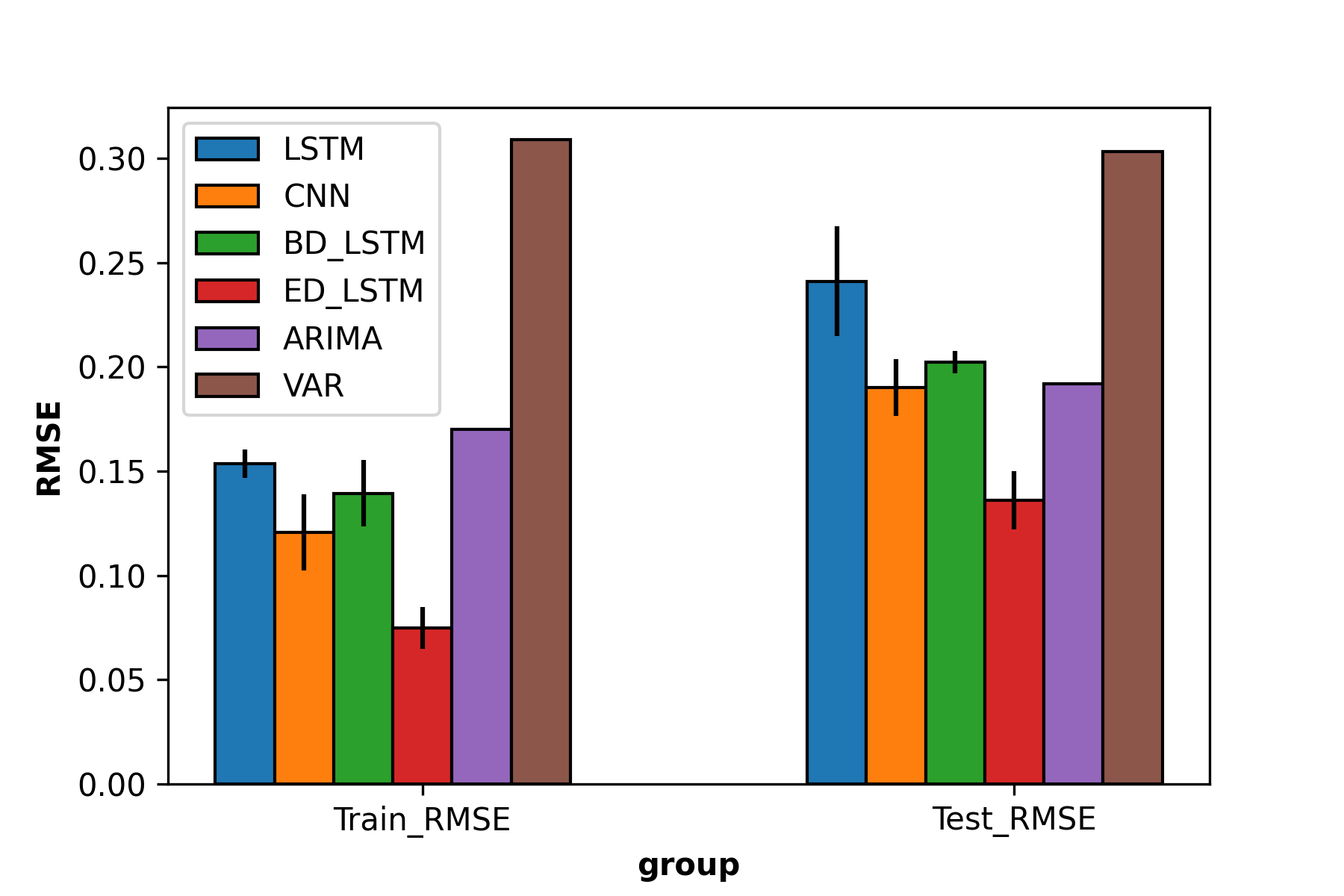}
         \caption{RMSE mean across 3-step prediction horizon}
     \end{subfigure}
     \hfill
     \begin{subfigure}[h]{0.45\textwidth}
         \centering
         \includegraphics[width=\textwidth]{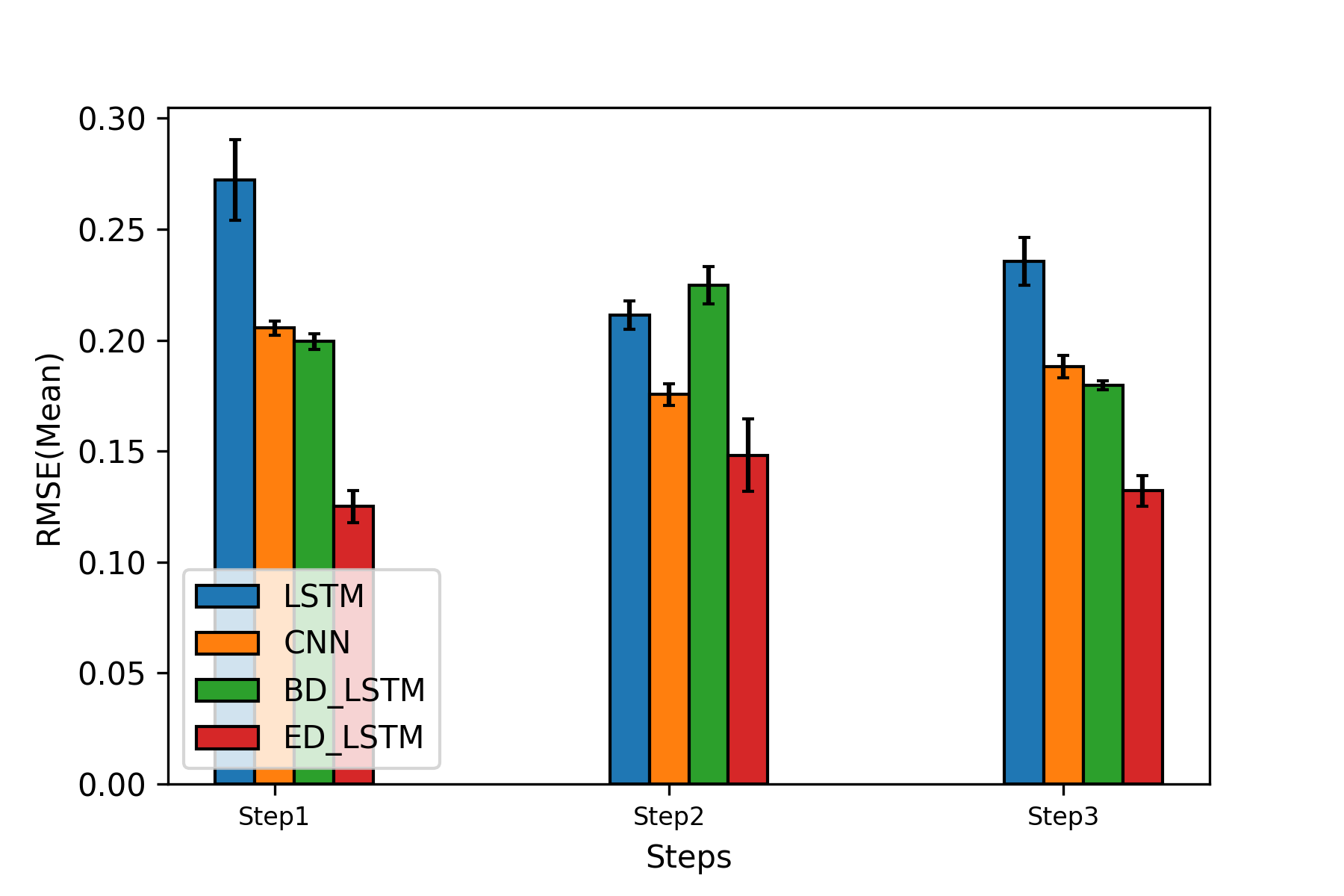}
         \caption{RMSE of each prediction horizon}
     \end{subfigure}
        \caption{Australia: performance evaluation of respective methods (RMSE mean and 95$\%$ confidence intervals as error bar).}
        \label{fig:Australia-RMSE}
\end{figure}

\begin{table*}
    \caption{Australia: Prediction accuracy showing RMSE mean and 95 $\%$ confidence interval (±).}
    \label{tab:Australia-table}
\begin{tabular}{c c c c c c c}
\hline
      & ARIMA & VAR & LSTM & BD-LSTM & ED-LSTM & CNN  \\ 
\hline
\hline
Train &0.1699  & 0.3088  &0.1537$\pm$0.0048 &0.1393$\pm$0.0056 & 0.0748$\pm$0.0035 & 0.1207$\pm$0.0.006\\ 
Test &0.1919  & 0.3032  &0.241$\pm$0.0187 &0.2022$\pm$0.0019 & 0.1359$\pm$0.0048 & 0.1901$\pm$0.0049\\ 
\hline
Step1 &  &  &0.2721$\pm$0.0065 &0.1994$\pm$0.0013 & 0.1251$\pm$0.0026 & 0.2054$\pm$0.0012\\ 
Step2 &  &   &0.2112$\pm$0.0022 &0.2247$\pm$0.0028 & 0.1481$\pm$0.0051 & 0.1755$\pm$0.0016\\ 
Step3 &  &  &0.2355$\pm$0.0036 &0.1798$\pm$0.0007 & 0.1321$\pm$0.0023 & 0.188$\pm$0.0017\\ 
\hline
\hline
\end{tabular}
\end{table*} 

In Figure \ref{fig:Australia-RMSE} Panel (a) for the case of Australia (Table \ref{tab:Australia-table}), we see that the  deep learning models produce better accuracy (RMSE) than the traditional models (ARIMA and VAR) both on the training data and test datasets. The traditional time series models provide more consistent performance across the two sets. The results of the ED-LSTM model appear consistently superior to all other models, while the prediction capabilities of CNN and ARIMA are approximately equivalent. We find that BD-LSTM is only slightly behind the former two models. The LSTM model ranks fifth, but with the largest variance, meaning that these models are the most unstable. Note that the traditional models do not feature step-ahead predictions in Table \ref{tab:Australia-table}, and for the rest of the countries hereon.   Figure \ref{fig:Australia-RMSE}  Panel (b) shows ED-LSTM provides the best accuracy.


\begin{table*}
    \caption{Canada: Prediction accuracy showing RMSE mean and 95 $\%$ confidence interval (±).}
    \label{tab:Canada-table}
\begin{tabular}{c c c c c c c}
\hline
      & ARIMA & VAR & LSTM & BD-LSTM & ED-LSTM & CNN  \\ 
\hline
\hline
Train &0.1873  & 0.2260 &0.1658$\pm$0.0081 &0.1553$\pm$0.0057 & 0.1642$\pm$0.0040 & 0.1671$\pm$0.0.009\\ 
Test &0.2630  & 0.1277  &0.4422$\pm$0.06 &0.2624$\pm$0.012 & 0.2345$\pm$0.0037 & 0.2507$\pm$0.0065\\ 
\hline
Step1 &  &  &0.4733$\pm$0.0188 &0.2884$\pm$0.0024 & 0.2938$\pm$0.0013 & 0.2937$\pm$0.0027\\ 
Step2 &  &   &0.3562$\pm$0.0188 &0.2481$\pm$0.0002 & 0.2319$\pm$0.0018 & 0.2381$\pm$0.0004\\ 
Step3 &  &  &0.4817$\pm$0.0345 &0.2485$\pm$0.0049 & 0.1573$\pm$0.0026 & 0.2133$\pm$0.0011\\ 
\hline
\hline
\end{tabular}
\end{table*}


 \begin{figure}[htbp!]
     \centering
     \begin{subfigure}{0.45\textwidth}
         \centering
         \includegraphics[width=\textwidth]{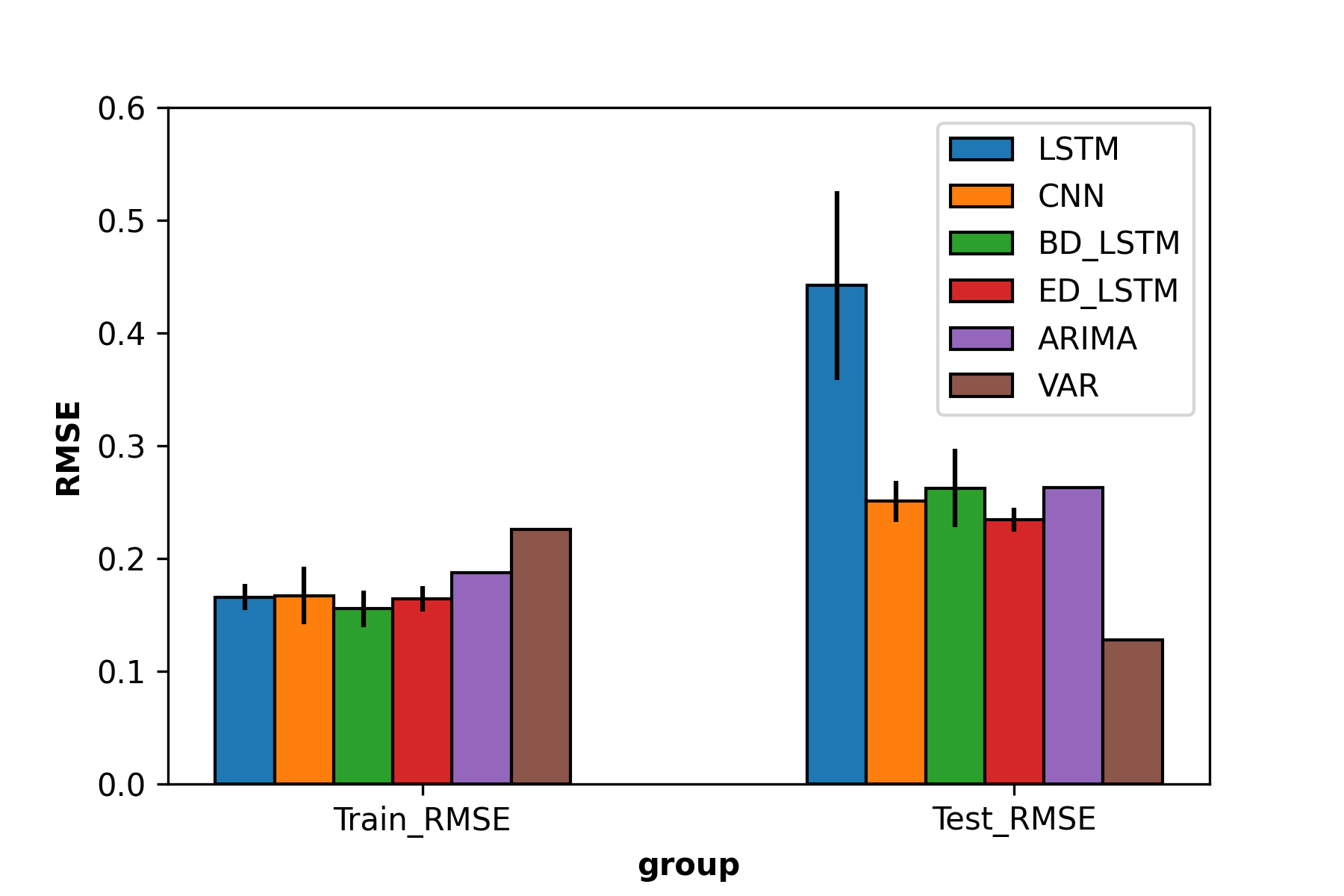}
         \caption{RMSE mean across 3-step prediction horizon}
     \end{subfigure}
     \hfill
     \begin{subfigure}{0.45\textwidth}
         \centering
         \includegraphics[width=\textwidth]{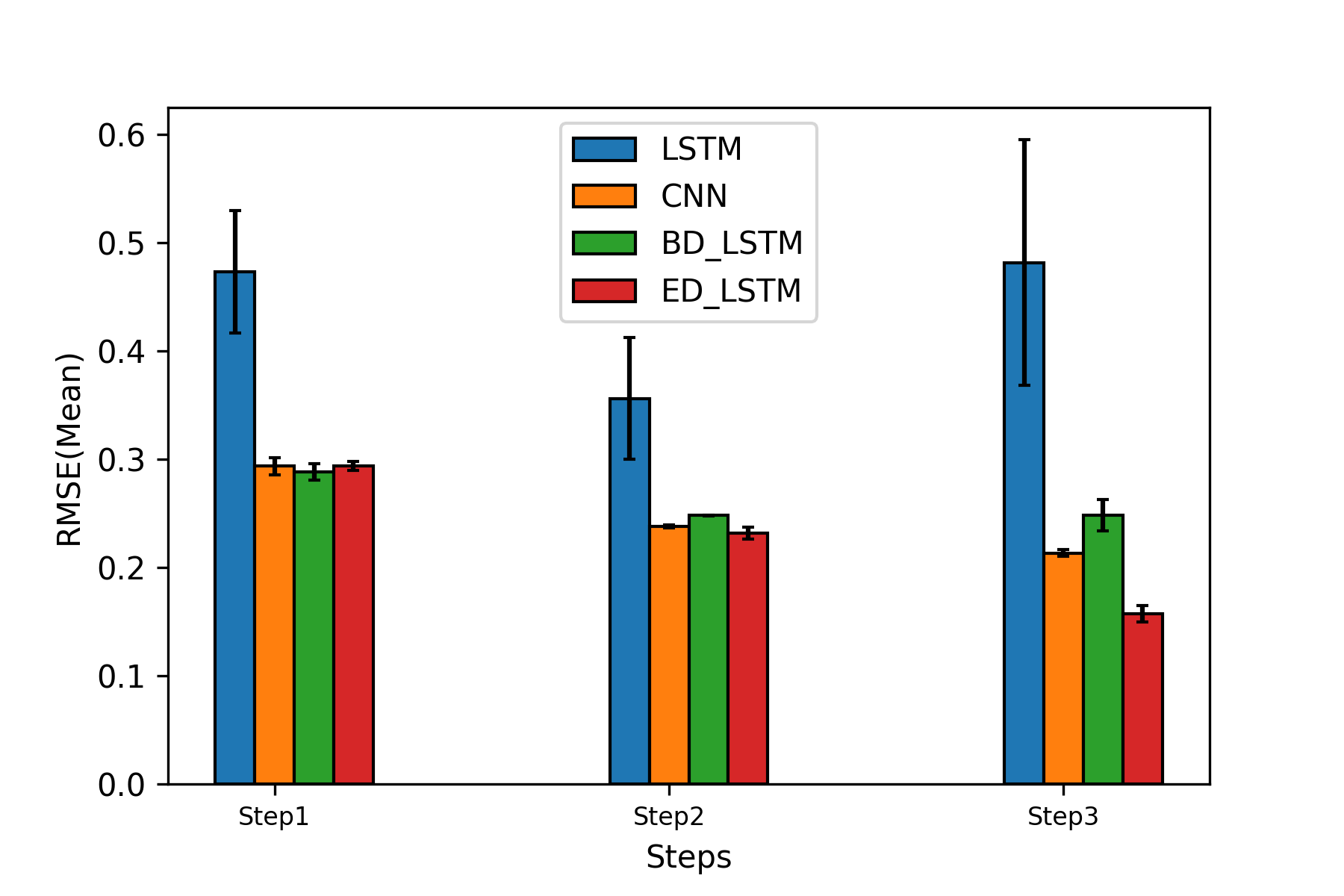}
         \caption{RMSE of each prediction horizon}
     \end{subfigure}
   
        \caption{Canada: performance evaluation of respective methods (RMSE mean and 95$\%$ confidence intervals as error bar).}
        \label{fig:Canada-RMSE}
\end{figure}

\begin{figure}[htbp!]
     \centering
     
     \begin{subfigure}{0.45\textwidth}
         \centering
         \includegraphics[width=\textwidth]{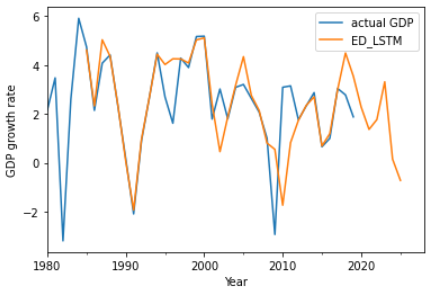}
         \caption{Best ED-LSTM model}
     \end{subfigure}
     \hfill
     \begin{subfigure}{0.45\textwidth}
         \centering
         \includegraphics[width=\textwidth]{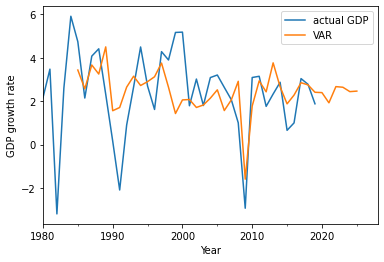}
         \caption{VAR model}
     \end{subfigure}
        \caption{Canada: prediction plots showing the performance of the respective models.}
        \label{fig:Canada-RMSElineplot}
\end{figure}

We present the results for Canada  in Figure \ref{fig:Canada-RMSE} and Table \ref{tab:Canada-table}, where the VAR model provides the best performance for the test dataset,  exceeding its own training set results. The other four models have relatively similar results, with the ED-LSTM exhibiting slightly better performance than CNN. We find that CNN ranks second among the deep learning models. The RMSE of  LSTM is the worst of these models, and its variance is substantially large, indicating that the model is extremely volatile. Panel (b)  in Figure \ref{fig:Canada-RMSE} demonstrates that ED-LSTM, BD-LSTM and CNN are approximately equivalent at the first step of prediction, but as the number of steps increases  the ED-LSTM demonstrates improved performance.  We compare the first-step prediction plot of the ED-LSTM with  VAR   in Figure \ref{fig:Canada-RMSElineplot} with actual GDP.

 \begin{figure}[htbp!]
     \centering
     \begin{subfigure}[h]{0.45\textwidth}
         \centering
         \includegraphics[width=\textwidth]{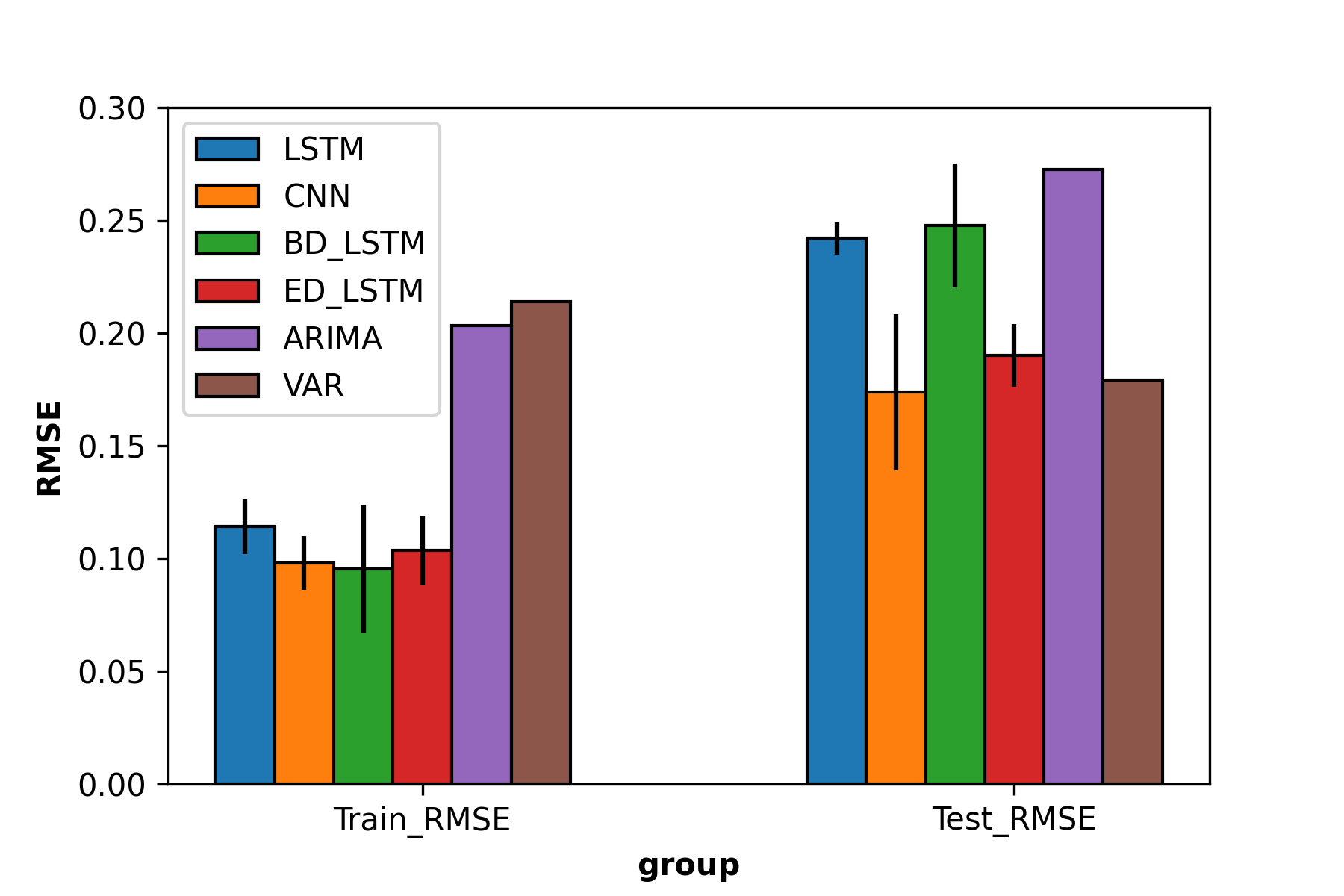}
         \caption{RMSE mean across 3-step prediction horizon}
     \end{subfigure}
     \hfill
     \begin{subfigure}[h]{0.45\textwidth}
         \centering
         \includegraphics[width=\textwidth]{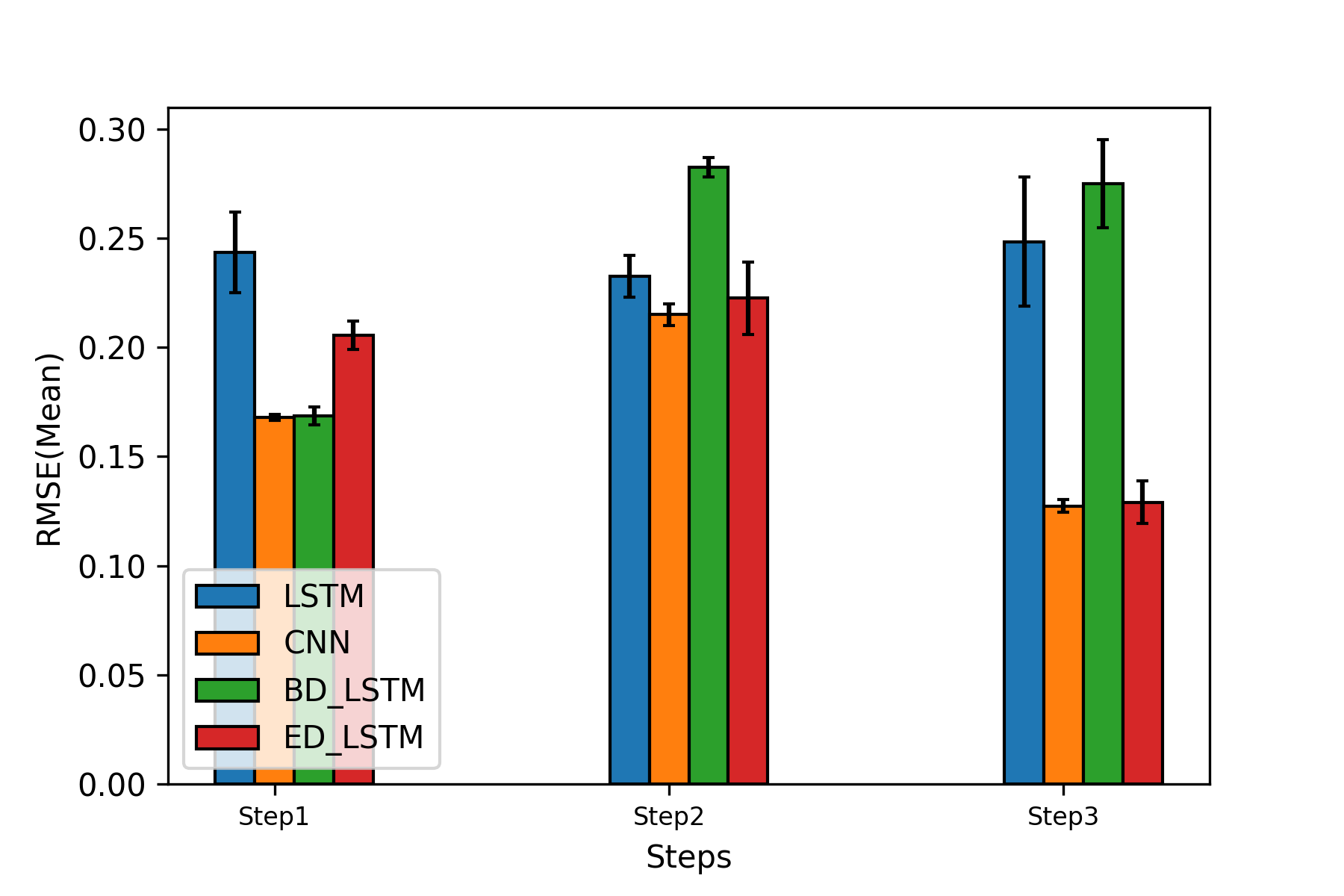}
         \caption{RMSE of each prediction horizon}
     \end{subfigure}
        \caption{France: performance evaluation of respective methods (RMSE mean and 95$\%$ confidence interval as error bar)}
        \label{fig:France-RMSE}
\end{figure}

\begin{table*}
    \caption{France: Prediction accuracy showing RMSE mean and 95 $\%$ confidence interval (±).}
    \label{tab:France-table}
\begin{tabular}{c c c c c c c}
\hline
      & ARIMA & VAR & LSTM & BD-LSTM & ED-LSTM & CNN  \\ 
\hline
\hline
Train &0.2034  & 0.2138 &0.1144$\pm$0.0087 &0.0954$\pm$0.0102 & 0.1035$\pm$0.0055 & 0.0979$\pm$0.0.0042\\ 
Test &0.2723  & 0.1793 &0.242$\pm$0.0051 &0.2477$\pm$0.0098 & 0.1901$\pm$0.0049 & 0.1738$\pm$0.0124\\ 
\hline
Step1 &  &  &0.2433$\pm$0.0062 &0.1686$\pm$0.0013 & 0.2055$\pm$0.0022 & 0.1679$\pm$0.0004\\ 
Step2 &  &   &0.2325$\pm$0.0032 &0.2824$\pm$0.0015 & 0.2224$\pm$0.0052 & 0.2149$\pm$0.0017\\ 
Step3 &  &  &0.2483$\pm$0.0098 &0.2748$\pm$0.0067 & 0.1273$\pm$0.0010 & 0.1442$\pm$0.0012\\
\hline
\hline
\end{tabular}
\end{table*}

Figure \ref{fig:France-RMSE} presents the results for forecasting French GDP (Table \ref{tab:France-table}), where we find that the ARIMA model is worse than all the models, while the VAR model occupies the first position. Among the four deep learning models, CNN has a better mean RMSE but slightly higher variance than the ED-LSTM for the test dataset.  

 \begin{figure}[htbp!]
     \centering
     \begin{subfigure}[h]{0.45\textwidth}
         \centering
         \includegraphics[width=\textwidth]{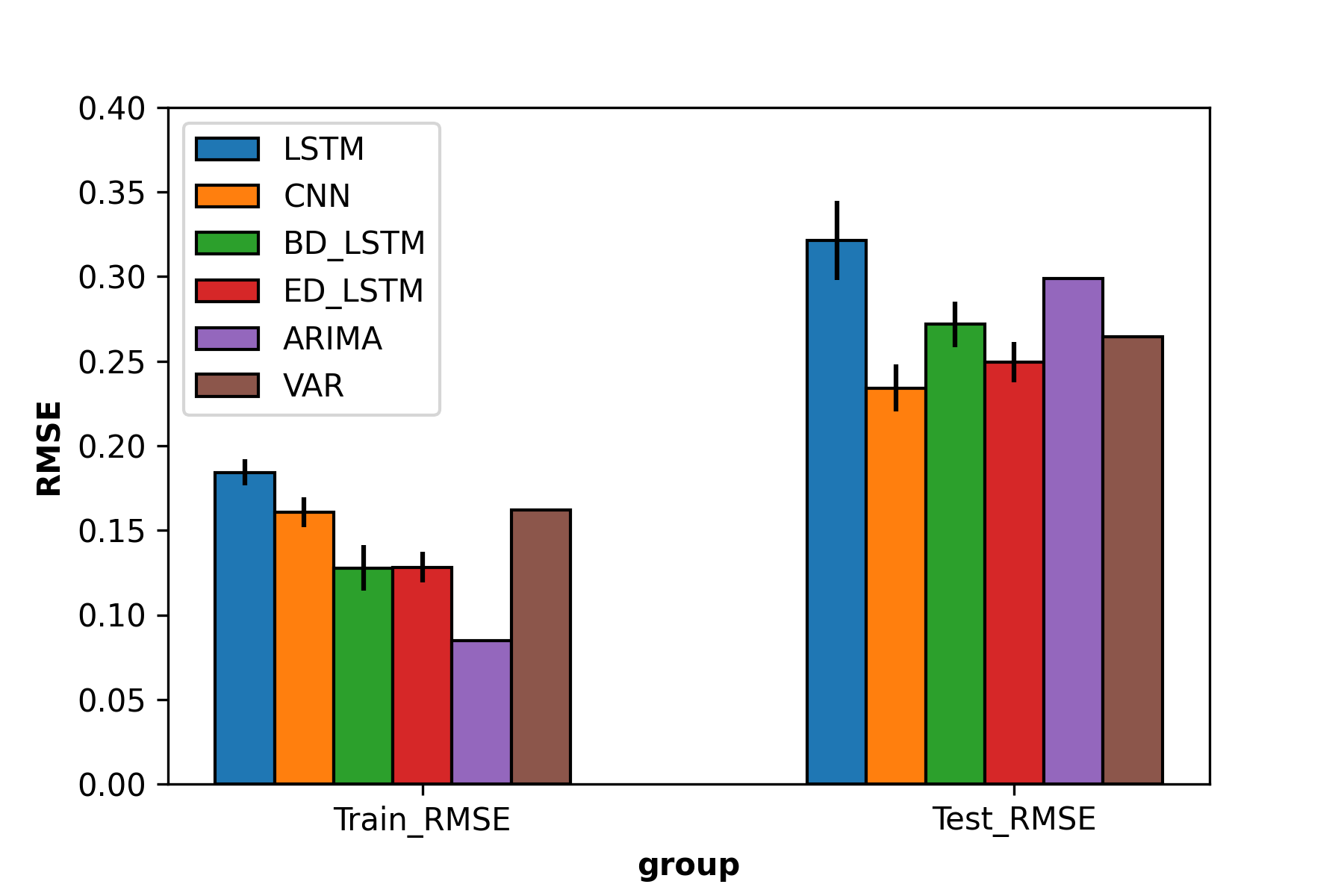}
         \caption{RMSE mean across 3-step prediction horizon}
     \end{subfigure}
     \hfill
     \begin{subfigure}[h]{0.45\textwidth}
         \centering
         \includegraphics[width=\textwidth]{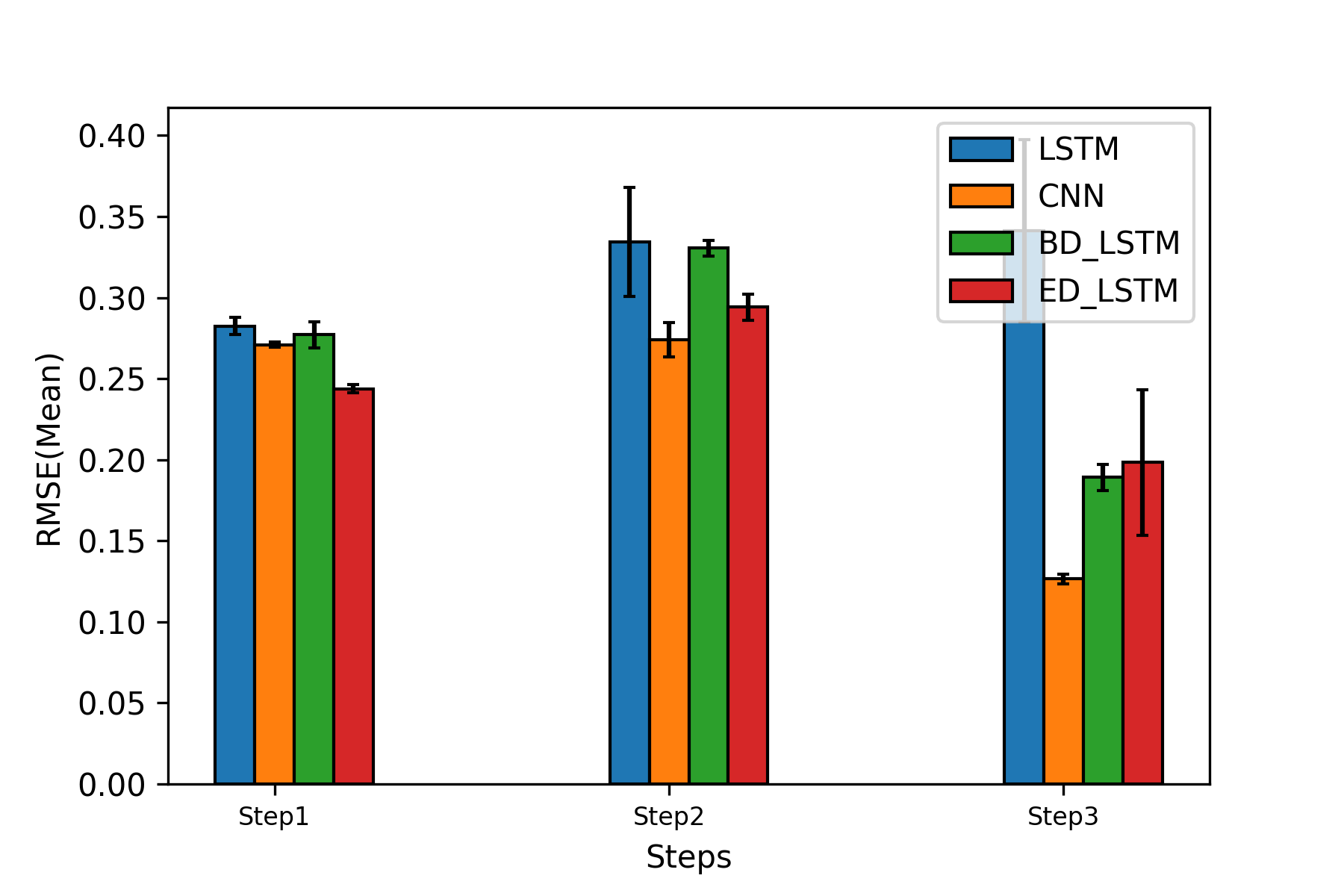}
         \caption{RMSE of each prediction horizon}
     \end{subfigure}
        \caption{Germany: performance evaluation of respective methods (RMSE mean and 95$\%$ confidence intervals as error bar).}
        \label{fig:Germany-RMSE}
\end{figure}

\begin{table*}
    \caption{Germany: Prediction accuracy showing RMSE mean and 95 $\%$ confidence interval (±).}
    \label{tab:Germany-table}
\begin{tabular}{c c c c c c c}
\hline
      & ARIMA & VAR & LSTM & BD-LSTM & ED-LSTM & CNN  \\ 
\hline
\hline
Train &0.0847  & 0.1620 &0.1843$\pm$0.0056 &0.1277$\pm$0.0047 & 0.1281$\pm$0.0032 & 0.1607$\pm$0.0.032\\ 
Test &0.2987  & 0.2644 &0.3215$\pm$0.016 &0.2719$\pm$0.0048 & 0.2495$\pm$0.0042 & 0.2342$\pm$0.0050\\ 
\hline
Step1 &  &  &0.2825$\pm$0.0018 &0.2771$\pm$0.003 & 0.2437$\pm$0.0008 & 0.2709$\pm$0.0018\\ 
Step2 &  &   &0.3344$\pm$0.0112 &0.3306$\pm$0.0017 & 0.2941$\pm$0.003 & 0.2741$\pm$0.0036\\ 
Step3 &  &  &0.3413$\pm$0.0187 &0.1891$\pm$0.0027 & 0.1983$\pm$0.015 & 0.1266$\pm$0.001\\ 
\hline
\hline
\end{tabular}
\end{table*}

Figure \ref{fig:Germany-RMSE} presents the prediction performance for Germany (Table \ref{tab:Germany-table}), where Panel (a) shows that the CNN and ED-LSTM models outperform traditional time series models. We find that CNN ranks first, while ED-LSTM and VAR are in the second and third positions for the test dataset, respectively. 

 \begin{figure}[htbp!]
     \centering
     \begin{subfigure}[h]{0.45\textwidth}
         \centering
         \includegraphics[width=\textwidth]{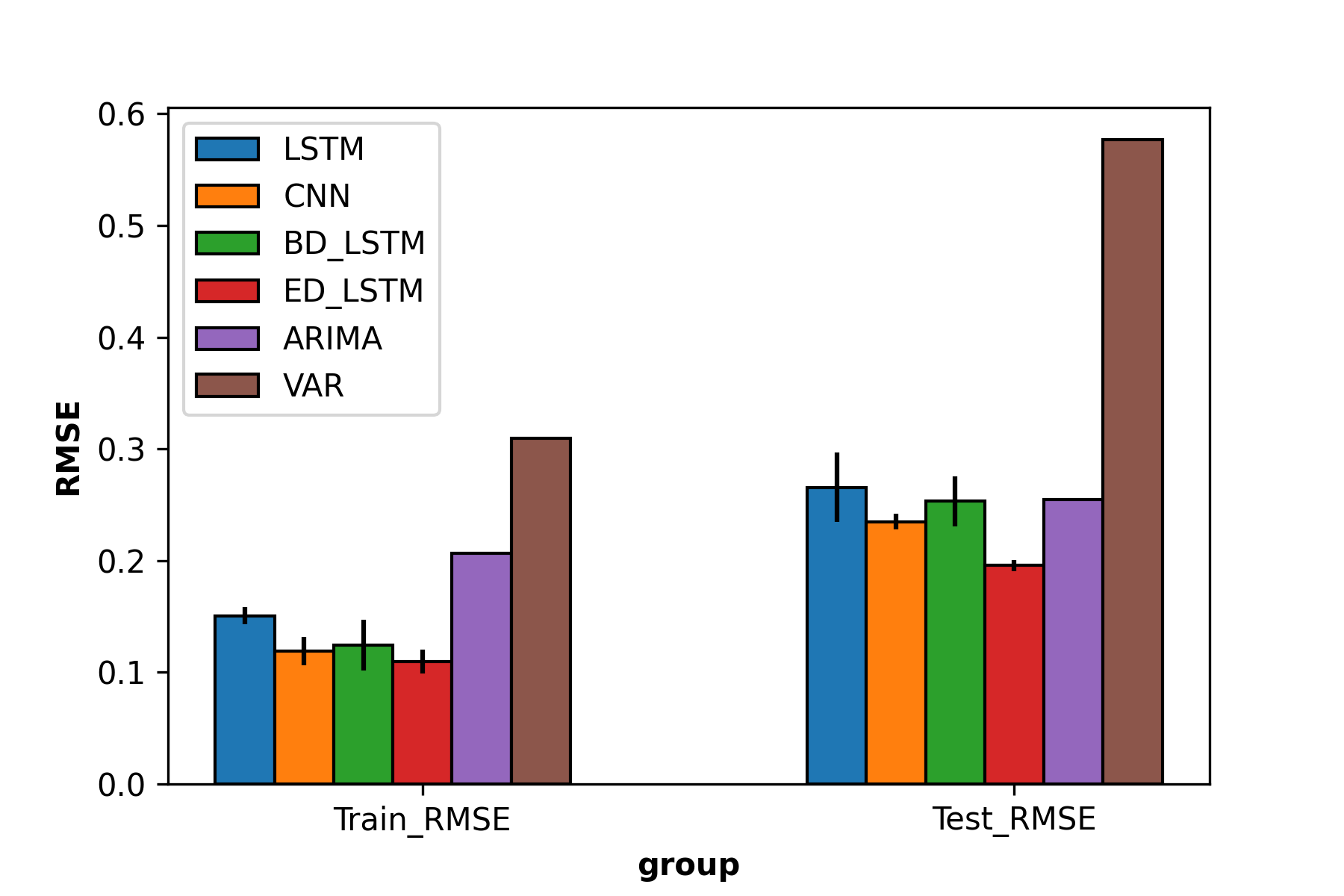}
         \caption{RMSE mean across 3-step prediction horizon}
     \end{subfigure}
     \hfill
     \begin{subfigure}[h]{0.45\textwidth}
         \centering
         \includegraphics[width=\textwidth]{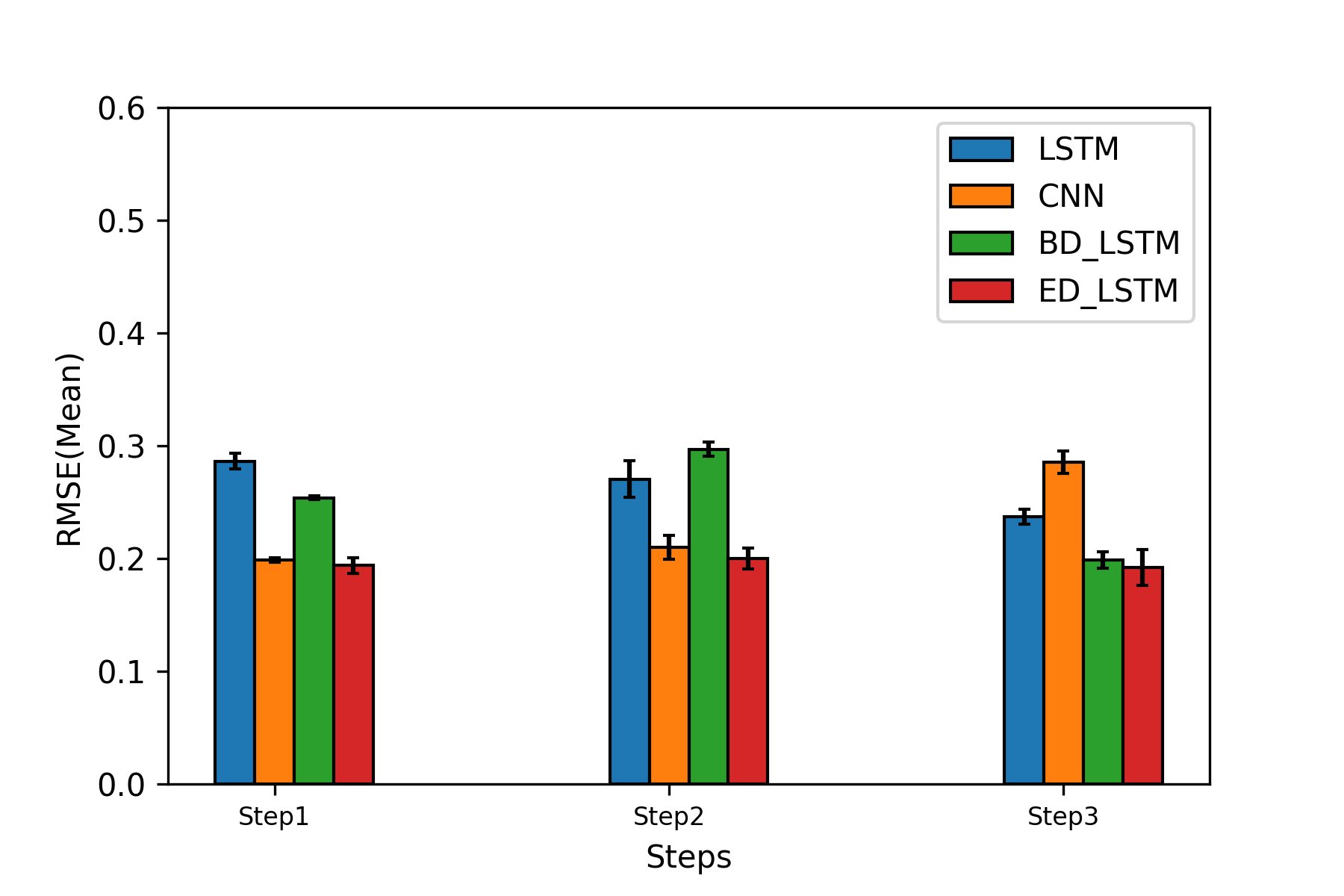}
         \caption{RMSE of each prediction horizon}
     \end{subfigure}
        \caption{Italy: performance evaluation of respective methods (RMSE mean and 95$\%$ confidence intervals as error bar).}
        \label{fig:Italy-RMSE}
\end{figure}

\begin{table*}
    \caption{Italy: RMSE mean and 95 $\%$ confidence interval (±).}
    \label{tab:Italy-table}
\begin{tabular}{c c c c c c c}
\hline
      & ARIMA & VAR & LSTM & BD-LSTM & ED-LSTM & CNN  \\ 
\hline
\hline
Train &0.2069  & 0.3095 &0.1506$\pm$0.0055 &0.1245$\pm$0.0081 & 0.1098$\pm$0.0037 & 0.1191$\pm$0.0.045\\ 
Test &0.2544  & 0.5767 &0.2656$\pm$0.0223 &0.2531$\pm$0.0080 & 0.1956$\pm$0.0017 & 0.2348$\pm$0.0025\\ 
\hline
Step1 &  &  &0.2863$\pm$0.0023 &0.2538$\pm$0.0006 & 0.1938$\pm$0.0023 & 0.1988$\pm$0.0006\\ 
Step2 &  &   &0.2705$\pm$0.0054 &0.291$\pm$0.002 & 0.2001$\pm$0.031 & 0.21$\pm$0.0107\\ 
Step3 &  &  &0.2371$\pm$0.0023 &0.1988$\pm$0.0024 & 0.1919$\pm$0.0052 & 0.2856$\pm$0.033\\ 
\hline
\hline
\end{tabular}
\end{table*}

We present the results for Italy  in Figure \ref{fig:Italy-RMSE} and Table \ref{tab:Italy-table}, where the CNN and ED-LSTM models provide the best performance for  both training and test datasets. In addition, they appear more stable than the other deep learning models. Figure \ref{fig:Italy-RMSE} Panel (b) indicates that the ED-LSTM model has stable and strong performance across all three steps.

 \begin{figure}[htbp!]
     \centering
     \begin{subfigure}[h]{0.45\textwidth}
         \centering
         \includegraphics[width=\textwidth]{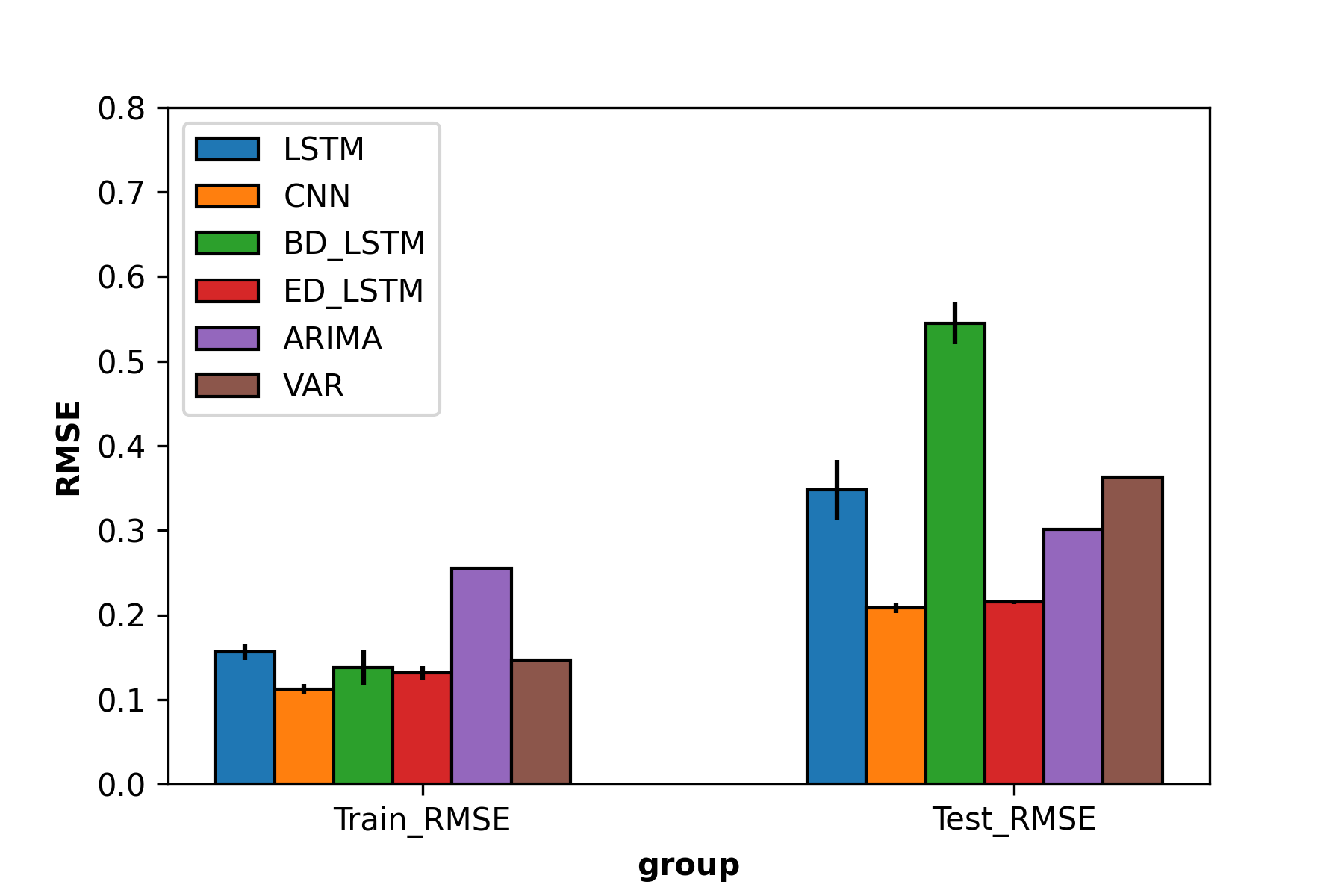}
         \caption{RMSE across 3-step prediction horizon}
     \end{subfigure}
     \hfill
     \begin{subfigure}[h]{0.45\textwidth}
         \centering
         \includegraphics[width=\textwidth]{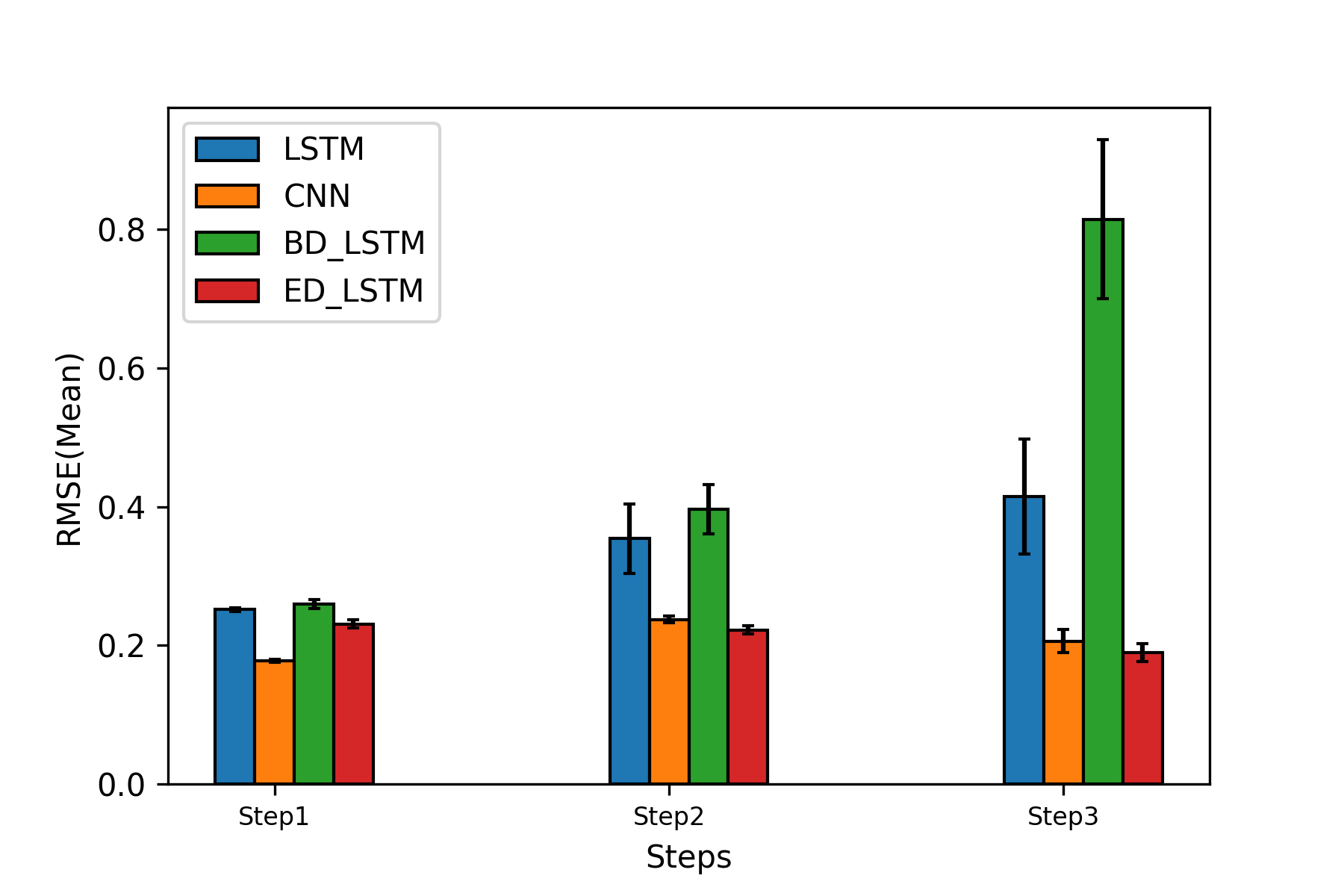}
         \caption{RMSE of each prediction horizon}
     \end{subfigure}
        \caption{Japan: performance evaluation of respective methods (RMSE mean and 95$\%$ confidence intervals as error bar).}
        \label{fig:Japan-RMSE}
\end{figure}

\begin{table*}
    \caption{Japan: RMSE mean and 95 $\%$ confidence interval (±).}
    \label{tab:Japan-table}
\begin{tabular}{c c c c c c c}
\hline
      & ARIMA & VAR & LSTM & BD-LSTM & ED-LSTM & CNN  \\ 
\hline
\hline
Train &0.2554  & 0.1466 &0.1561$\pm$0.0066 &0.1375$\pm$0.0075 & 0.1312$\pm$0.0031 & 0.1125$\pm$0.0.0021\\ 
Test &0.3014  & 0.3626 &0.3483$\pm$0.0253 &0.5448$\pm$0.0087 & 0.2151$\pm$0.0009 & 0.2086$\pm$0.0022\\ 
\hline 
Step1 &  &  &0.2516$\pm$0.0009 &0.2592$\pm$0.0022 & 0.2308$\pm$0.0019 & 0.1778$\pm$0.0006\\ 
Step2 &  &   &0.354$\pm$0.0017 &0.3966$\pm$0.0122 & 0.2221$\pm$0.021 & 0.2374$\pm$0.0018\\ 
Step3 &  &  &0.4143$\pm$0.0272 &0.8145$\pm$0.0382 & 0.1899$\pm$0.0043 & 0.206$\pm$0.006\\ 
\hline
\hline
\end{tabular}
\end{table*}

We present the results for Japan in Figure \ref{fig:Japan-RMSE} and Table \ref{tab:Japan-table},  where CNN and ED-LSTM models provide superior prediction performance overall. The CNN model has slightly lower errors in the test data, and the other models have significantly higher errors in both training and test data. Looking at Figure \ref{fig:Japan-RMSE} Panel (b), we observe that as the number of steps increases the CNN and ED-LSTM models maintain consistent performance, while the other deep learning models deteriorate.

 \begin{figure}[htbp!]
     \centering
     \begin{subfigure}[h]{0.45\textwidth}
         \centering
         \includegraphics[width=\textwidth]{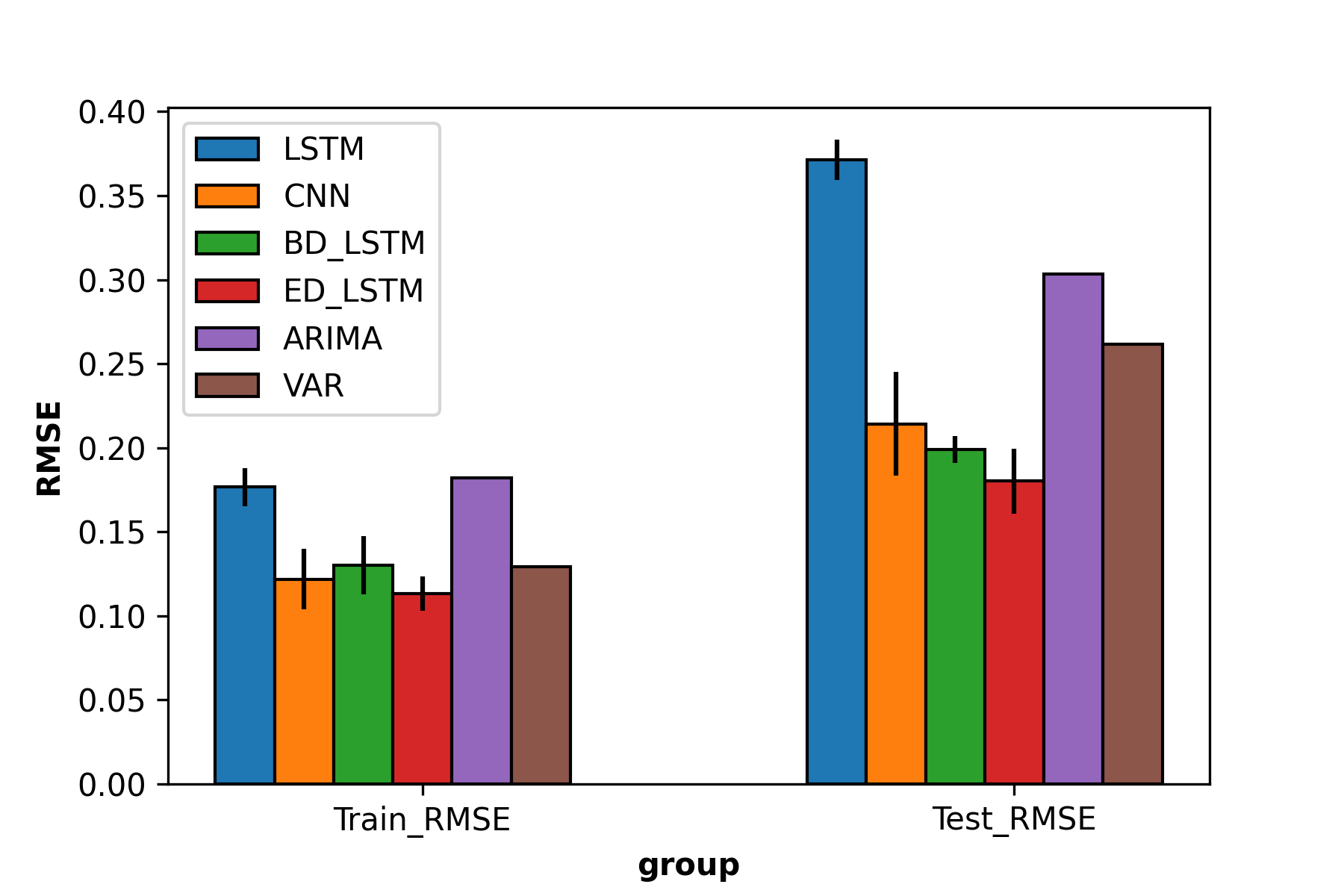}
         \caption{RMSE mean across 3-step prediction horizon}
     \end{subfigure}
     \hfill
     \begin{subfigure}[h]{0.45\textwidth}
         \centering
         \includegraphics[width=\textwidth]{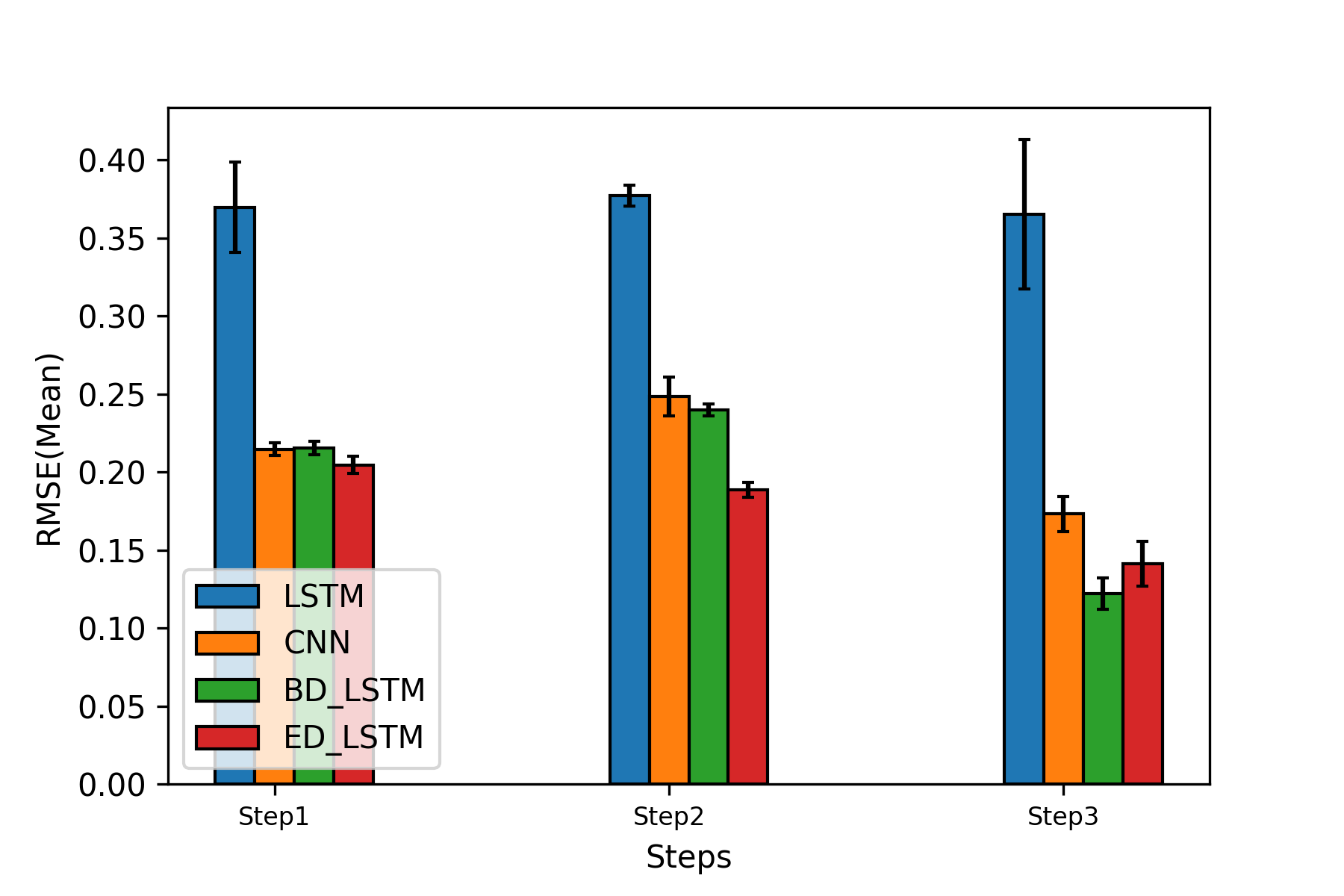}
         \caption{RMSE of each prediction horizon}
     \end{subfigure}
        \caption{United Kingdom: performance evaluation of respective methods (RMSE mean and 95$\%$ confidence intervals as error bar).}
        \label{fig:United_Kingdom_RMSE}
\end{figure}

\begin{table*}
    \caption{United Kingdom: RMSE mean and 95 $\%$ confidence interval (±).}
    \label{tab:United_Kingdom_table}
\begin{tabular}{c c c c c c c}
\hline
      & ARIMA & VAR & LSTM & BD-LSTM & ED-LSTM & CNN  \\ 
\hline
\hline
Train &0.1821  & 0.1291 &0.1767$\pm$0.0081 &0.1302$\pm$0.0062 & 0.1132$\pm$0.0036 & 0.1219$\pm$0.0.0064\\ 
Test &0.3035  & 0.2614&0.3714$\pm$0.00843 &0.199$\pm$0.0028 & 0.1803$\pm$0.0069 & 0.2143$\pm$0.011\\ 
\hline 
Step1 &  &  &0.3697$\pm$0.009 &0.2154$\pm$0.0015 & 0.2046$\pm$0.0018 & 0.2146$\pm$0.0013\\ 
Step2 &  &   &0.3772$\pm$0.0025 &0.2397$\pm$0.0013 & 0.1886$\pm$0.015 & 0.2484$\pm$0.041\\ 
Step3 &  &  &0.3652$\pm$0.0162 &0.122$\pm$0.0033 & 0.1411$\pm$0.0048 & 0.1731$\pm$0.0037\\ 
\hline
\hline
\end{tabular}
\end{table*}

 We present the results for the United Kingdom  in Figure \ref{fig:United_Kingdom_RMSE} and Table \ref{tab:United_Kingdom_table}. In  \ref{fig:United_Kingdom_RMSE} Panel (a), apart from the LSTM, the deep learning models perform better than traditional time series models on the test data. The ED-LSTM appears to provide the best performance; however, its variance upper bound exceeds the mean performance of the BD-LSTM model. 
 
 \begin{figure}[htbp!]
     \centering
     \begin{subfigure}[h]{0.45\textwidth}
         \centering
         \includegraphics[width=\textwidth]{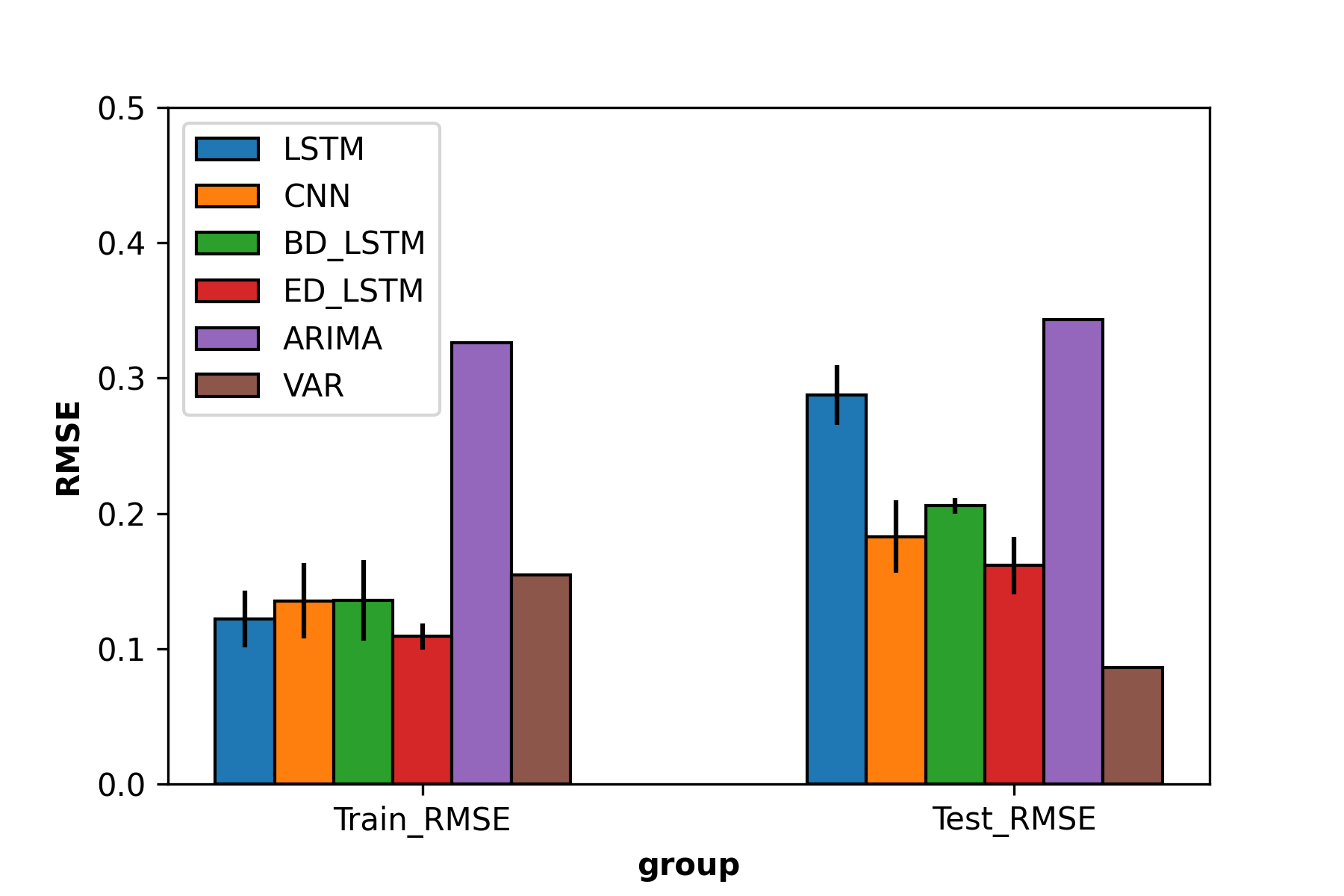}
         \caption{RMSE mean across 3-step prediction horizon}
     \end{subfigure}
     \hfill
     \begin{subfigure}[h]{0.45\textwidth}
         \centering
         \includegraphics[width=\textwidth]{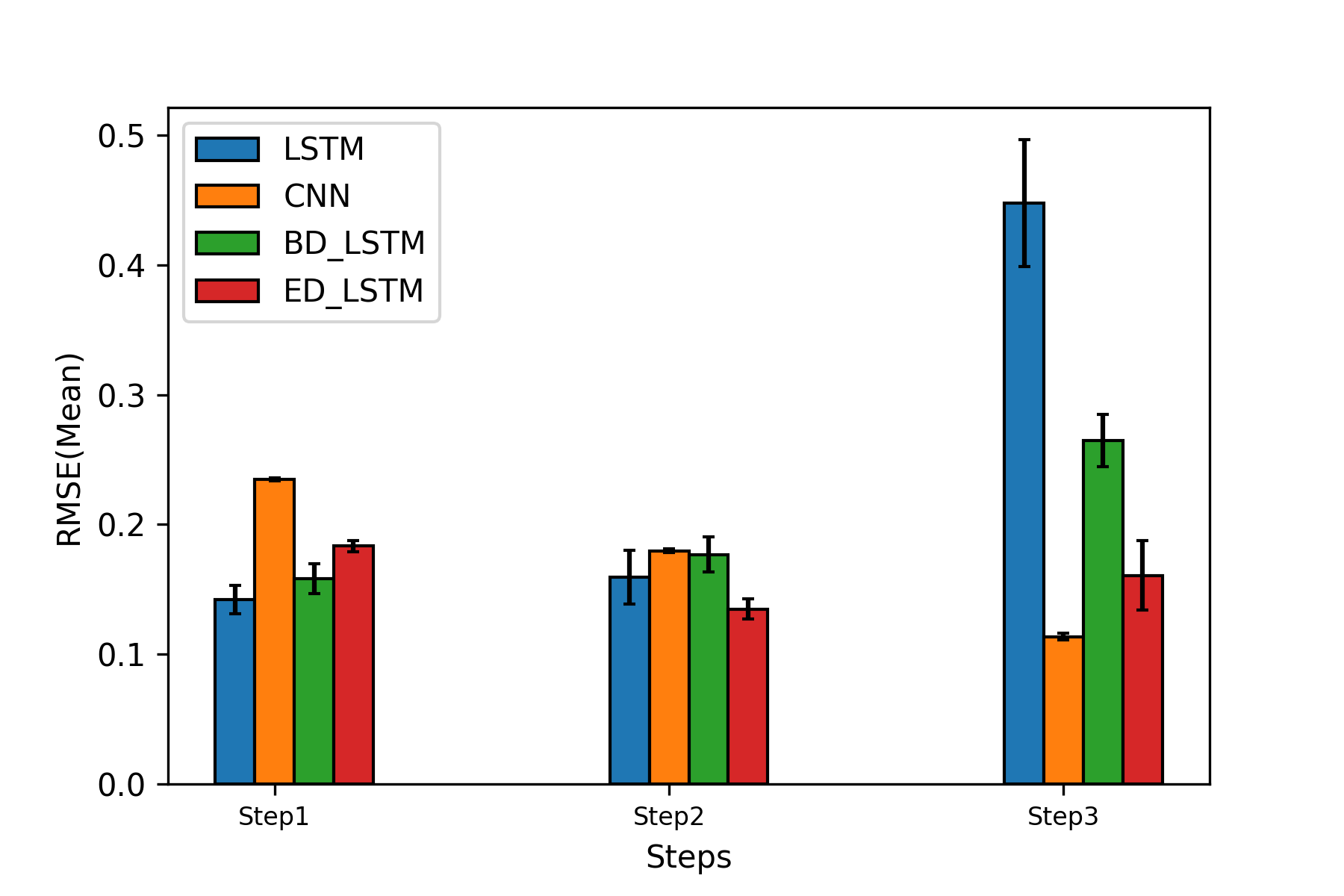}
         \caption{RMSE of each prediction horizon}
     \end{subfigure} 
        \caption{United States: performance evaluation of respective methods (RMSE mean and 95$\%$ confidence interval as error bar).}
        \label{fig:United_States_RMSE}
\end{figure}

 \begin{figure}[htbp!]
     \centering 
     \begin{subfigure}[h]{0.45\textwidth}
         \centering
         \includegraphics[width=\textwidth]{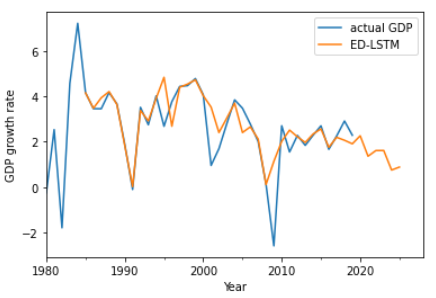}
         \caption{Best ED-LSTM model}
     \end{subfigure}
     \hfill
     \begin{subfigure}[h]{0.45\textwidth}
         \centering
         \includegraphics[width=\textwidth]{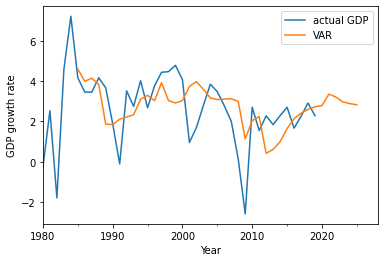}
         \caption{VAR model}
     \end{subfigure}
        \caption{United States: prediction plot using selected models comparing with actual GDP.}
        \label{fig:United_States_RMSEplot}
\end{figure}

\begin{table*}
    \caption{United States: Prediction accuracy showing RMSE mean and 95 $\%$ confidence interval (±).}
    \label{tab:United_States_table}
\begin{tabular}{c c c c c c c}
\hline
      & ARIMA & VAR & LSTM & BD-LSTM & ED-LSTM & CNN  \\ 
\hline
\hline
Train &0.3263  & 0.1544 &0.1221$\pm$0.0149 &0.1359$\pm$0.0106 & 0.109$\pm$0.0034 & 0.1354$\pm$0.0.0099\\ 
Test &0.3435  & 0.0863&0.2875$\pm$0.0159 &0.2056$\pm$0.0021 & 0.1615$\pm$0.0075 & 0.1828$\pm$0.0095\\ 
\hline
Step1 &  &  &0.1421$\pm$0.004 &0.1581$\pm$0.0035 & 0.1835$\pm$0.0014 & 0.2347$\pm$0.0003\\ 
Step2 &  &   &0.1592$\pm$0.0073 &0.1768$\pm$0.043 & 0.1349$\pm$0.0026 & 0.1798$\pm$0.0004\\ 
Step3 &  &  &0.4475$\pm$0.0193 &0.2646$\pm$0.007 & 0.1608$\pm$0.0088 & 0.1135$\pm$0.0008\\
\hline
\hline
\end{tabular}
\end{table*}

We present the results for the United States in Figure \ref{fig:United_States_RMSE} and Table \ref{tab:United_States_table}, where Panel (a) indicates that the two traditional time series models rank first and last, the VAR model has the best performance among these 6 models while ARIMA performs worst. Among the deep learning models, the CNN and ED-LSTM perform better than the BD-LSTM and LSTM, although BD-LSTM has the least variance. In the first 3 steps, although the four models perform similarly in the first two steps,   as the number of steps increases, the prediction accuracy of LSTM and BD-LSTM gradually degrades. The ED-LSTM model performance is consistently stable, while the CNN improves as the forecast horizon increases.   Figure \ref{fig:United_States_RMSEplot} presents prediction plots of the United States GDP using ED-LSTM and VAR and compares them with actual the GDP.

 \begin{figure}[htbp!]
     \centering
     \begin{subfigure}[h]{0.45\textwidth}
         \centering
         \includegraphics[width=\textwidth]{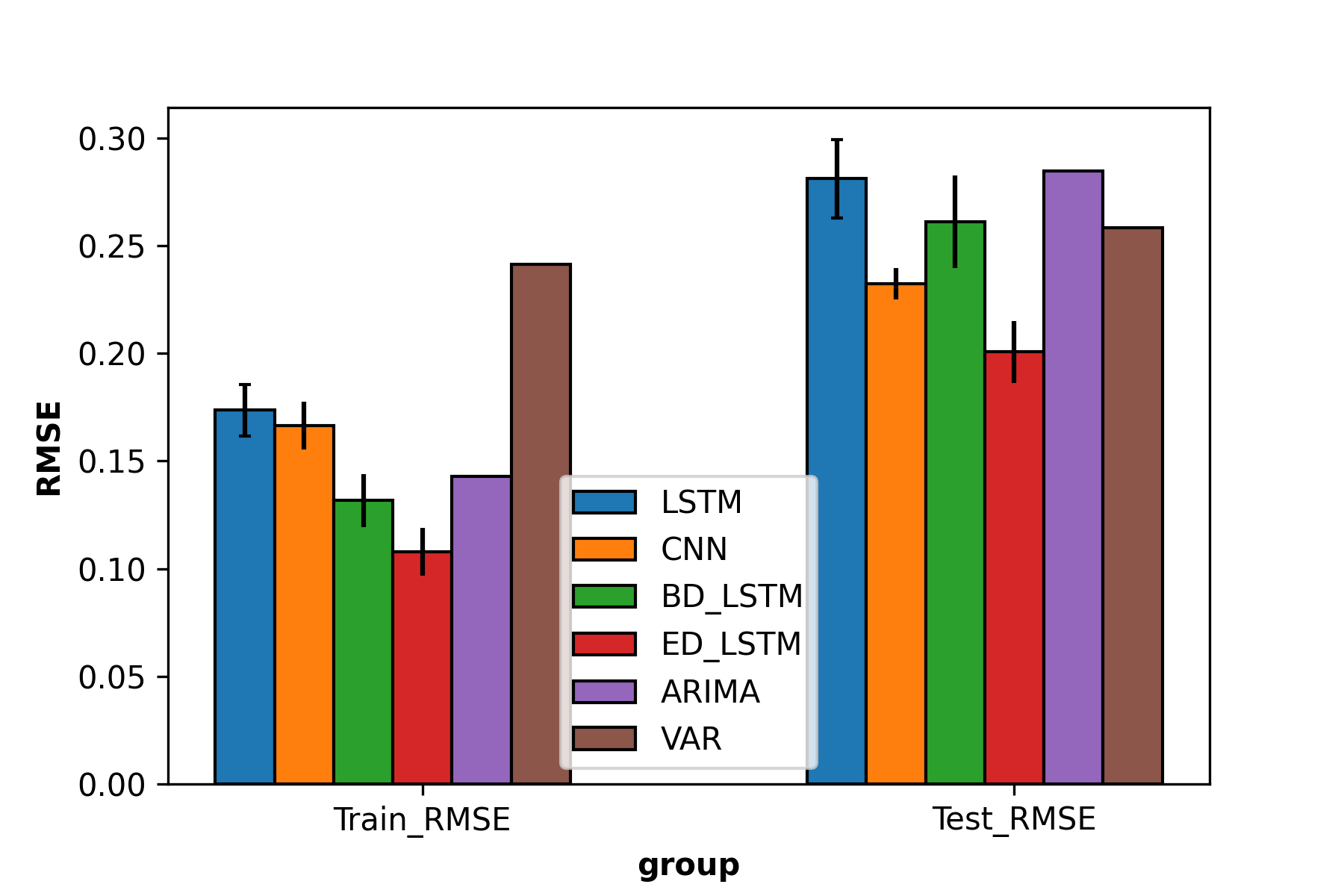}
         \caption{Mean RMSE across 3-steps of developed countries}
     \end{subfigure}
     \hfill
     \begin{subfigure}[h]{0.45\textwidth}
         \centering
         \includegraphics[width=\textwidth]{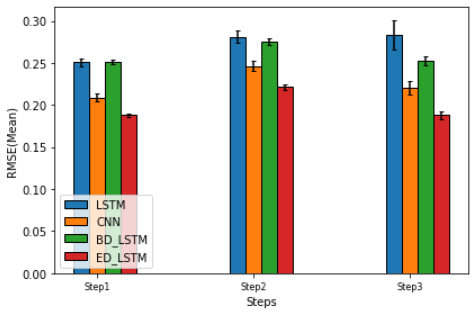}
         \caption{RMSE of each prediction horizon}
     \end{subfigure}
        \caption{Developed countries: performance evaluation of respective methods (RMSE mean and 95$\%$ confidence intervals as error bar).}
        \label{fig:developed_country}
\end{figure}

Figure \ref{fig:developed_country} provides a summary of the prediction results of developed countries. We observe that CNN and ED-LSTM are far better than other models for the test dataset. BD-LSTM and two traditional models have similar results, while LSTM's results are obviously worse than other models. In each step, the deep learning models have very different trends in pairs, as the number of steps increases, CNN and ED-LSTM perform better and better, proving that CNN and ED-LSTM have the ability to achieve long-term prediction, while LSTM and BD-LSTM are completely opposite, and the increase in the number of steps directly leads to an increase in RMSE, indicating that their predictive ability for multistep-ahead prediction is relatively weak. 

\subsubsection{Results: Developing Countries}
 
This section presents the forecast results for developing countries that are typically defined by countries that are still exploring the path of national development \cite{hudson2003international}, which at times result in irregularities in their economic cycles \cite{bussmann2006trade}. At times they have no stable trends, a high probability of abnormal rises or falls, and the transparency and credibility of their data are often questioned, making forecasting more difficult \cite{feng2001political}. In addition, the lack of historical data for Russia due to the collapse of the Soviet Union in 1991 has undoubtedly added to the difficulty of model fitting and forecasting.

 \begin{figure}[htbp!]
     \centering
     \begin{subfigure}[h]{0.45\textwidth}
         \centering
         \includegraphics[width=\textwidth]{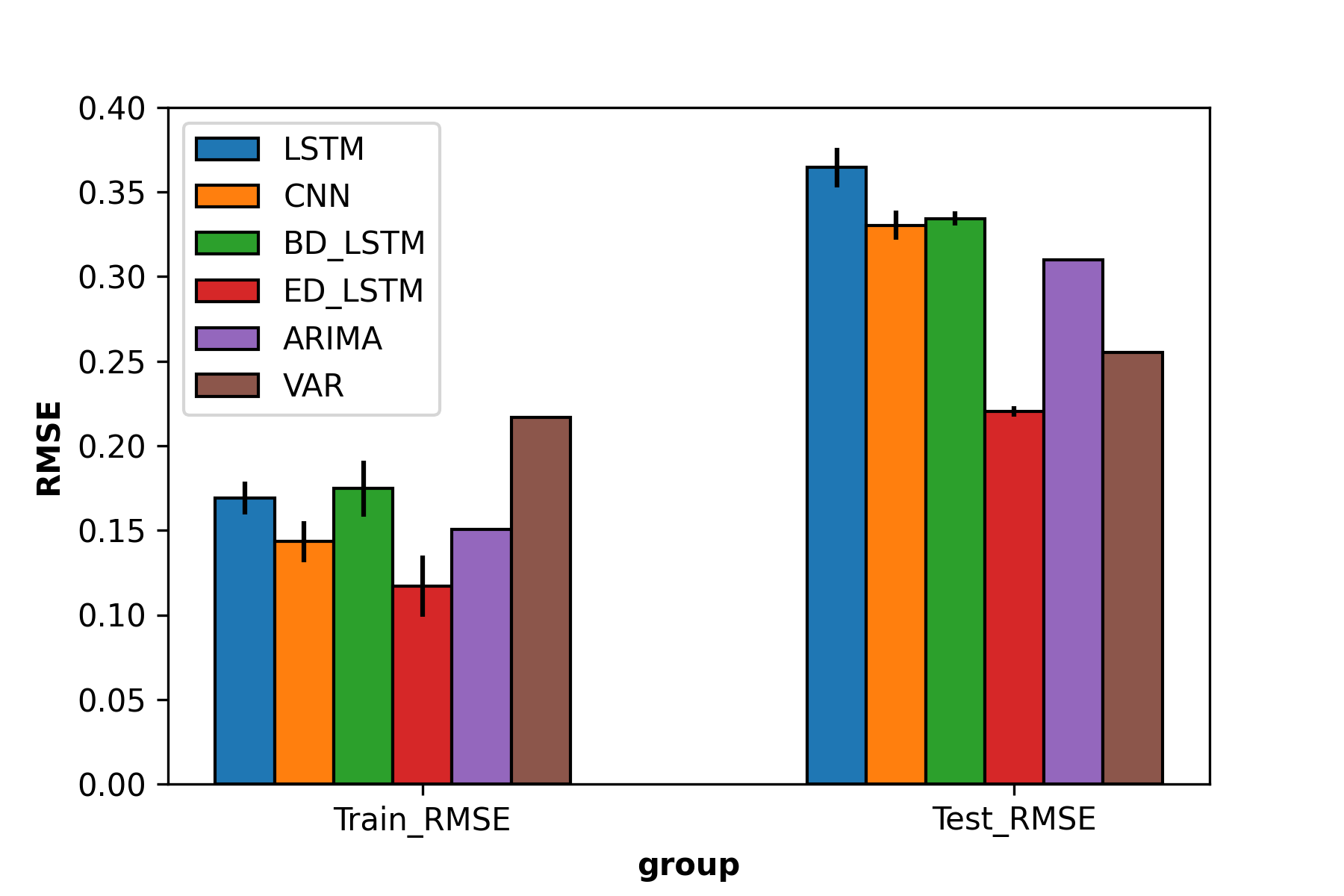}
         \caption{RMSE mean across 3-step prediction horizon}
     \end{subfigure}
     \hfill
     \begin{subfigure}[h]{0.45\textwidth}
         \centering
         \includegraphics[width=\textwidth]{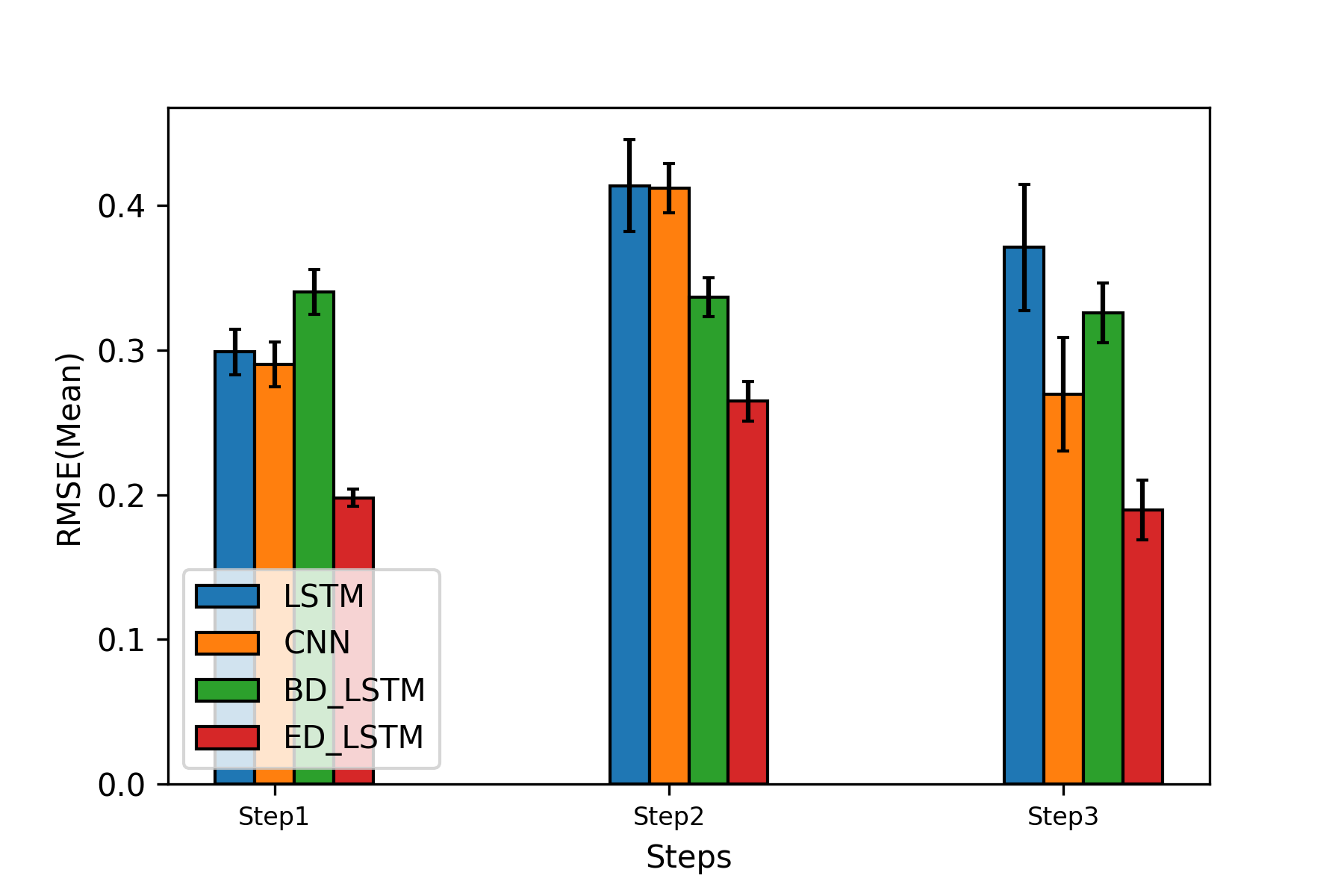}
         \caption{RMSE of each prediction horizon}
     \end{subfigure}
        \caption{Brazil: performance evaluation of respective methods (RMSE mean and 95$\%$ confidence intervals as error bar)}
        \label{fig:Brazil_RMSE}
\end{figure}

\begin{table*}
    \caption{Brazil: Prediction accuracy showing RMSE mean and 95 $\%$ confidence interval (±).}
    \label{tab:Brazil_table}
\begin{tabular}{c c c c c c c}
\hline
      & ARIMA & VAR & LSTM & BD-LSTM & ED-LSTM & CNN  \\ 
\hline
\hline
Train &0.1507  & 0.2169 &0.1689$\pm$0.0064 &0.1747$\pm$0.0055 & 0.1169$\pm$0.0061 & 0.1434$\pm$0.0.004\\ 
Test &0.3097  & 0.2549&0.3645$\pm$0.082 &0.3344$\pm$0.0014 & 0.2203$\pm$0.0011 & 0.3303$\pm$0.0028\\ 
\hline
Step1 &  &  &0.2988$\pm$0.005 &0.3402$\pm$0.005 & 0.1979$\pm$0.005 & 0.2901$\pm$0.002\\ 
Step2 &  &   &0.4137$\pm$0.0106 &0.3367$\pm$0.004 & 0.2647$\pm$0.004 & 0.412$\pm$0.006\\ 
Step3 &  &  &0.371$\pm$0.0142 &0.3259$\pm$0.006 & 0.1894$\pm$0.006 & 0.2694$\pm$0.0131\\ 
\hline
\hline
\end{tabular}
\end{table*}

We present the results for Brazil in Figure \ref{fig:Brazil_RMSE} and Table \ref{tab:Brazil_table}.  We observe that only ED-LSTM is better than the traditional time series   models for the test dataset, and LSTM remains the worst-performing deep learning model. Examining the step-by-step predictions, we see that the RMSE of ED-LSTM deteriorates in the second step, but remains the best model for every step.
 
 \begin{figure}[htbp!]
     \centering
     \begin{subfigure}[h]{0.45\textwidth}
         \centering
         \includegraphics[width=\textwidth]{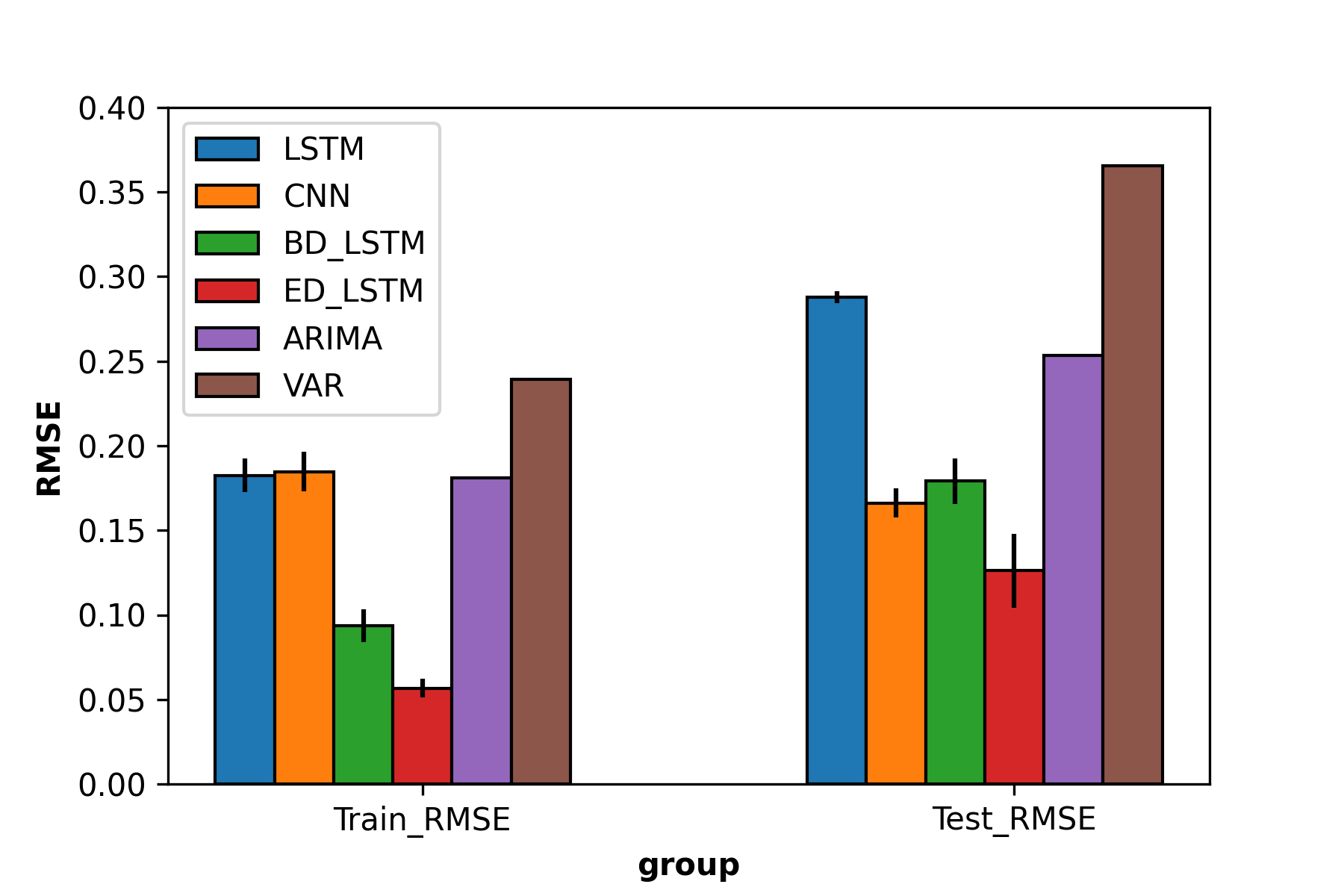}
         \caption{RMSE mean across 3-step prediction horizon}
     \end{subfigure}
     \hfill
     \begin{subfigure}[h]{0.45\textwidth}
         \centering
         \includegraphics[width=\textwidth]{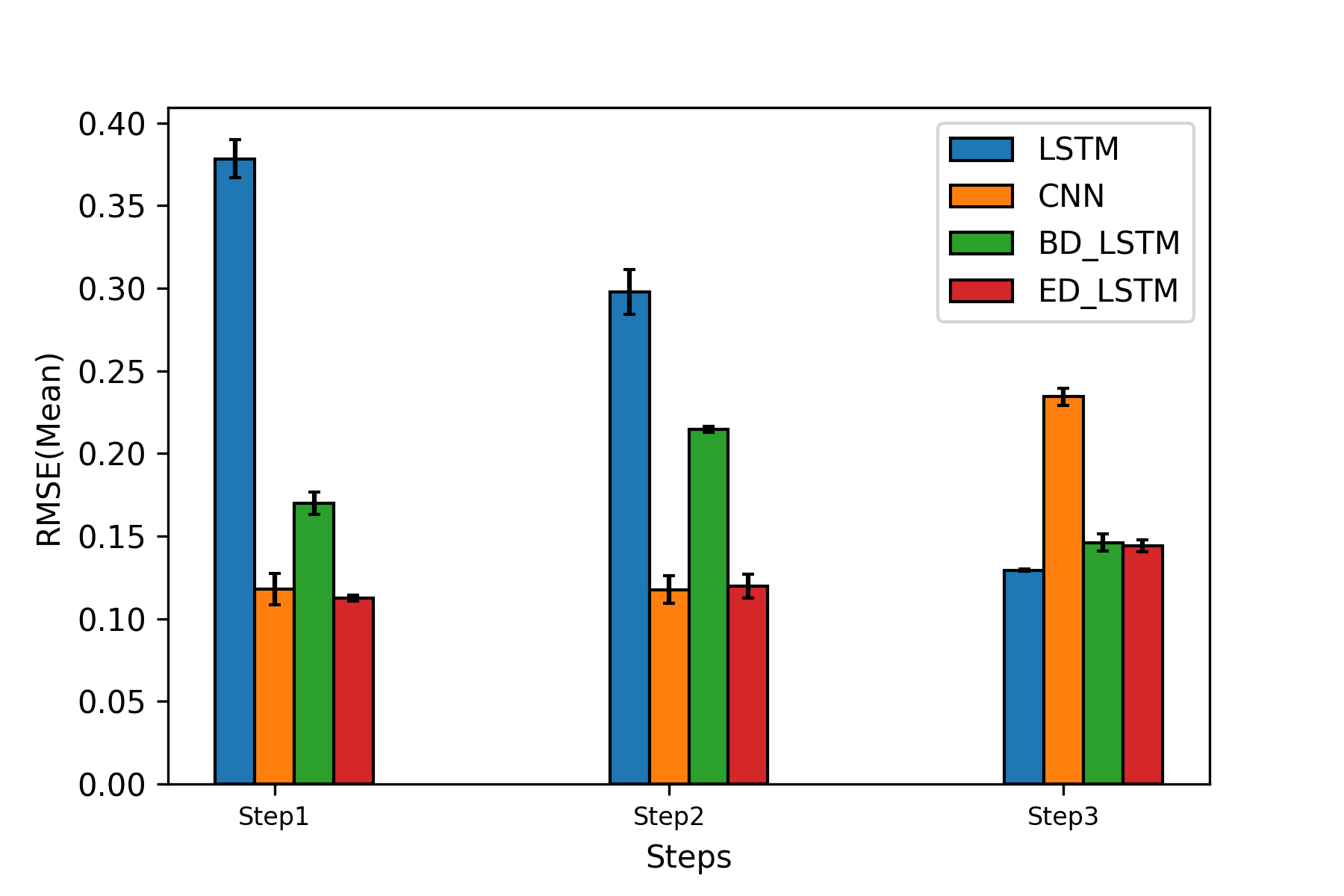}
         \caption{RMSE of each prediction horizon}
     \end{subfigure}
        \caption{China: performance evaluation of respective methods (RMSE mean and 95$\%$ confidence intervals as error bar).}
        \label{fig:China_RMSE}
\end{figure}

\begin{table*}
    \caption{China: Prediction accuracy showing RMSE mean and 95 $\%$ confidence interval (±).}
    \label{tab:China_table}
\begin{tabular}{c c c c c c c}
\hline
      & ARIMA & VAR & LSTM & BD-LSTM & ED-LSTM & CNN  \\ 
\hline
\hline
Train &0.1808 & 0.2394 &0.1825$\pm$0.0066 &0.0937$\pm$0.0032 & 0.0567$\pm$0.0018 & 0.1846$\pm$0.0.004\\ 
Test &0.2536  & 0.3656&0.2879$\pm$0.0024 &0.1792$\pm$0.0045 & 0.1262$\pm$0.0073 & 0.1662$\pm$0.0028\\ 
\hline
Step1 &  &  &0.3783$\pm$0.038 &0.1701$\pm$0.0023 & 0.1123$\pm$0.006 & 0.118$\pm$0.0032\\ 
Step2 &  &   &0.2977$\pm$0.0045 &0.2146$\pm$0.0006 & 0.1198$\pm$0.0024 & 0.1177$\pm$0.0028\\ 
Step3 &  &  &0.1294$\pm$0.0003 &0.146$\pm$0.0017 & 0.1442$\pm$0.0012 & 0.2343$\pm$0.0054\\ 
\hline
\hline
\end{tabular}
\end{table*}

We present the results for China in Figure \ref{fig:China_RMSE} and Table \ref{tab:China_table},  where we observe that except for LSTM, the other three deep learning models are significantly better than the traditional time series models. The RMSE  of LSTM is only slightly worse than ARIMA, and   also much better than VAR for the test dataset. Among the top three deep learning models, ED-LSTM ranks first, but the confidence interval (error bar)  remains stable when compared with CNN and BD-LSTM.

 \begin{figure}[htbp!]
     \centering
     \begin{subfigure}[h]{0.45\textwidth}
         \centering
         \includegraphics[width=\textwidth]{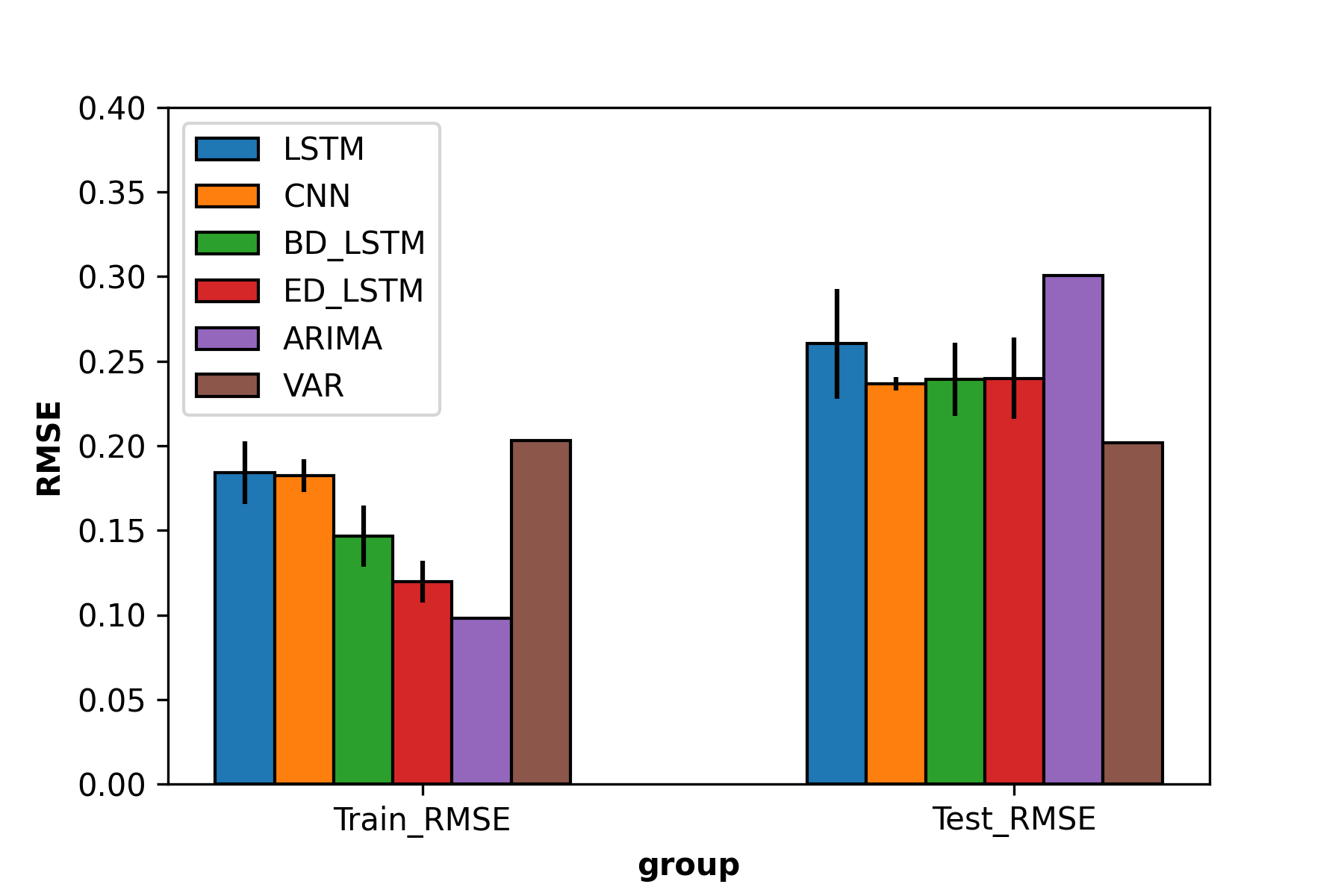}
         \caption{RMSE mean across 3-step prediction horizon}
     \end{subfigure}
     \hfill
     \begin{subfigure}[h]{0.45\textwidth}
         \centering
         \includegraphics[width=\textwidth]{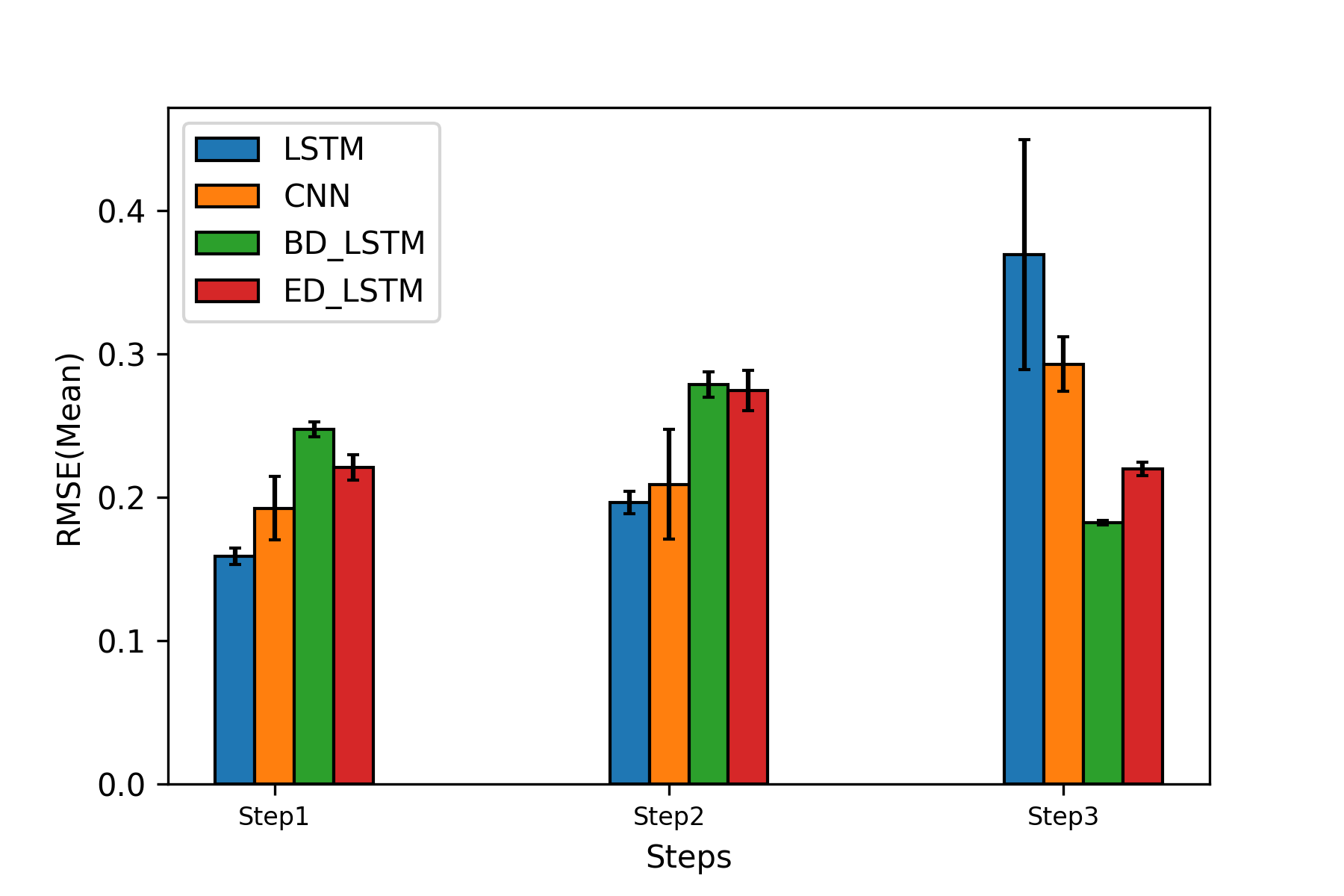}
         \caption{RMSE of each prediction horizon}
     \end{subfigure} 
        \caption{India: performance evaluation of respective methods (RMSE mean and 95$\%$ confidence intervals as error bar).}
        \label{fig:India_RMSE}
\end{figure}

 \begin{figure}[htbp!]
     \centering
     
     \begin{subfigure}[h]{0.45\textwidth}
         \centering
         \includegraphics[width=\textwidth]{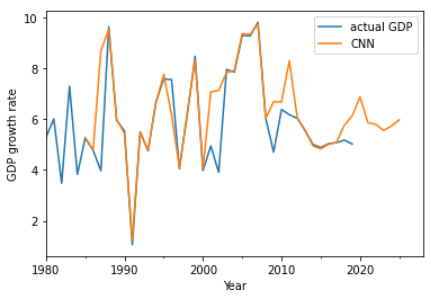}
         \caption{CNN model}
     \end{subfigure}
     \hfill
     \begin{subfigure}[h]{0.45\textwidth}
         \centering
         \includegraphics[width=\textwidth]{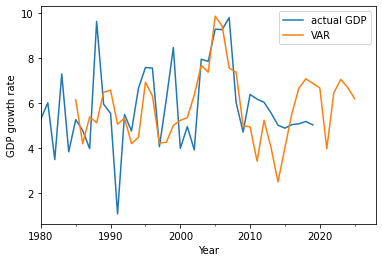}
         \caption{VAR model}
     \end{subfigure}
        \caption{India: prediction plots of selected models comparing with actual GDP growth rate.}
        \label{fig:India_RMSEplot}
\end{figure}

\begin{table*}
    \caption{India: Prediction accuracy showing RMSE mean and 95 $\%$ confidence interval (±).}
    \label{tab:India_table}
\begin{tabular}{c c c c c c c}
\hline
      & ARIMA & VAR & LSTM & BD-LSTM & ED-LSTM & CNN  \\ 
\hline
\hline
Train &0.0982 & 0.2032 &0.1842$\pm$0.0124 &0.1465$\pm$0.006 & 0.1197$\pm$0.0041 & 0.1823$\pm$0.0.0033\\ 
Test &0.3008  & 0.2019&0.2604$\pm$0.02164 &0.2394$\pm$0.0072 & 0.2398$\pm$0.0081 & 0.2366$\pm$0.0013\\ 
\hline
Step1 &  &  &0.1587$\pm$0.019 &0.2475$\pm$0.0017 & 0.2206$\pm$0.0033 & 0.1923$\pm$0.007\\ 
Step2 &  &   &0.1961$\pm$0.0023 &0.2784$\pm$0.003 & 0.2745$\pm$0.0047 & 0.209$\pm$0.0128\\ 
Step3 &  &  &0.369$\pm$0.027 &0.1821$\pm$0.0005 & 0.2197$\pm$0.0015 & 0.2927$\pm$0.0063\\ 
\hline
\hline
\end{tabular}
\end{table*}

We present the results for India  in Figure \ref{fig:India_RMSE} and Table \ref{tab:India_table},  where we observe that the mean RMSE of all the deep learning models are in a similar range for the test dataset, outperforming the ARIMA but worse than the VAR.   Figure \ref{fig:India_RMSEplot} presents the prediction plots of the best models and compares them with the actual GDP growth rate.

 \begin{figure}[htbp!]
     \centering
     \begin{subfigure}[h]{0.45\textwidth}
         \centering
         \includegraphics[width=\textwidth]{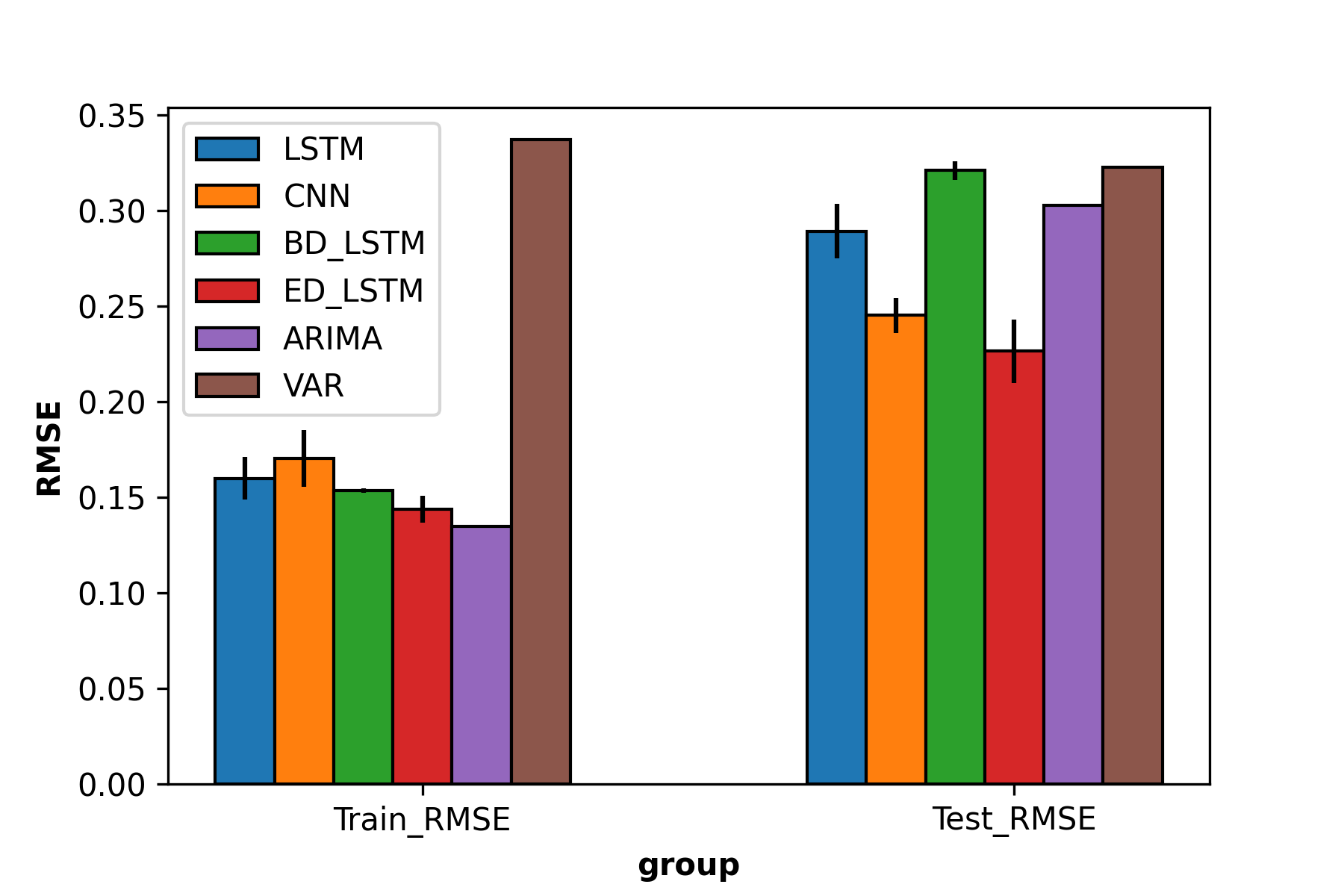}
         \caption{RMSE mean across 3-step prediction horizon}
     \end{subfigure}
     \hfill
     \begin{subfigure}[h]{0.45\textwidth}
         \centering
         \includegraphics[width=\textwidth]{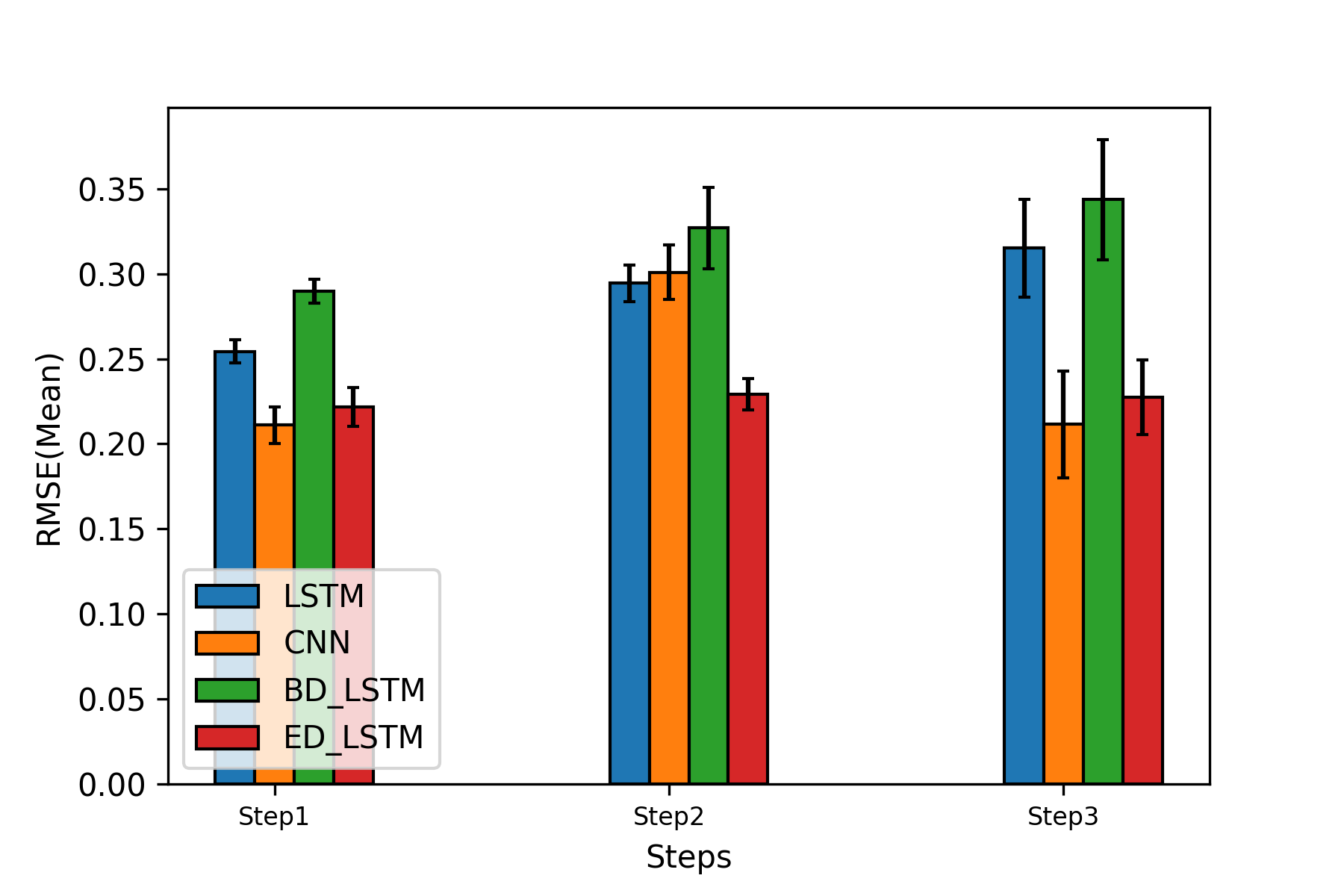}
         \caption{RMSE of each prediction horizon}
     \end{subfigure}
        \caption{Russia: performance evaluation of respective methods (RMSE mean and 95$\%$ confidence intervals as error bar).}
        \label{fig:Russia_RMSE}
\end{figure}

\begin{table*}
    \caption{Russia:  Prediction accuracy showing RMSE mean and 95 $\%$ confidence interval (±).}
    \label{tab:Russia_table}
\begin{tabular}{c c c c c c c}
\hline
      & ARIMA & VAR & LSTM & BD-LSTM & ED-LSTM & CNN  \\ 
\hline
\hline
Train &0.1348 & 0.3371 &0.1599$\pm$0.0072 &0.1535$\pm$0.0004 & 0.1437$\pm$0.002 & 0.1704$\pm$0.0.0049\\ 
Test &0.3029  & 0.3227&0.2892$\pm$0.0095 &0.3211$\pm$0.0016 & 0.2265$\pm$0.0055 & 0.2453$\pm$0.0031\\ 
\hline
Step1 &  &  &0.2543$\pm$0.0023 &0.2896$\pm$0.002& 0.2218$\pm$0.0038 & 0.211$\pm$0.0036\\ 
Step2 &  &   &0.2945$\pm$0.0036 &0.3269$\pm$0.079 & 0.2293$\pm$0.0032 & 0.3009$\pm$0.0053\\ 
Step3 &  &  &0.315$\pm$0.0095 &0.3436$\pm$0.0118 & 0.2274$\pm$0.0073 & 0.2115$\pm$0.0105\\ 
\hline
\hline
\end{tabular}
\end{table*}

We present the results for Russia in Figure \ref{fig:Russia_RMSE} and Table \ref{tab:Russia_table}, where the CNN and ED-LSTM models outperform all others for the test \ref{tab:Russia_table}, with the ED-LSTM exhibiting a slight advantage.  

 \begin{figure}[htbp!]
     \centering
     \begin{subfigure}[h]{0.45\textwidth}
         \centering
         \includegraphics[width=\textwidth]{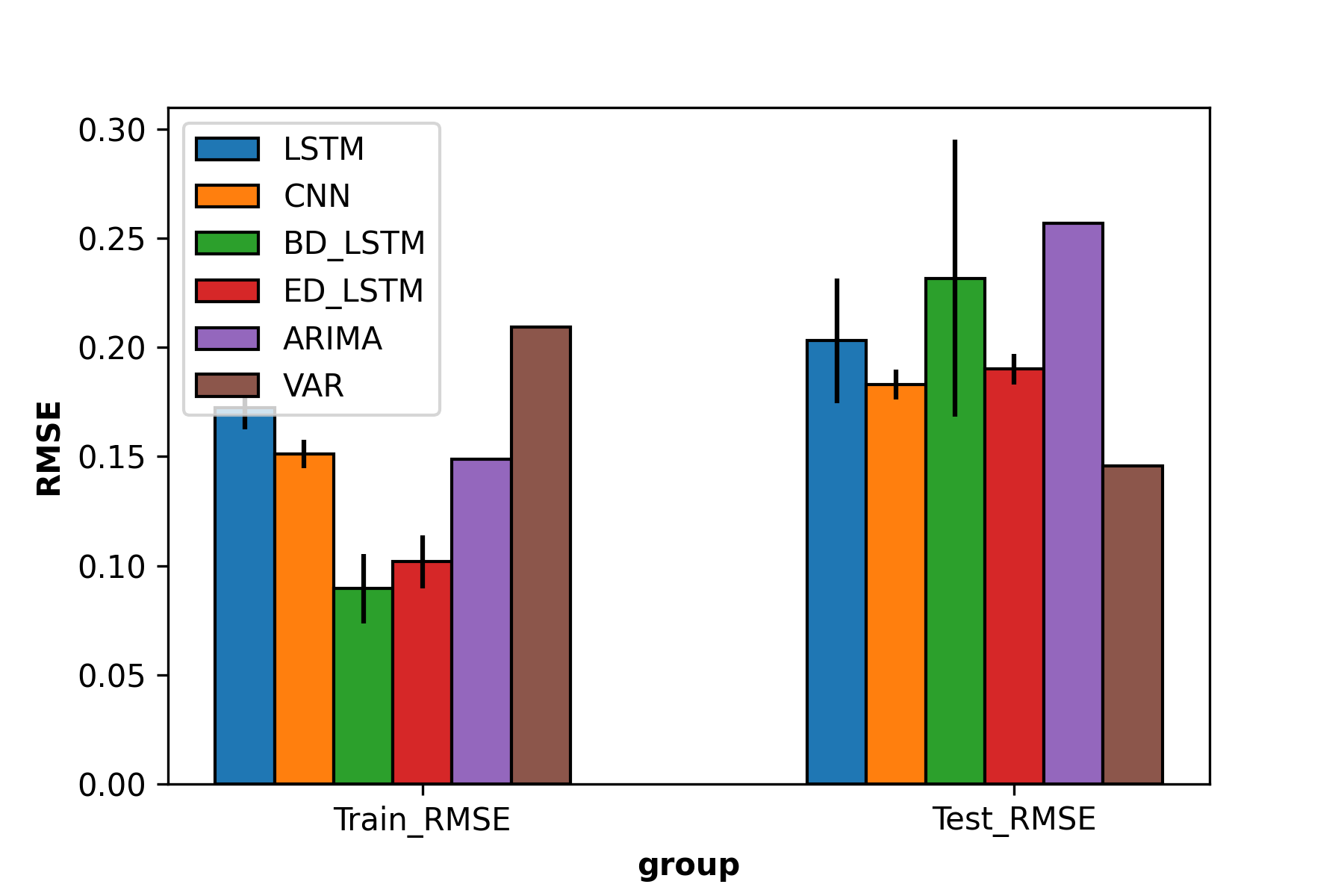}
         \caption{RMSE mean across 3-step prediction horizon}
     \end{subfigure}
     \hfill
     \begin{subfigure}[h]{0.45\textwidth}
         \centering
         \includegraphics[width=\textwidth]{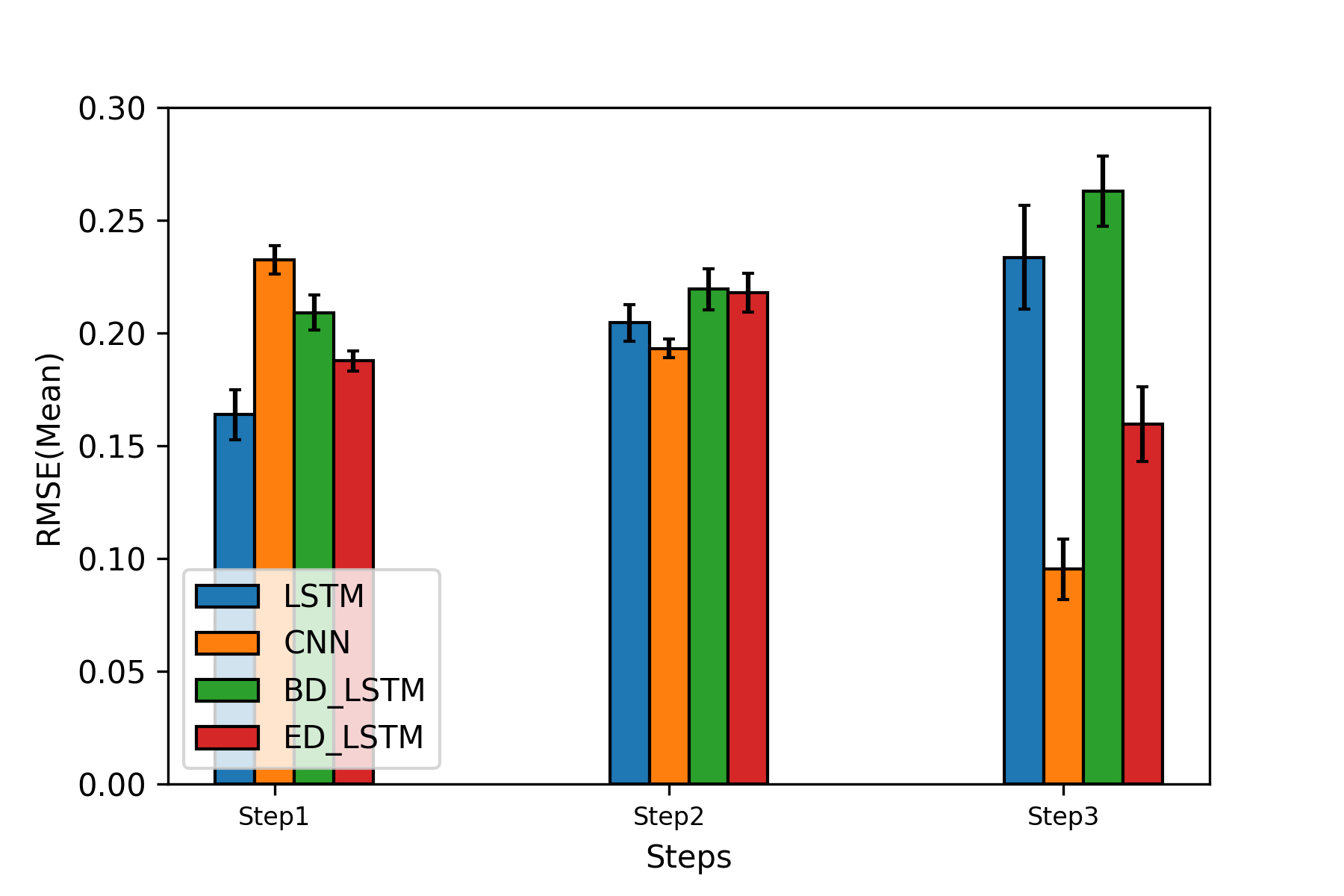}
         \caption{RMSE of each prediction horizon}
     \end{subfigure} 
        \caption{South Africa: performance evaluation of respective methods (RMSE mean and 95$\%$ confidence interval as error bar)}
        \label{fig:South_Africa_RMSE}
\end{figure}

\begin{table*}
    \caption{South Africa: Prediction accuracy showing RMSE mean and 95 $\%$ confidence interval (±).}
    \label{tab:South_Africa_table}
\begin{tabular}{c c c c c c c}
\hline
      & ARIMA & VAR & LSTM & BD-LSTM & ED-LSTM & CNN  \\ 
\hline
\hline
Train &0.1487 & 0.2093 &0.1725$\pm$0.0068 &0.0895$\pm$0.0053 & 0.1018$\pm$0.0041 & 0.1512$\pm$0.0.0022\\ 
Test &0.2569 & 0.1456&0.2031$\pm$0.0173 &0.2317$\pm$0.0211 & 0.1901$\pm$0.002 & 0.1831$\pm$0.0023\\ 
\hline
Step1 &  &  &0.1638$\pm$0.0037 &0.209$\pm$0.0023 & 0.1876$\pm$0.0015 & 0.2324$\pm$0.021\\ 
Step2 &  &   &0.2046$\pm$0.0027 &0.2195$\pm$0.003 & 0.2179$\pm$0.0029 & 0.1932$\pm$0.0014\\ 
Step3 &  &  &0.2335$\pm$0.07 &0.263$\pm$0.005 & 0.1595$\pm$0.0052 & 0.0952$\pm$0.0045\\ 
\hline
\hline
\end{tabular}
\end{table*}

We further present the results for South Africa in Figure \ref{fig:South_Africa_RMSE} and Table \ref{tab:South_Africa_table},  where we find a  similar trend when compared to the results for India. The ARIMA model, by contrast, ranks last and in the case of the  deep learning models, we find that CNN and ED-LSTM maintain consistently good performance in the test dataset, ranking first and second for the deep learning models.  

 \begin{figure}[htbp!]
     \centering
     \begin{subfigure}[h]{0.45\textwidth}
         \centering
         \includegraphics[width=\textwidth]{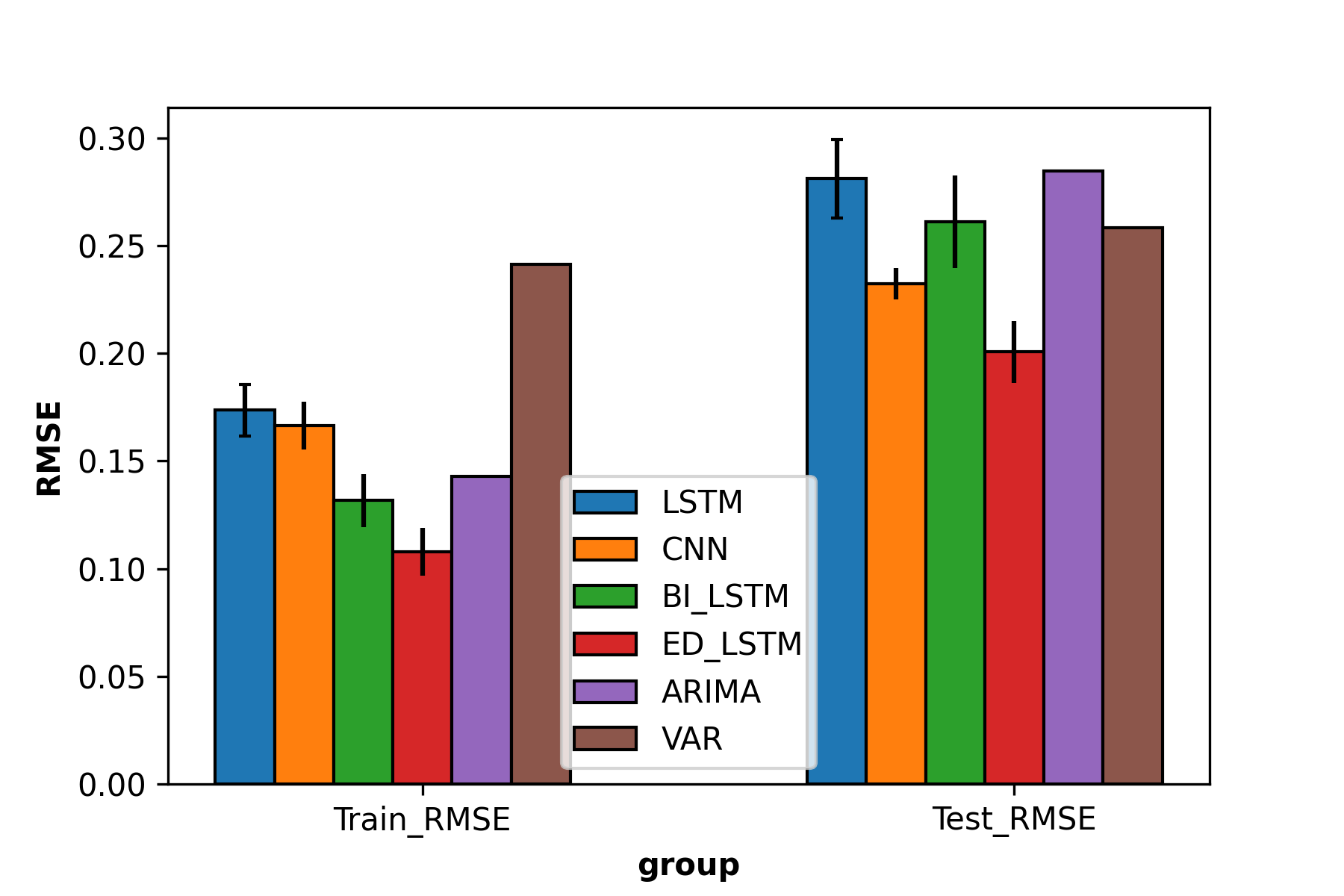}
         \caption{Mean RMSE across 3-steps for developing countries}
     \end{subfigure}
     \hfill
     \begin{subfigure}[h]{0.45\textwidth}
         \centering
         \includegraphics[width=\textwidth]{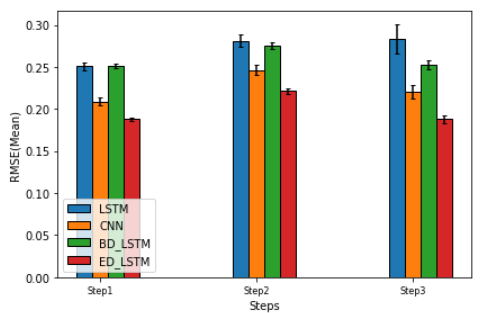}
         \caption{RMSE of each prediction horizon}
     \end{subfigure}
        \caption{Developing countries: performance evaluation of respective methods (RMSE mean and 95$\%$ confidence intervals as error bar).}
        \label{fig:developing_country}
\end{figure}

As in the previous section, Figure \ref{fig:developing_country} shows that CNN and ED-LSTM provide better overall performance than traditional time series models for the test dataset. Looking at the individual prediction horizon  (time step), ED-LSTM and CNN significantly outperform the other two models at every step, and the former is the first rank in every step. It is worth noting that for India and South Africa, the VAR performance is best according to the mean RMSE, but from the prediction curves, the VAR performance is not as good as the optimal deep learning models. 
 
\subsection{Recursive deep learning framework: results} 

The results in general show that both for the case of developed or developing countries (Table \ref{tab:rk_ark}), ED-LSTM ranks first; hence, we use ED-LSTM to predict the next decade's GDP growth rate. We use the PWT data and extend it recursively to forecast decadal growth rate. We begin the recursive strategy by forecasting features in the dataset until the features are long enough for us to forecast the GDP growth rate for the next decade.  

\begin{table*}[htbp!] 
    \caption{RMSE for feature prediction in recursive strategy in some countries (CSH represents  the share in GDP per-capita, where C represents household consumption, I refers to  the gross capital formation and G is government consumption).}
    \label{recursive:RMSE}
\centering 
\begin{tabular}{ l l  c c } 
\hline
Country & Feature & Train-RMSE& Test-RMSE  \\ 
\hline
\hline
Australia & Net population growth & 0.0432 & 0.0769 \\ 
          & CPI & 0.0528 & 0.1128 \\ 
          & Employment rate & 0.1595 & 0.1759 \\ 
Brazil & CSH-C & 0.1765 & 0.1254  \\ 
        & CSH-I & 0.2128 & 0.1728 \\ 
        & CSH-G & 0.1784 & 0.1292 \\ 

\hline
\hline
\end{tabular}
\end{table*}

\begin{figure}[htbp!]
     \centering
     \begin{subfigure}[h]{0.45\textwidth}
         \centering
         \includegraphics[width=\textwidth]{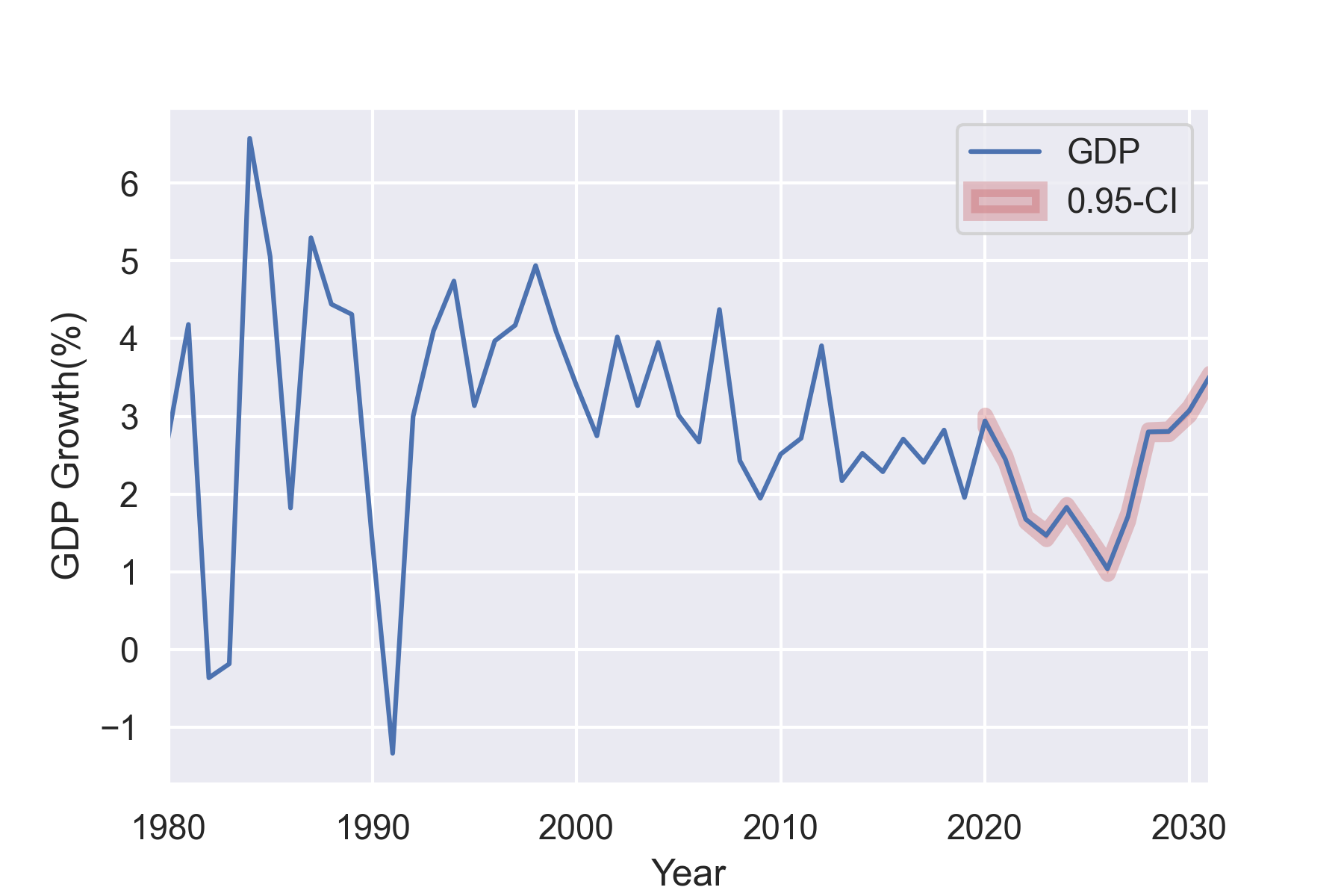}
         \caption{Australia GDP with the model prediction (mean) and uncertainty (95$\%$ CI). }
     \end{subfigure}
     \hfill
     \begin{subfigure}[h]{0.45\textwidth}
         \centering
         \includegraphics[width=\textwidth]{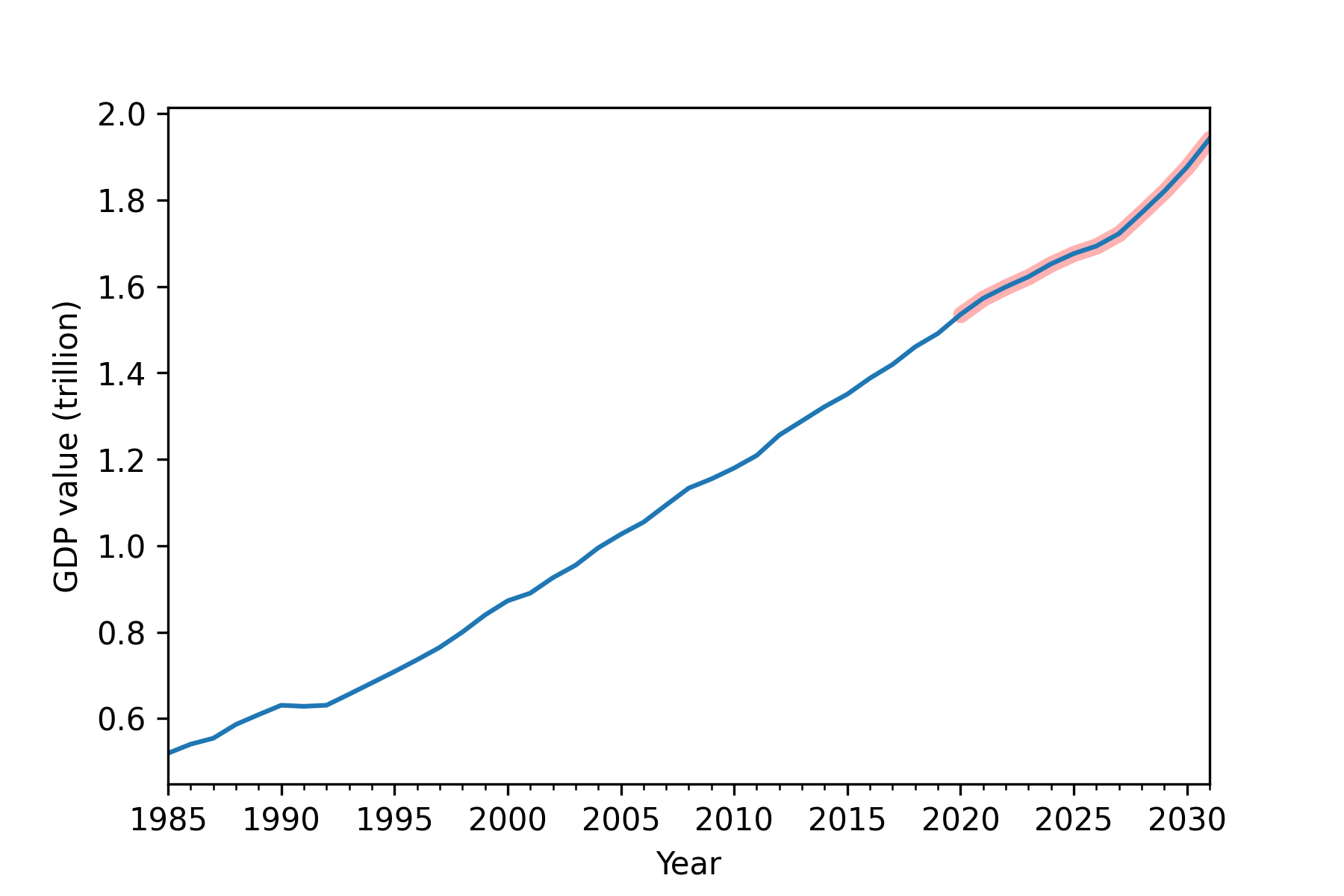}
         \caption{Australia GDP with the model prediction (mean) and uncertainty (95$\%$ CI).}
     \end{subfigure}
     \caption{Australia actual value and recursive multivariate ED-LSTM model prediction for next decade (2020-2031).}
        \label{fig:Australia_final_forecast}
\end{figure}

\begin{figure}[htbp!]
     \centering
     \begin{subfigure}[h]{0.45\textwidth}
         \centering
         \includegraphics[width=\textwidth]{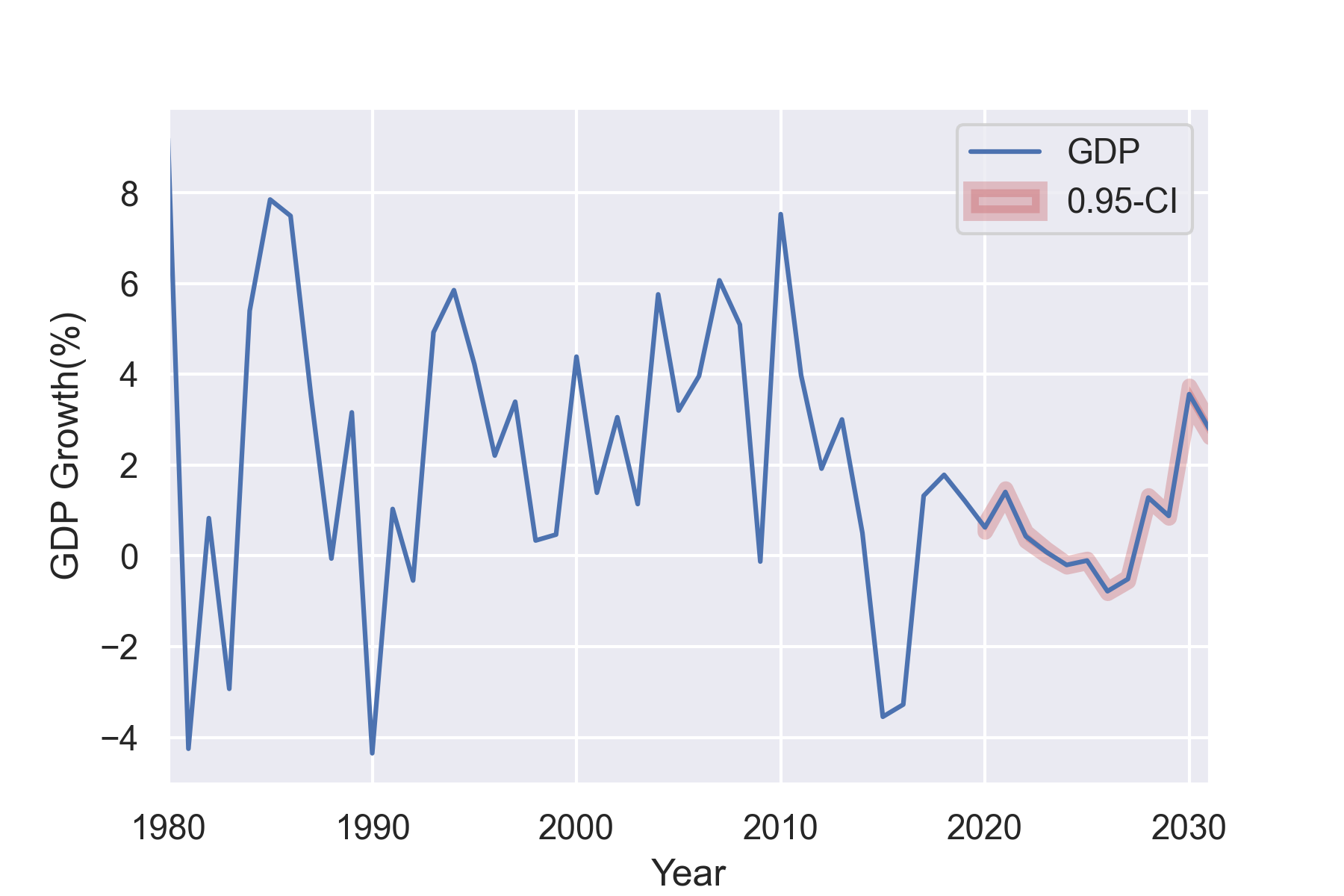}
         \caption{Brazil GDP with the model prediction (mean) and uncertainty (95$\%$ CI). }
     \end{subfigure}
     \hfill
     \begin{subfigure}[h]{0.45\textwidth}
         \centering
         \includegraphics[width=\textwidth]{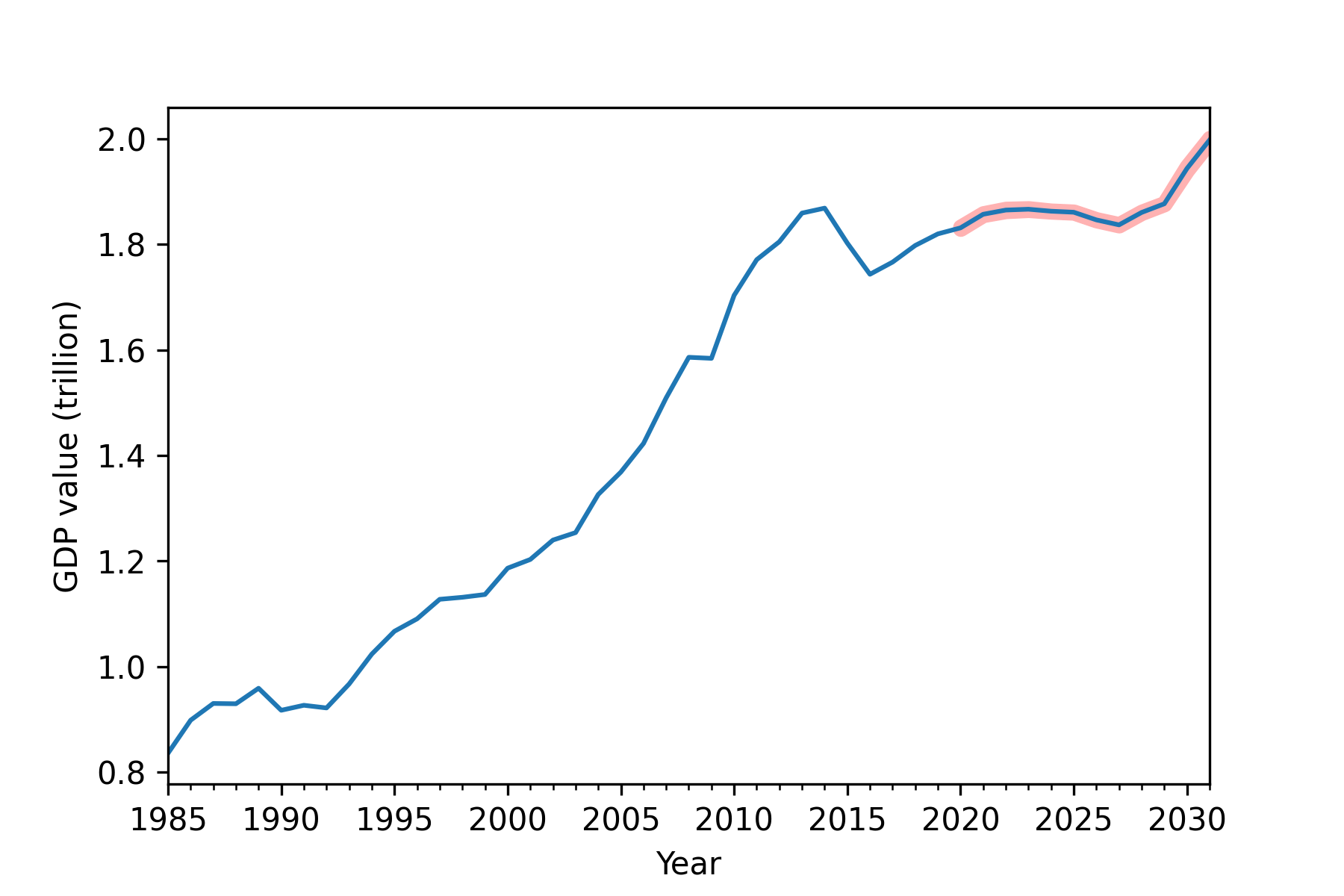}
         \caption{Brazil GDP with the model prediction (mean) and uncertainty (95$\%$ CI).}
     \end{subfigure}
     \caption{Brazil actual value and recursive multivariate ED-LSTM model prediction for next decade (2020-2031).}
        \label{fig:Brazil_final_forecast}
\end{figure}

\begin{figure}[htbp!]
     \centering
     \begin{subfigure}[h]{0.45\textwidth}
         \centering
         \includegraphics[width=\textwidth]{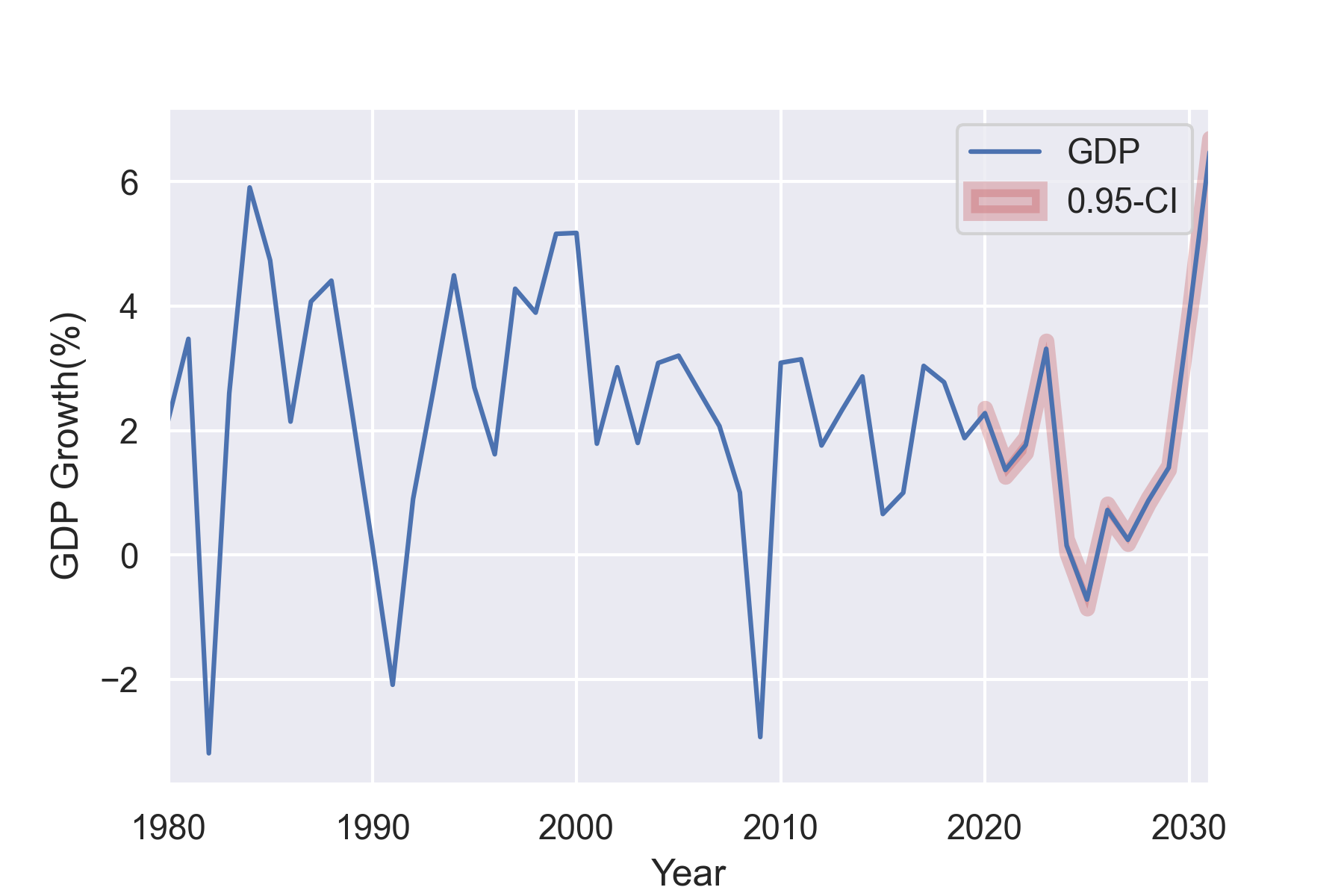}
         \caption{Canada GDP with the model prediction (mean) and uncertainty (95$\%$ CI). }
     \end{subfigure}
     \hfill
     \begin{subfigure}[h]{0.45\textwidth}
         \centering
         \includegraphics[width=\textwidth]{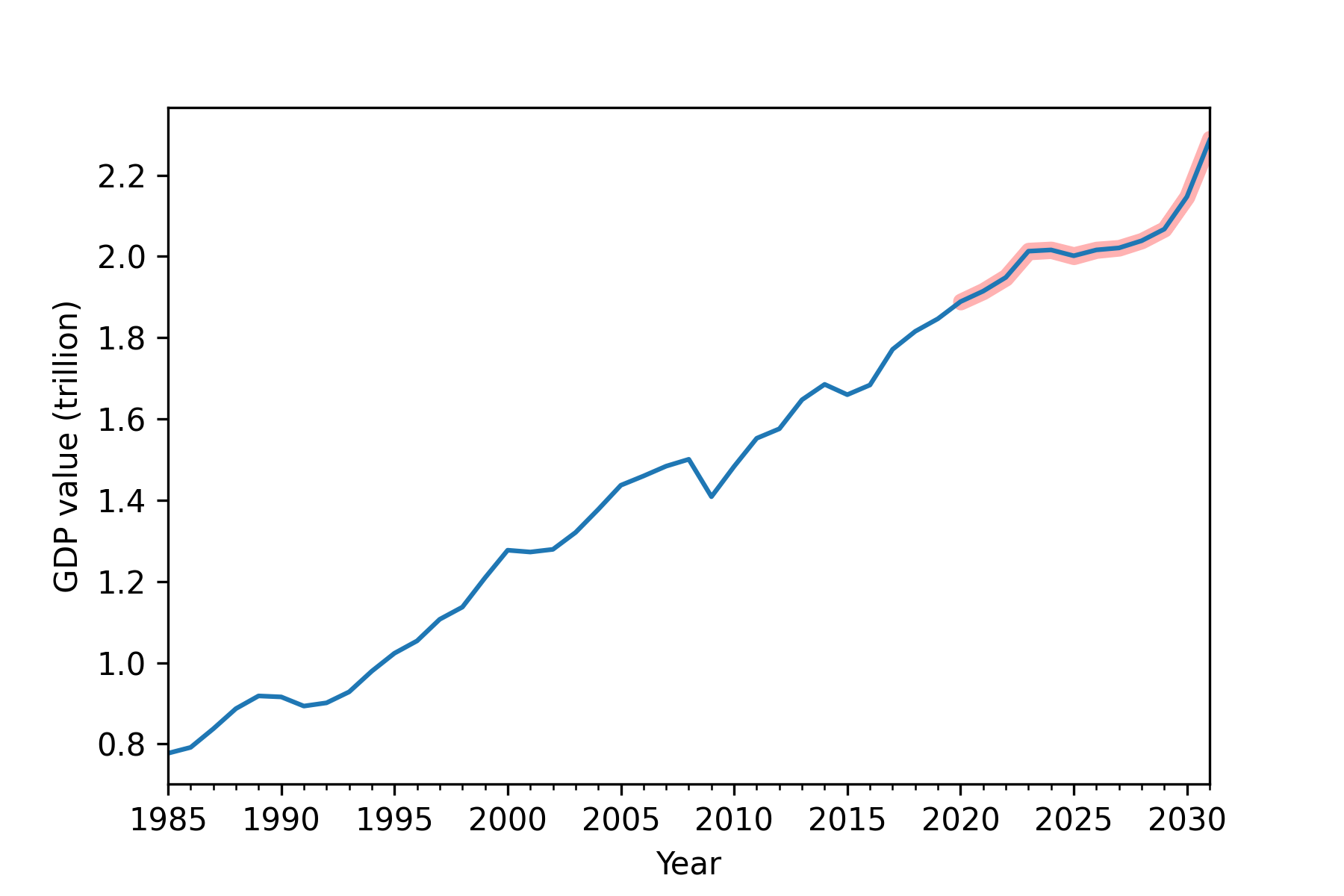}
         \caption{Canada GDP with the model prediction (mean) and uncertainty (95$\%$ CI).}
     \end{subfigure}
        \caption{Canada: actual value and recursive multivariate ED-LSTM model prediction for next decade (2020-2031).}
        \label{fig:Canada_final_forecast}
\end{figure}

\begin{figure}[htbp!]
     \centering
     \begin{subfigure}[h]{0.45\textwidth}
         \centering
         \includegraphics[width=\textwidth]{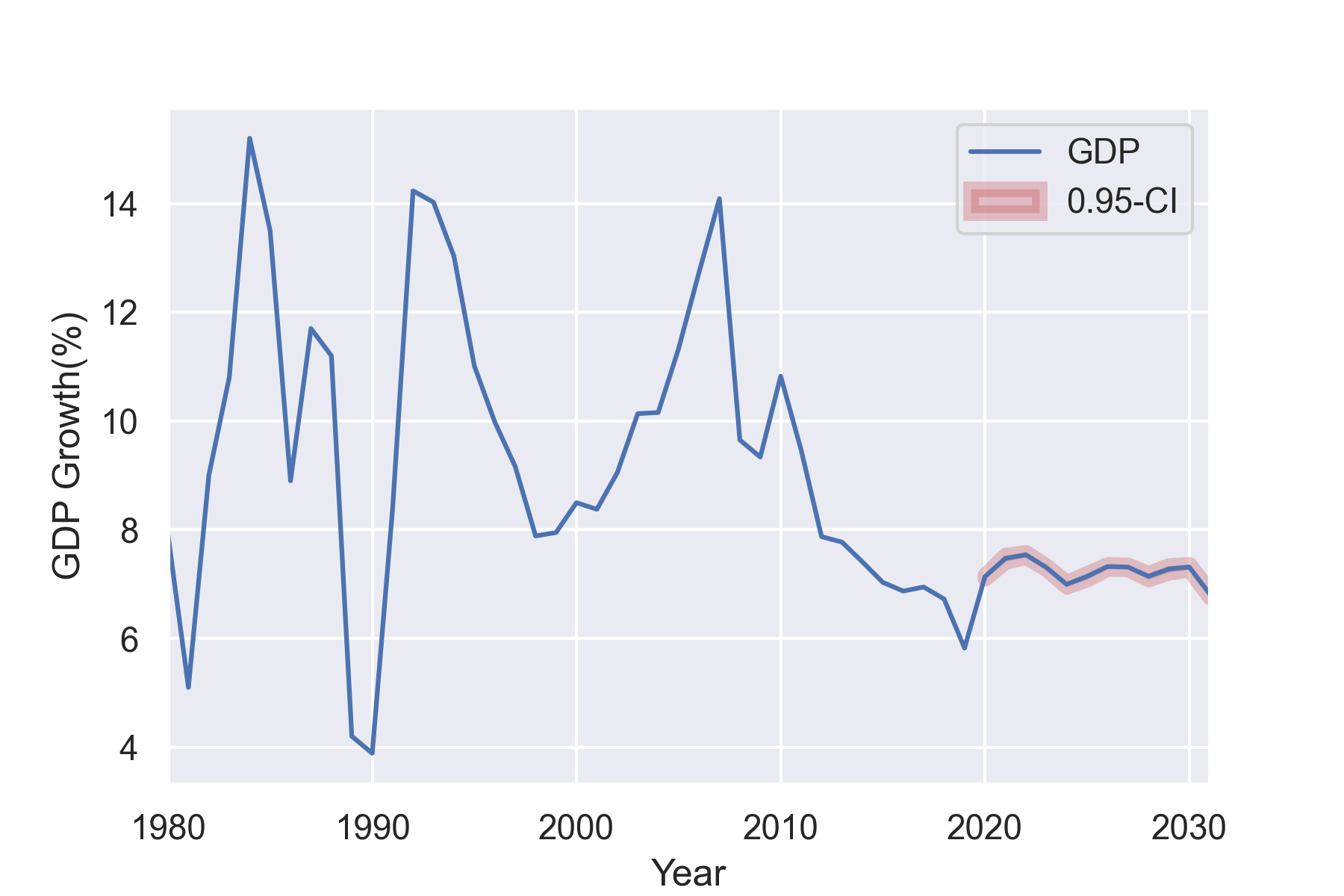}
         \caption{China GDP with the model prediction (mean) and uncertainty (95$\%$ CI). }
     \end{subfigure}
     \hfill
     \begin{subfigure}[h]{0.45\textwidth}
         \centering
         \includegraphics[width=\textwidth]{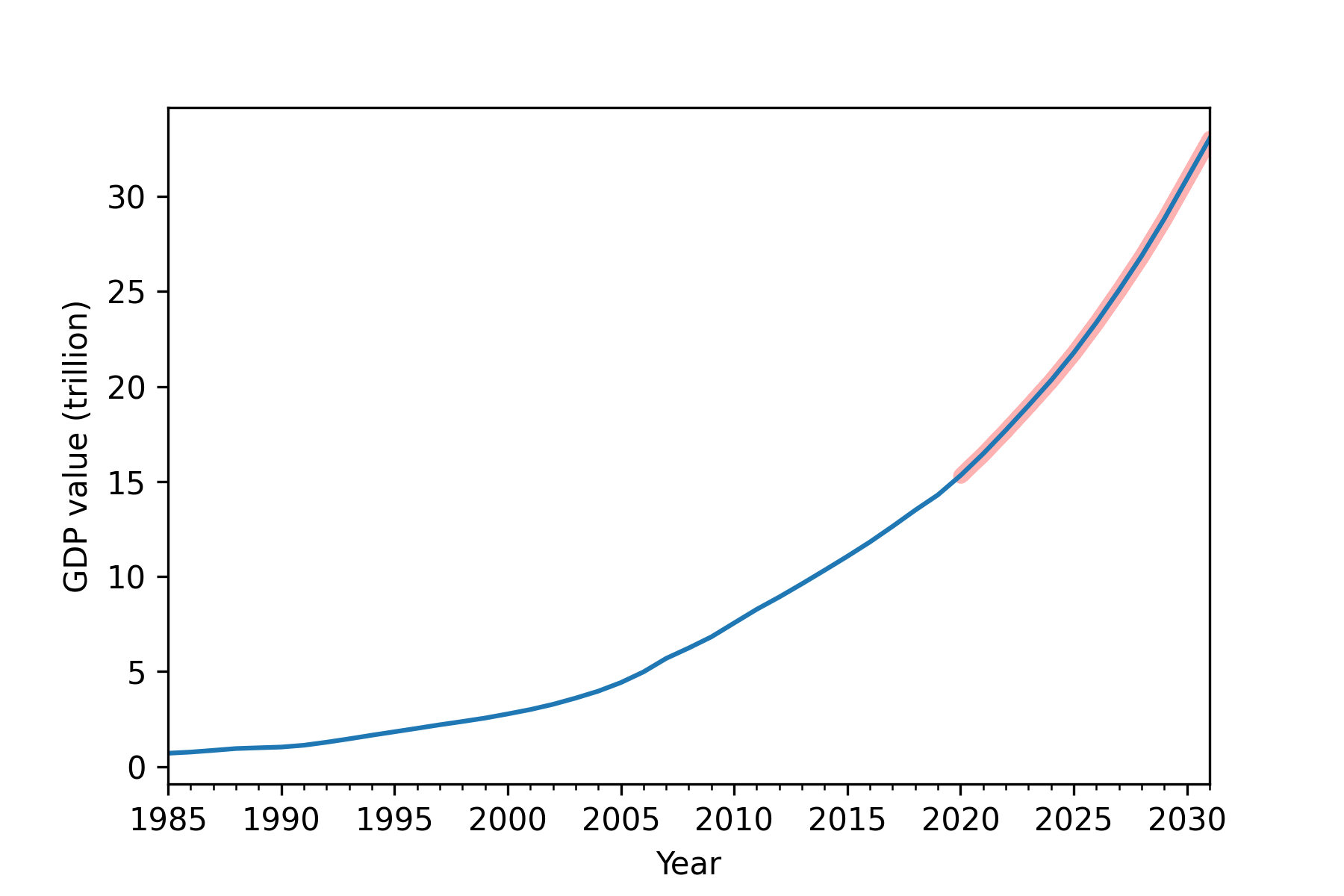}
         \caption{China GDP with the model prediction (mean) and uncertainty (95$\%$ CI).}
     \end{subfigure}
        \caption{China: actual value and recursive multivariate ED-LSTM model prediction for next decade (2020-2031).}
        \label{fig:China_final_forecast}
\end{figure}

\begin{figure}[htbp!]
     \centering
     \begin{subfigure}[h]{0.45\textwidth}
         \centering
         \includegraphics[width=\textwidth]{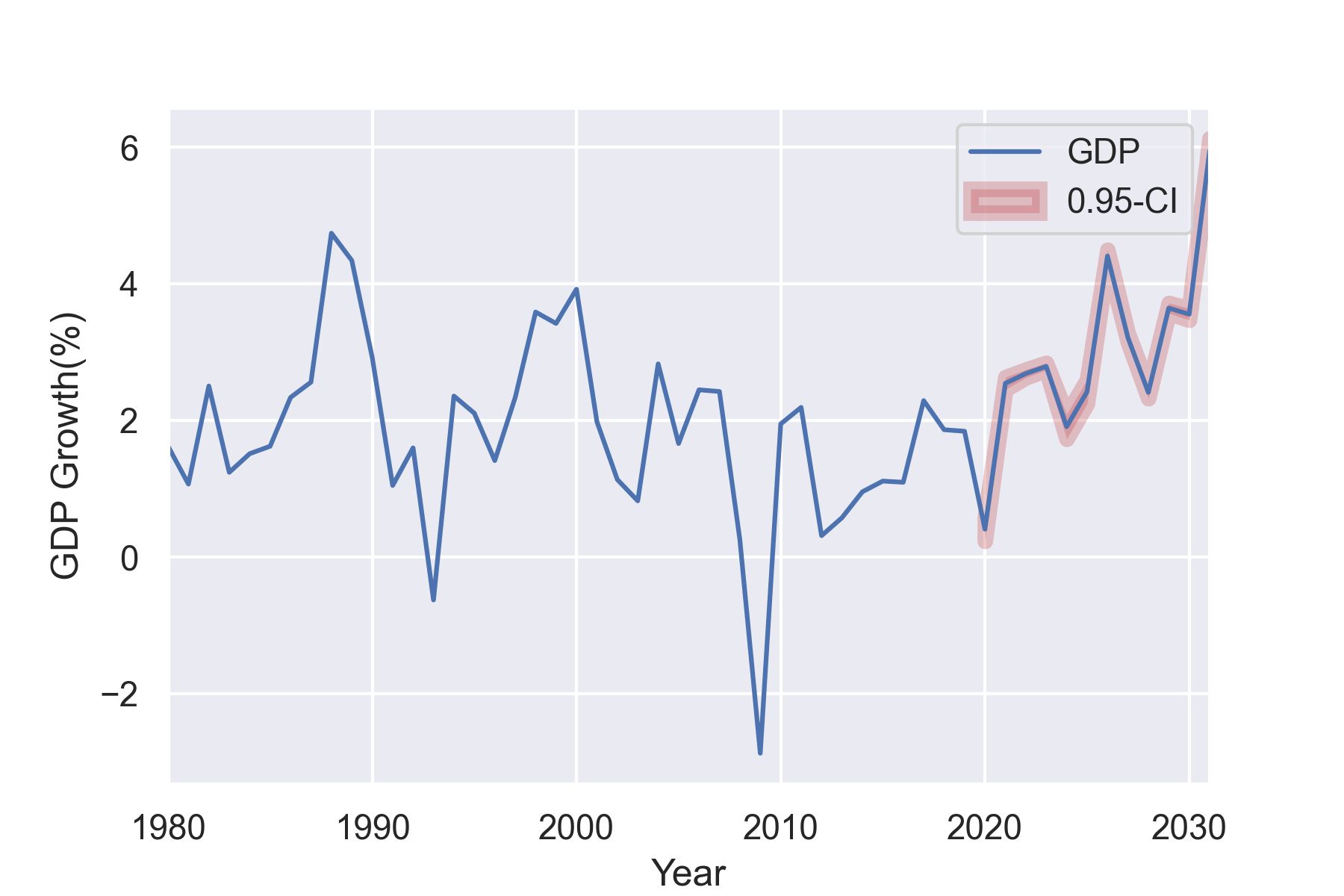}
         \caption{France GDP with the model prediction (mean) and uncertainty (95$\%$ CI). }
     \end{subfigure}
     \hfill
     \begin{subfigure}[h]{0.45\textwidth}
         \centering
         \includegraphics[width=\textwidth]{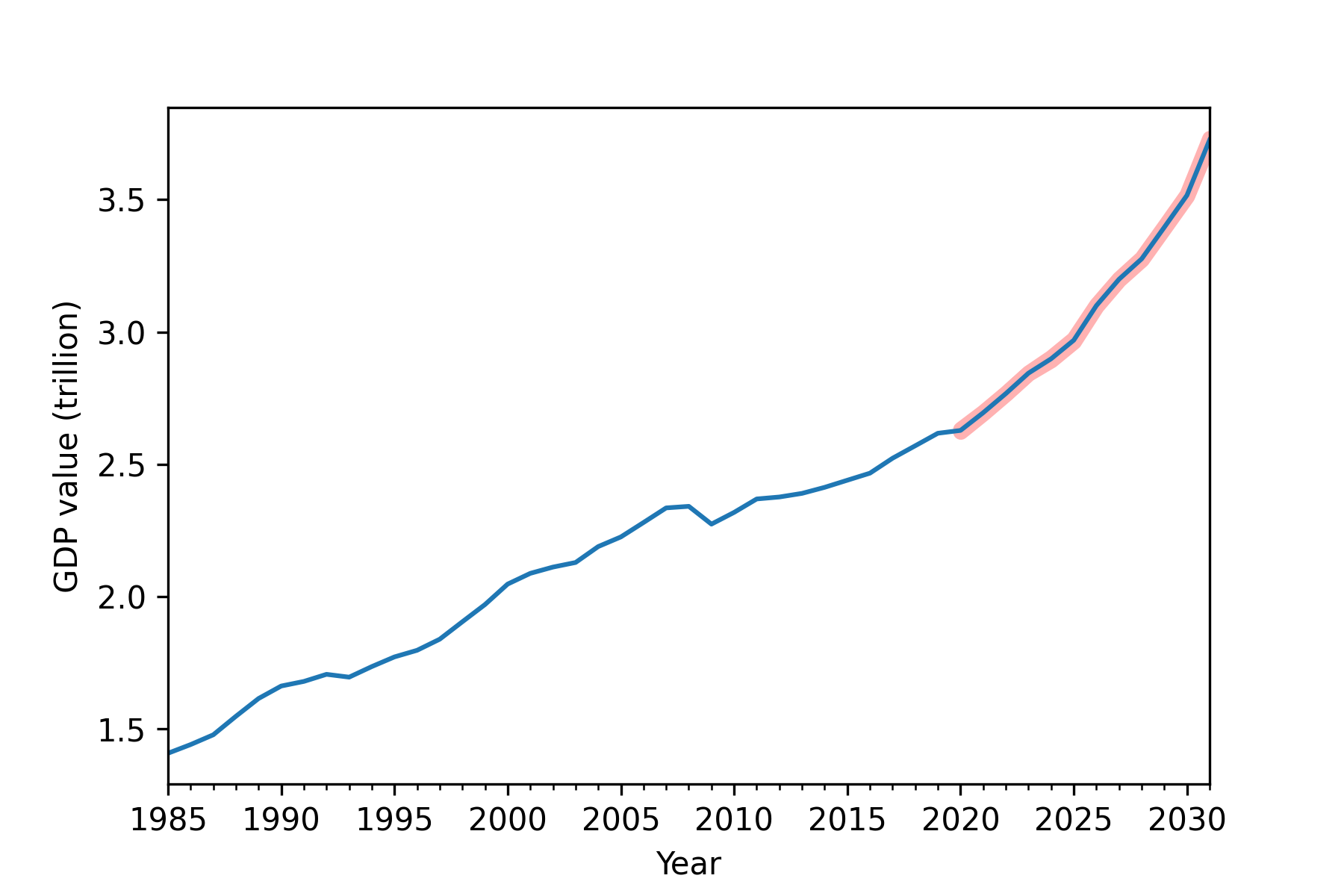}
         \caption{France GDP with the model prediction (mean) and uncertainty (95$\%$ CI).}
     \end{subfigure}
     \caption{France: actual value and recursive multivariate ED-LSTM model prediction for next decade (2020-2031).}
        \label{fig:France_final_forecast}
\end{figure}

\begin{figure}[htbp!]
     \centering
     \begin{subfigure}[h]{0.45\textwidth}
         \centering
         \includegraphics[width=\textwidth]{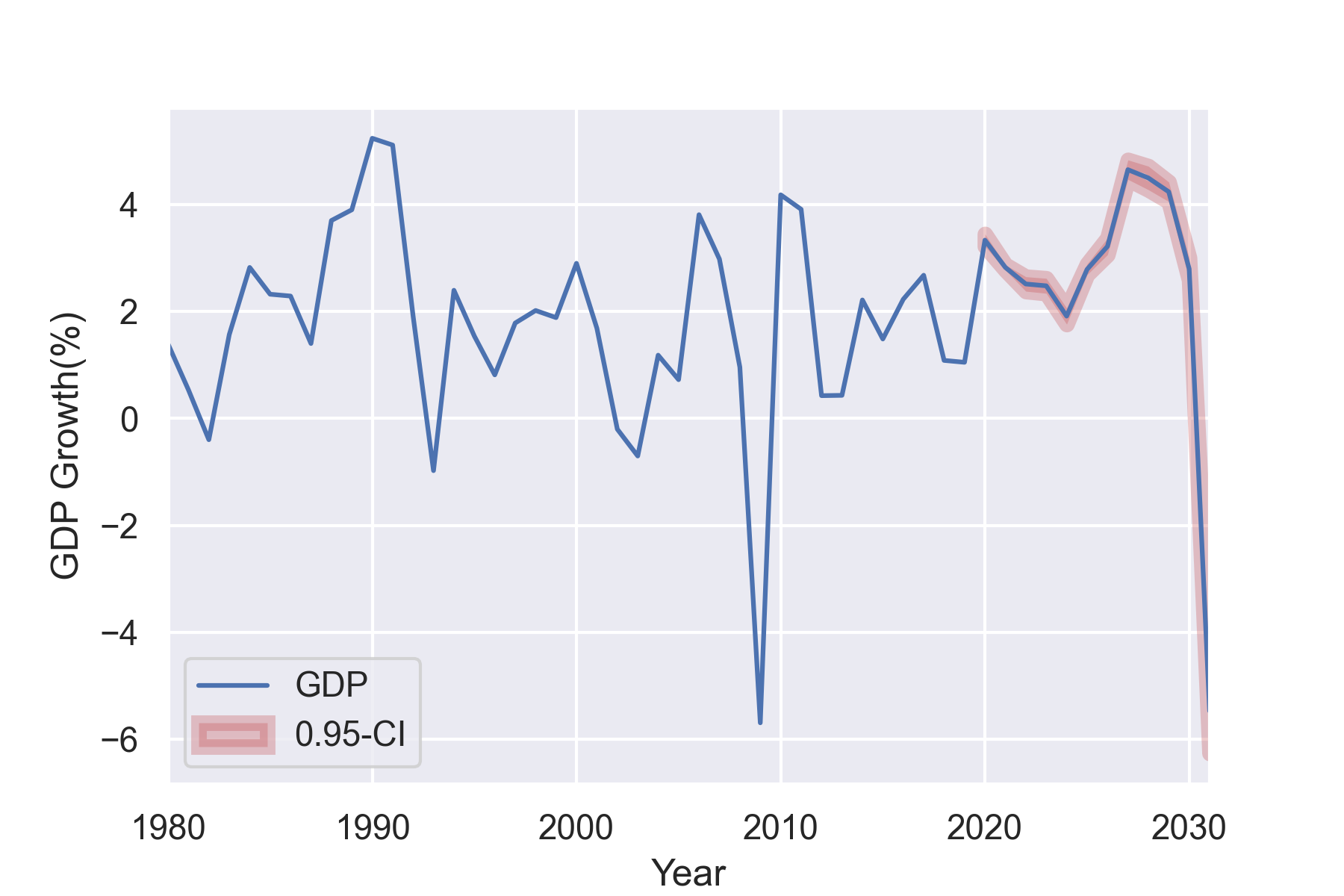}
         \caption{Germany GDP with the model prediction (mean) and uncertainty (95$\%$ CI). }
     \end{subfigure}
     \hfill
     \begin{subfigure}[htbp!]{0.45\textwidth}
         \centering
         \includegraphics[width=\textwidth]{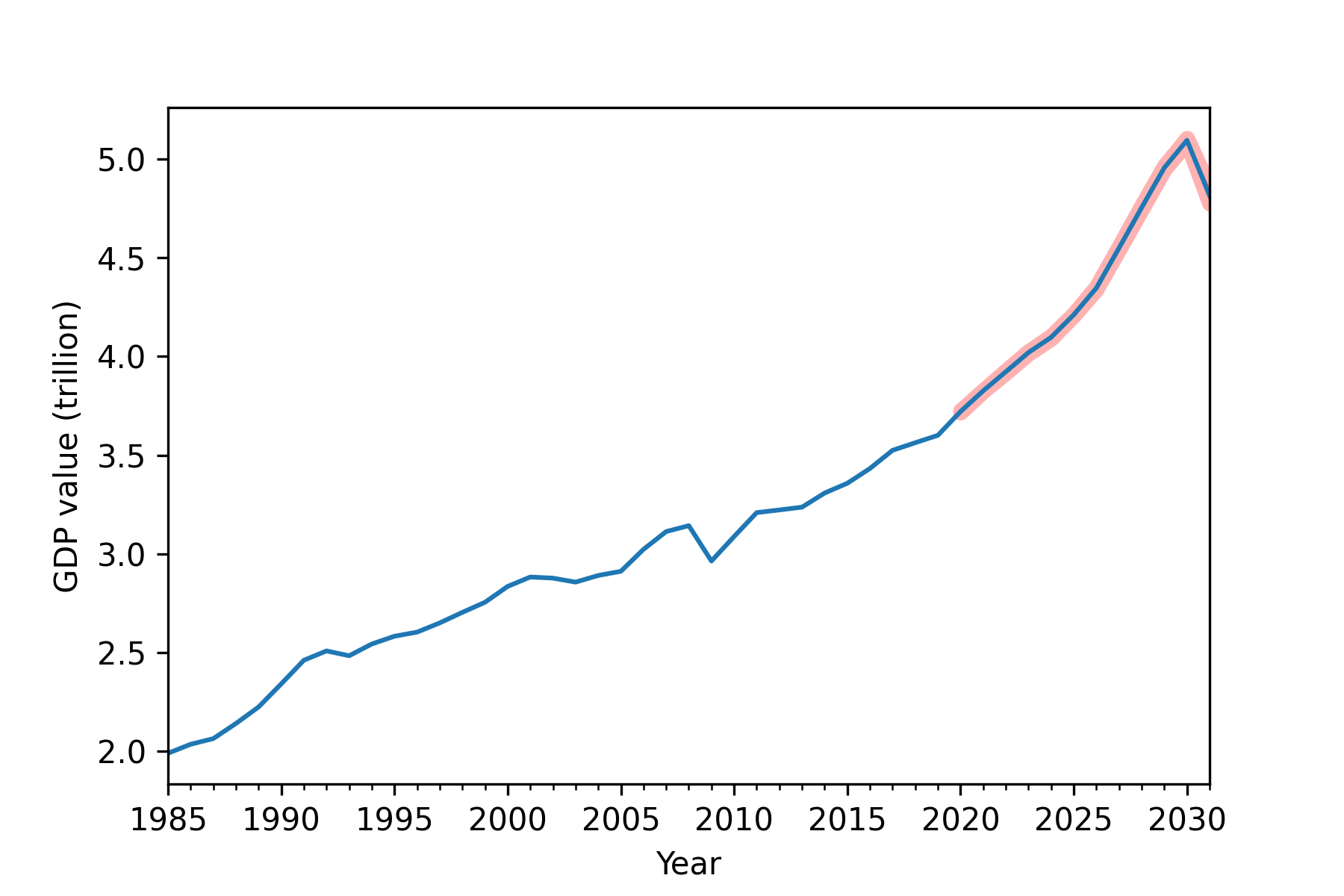}
         \caption{Germany GDP with the model prediction (mean) and uncertainty (95$\%$ CI).}
     \end{subfigure}
     \caption{Germany: actual value and recursive multivariate ED-LSTM model prediction for next decade (2020-2031).}
        \label{fig:Germany_final_forecast}
\end{figure}

\begin{figure}[htbp!]
     \centering
     \begin{subfigure}[h]{0.45\textwidth}
         \centering
         \includegraphics[width=\textwidth]{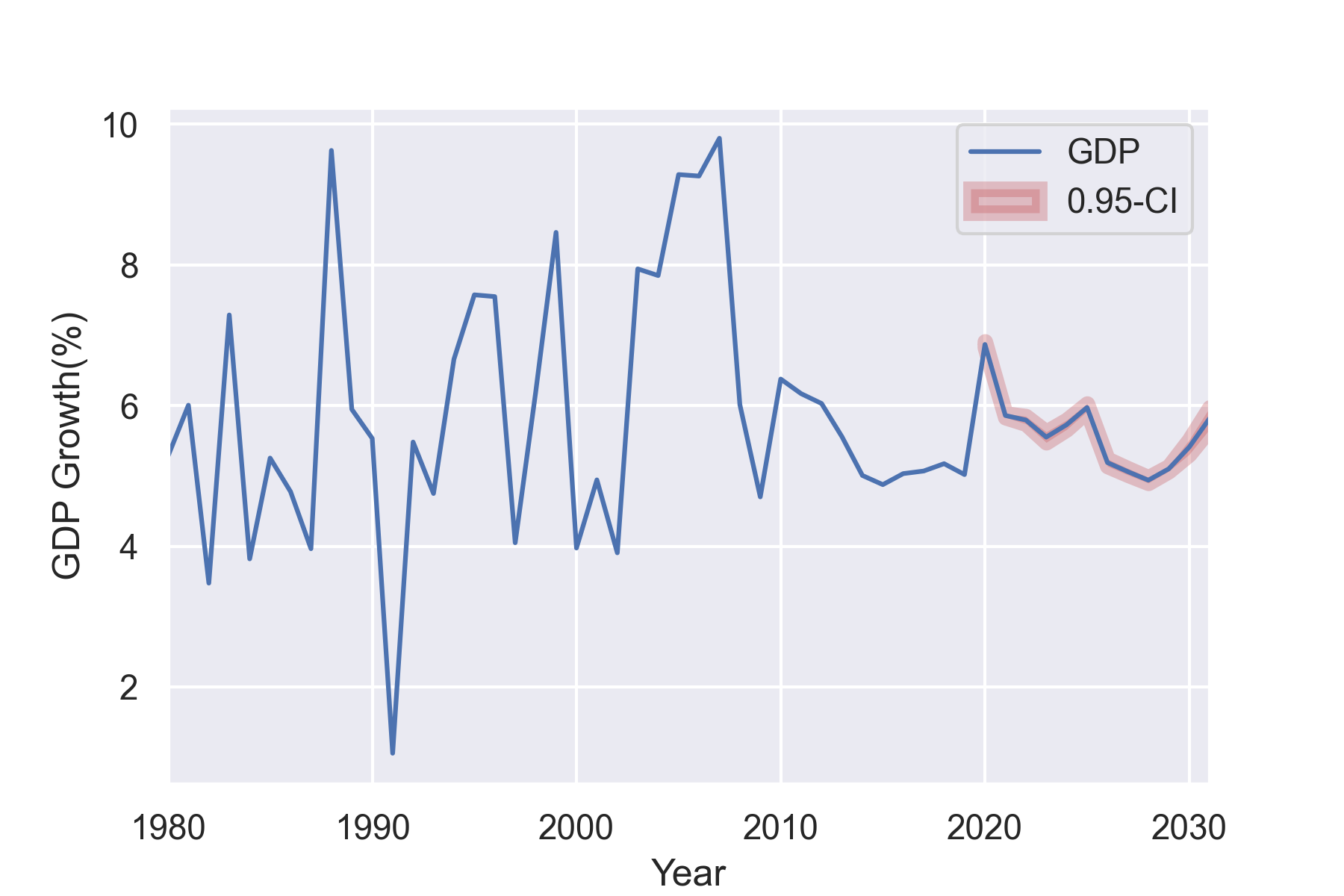}
         \caption{India GDP with the model prediction (mean) and uncertainty (95$\%$ CI). }
     \end{subfigure}
     \hfill
     \begin{subfigure}[htbp!]{0.45\textwidth}
         \centering
         \includegraphics[width=\textwidth]{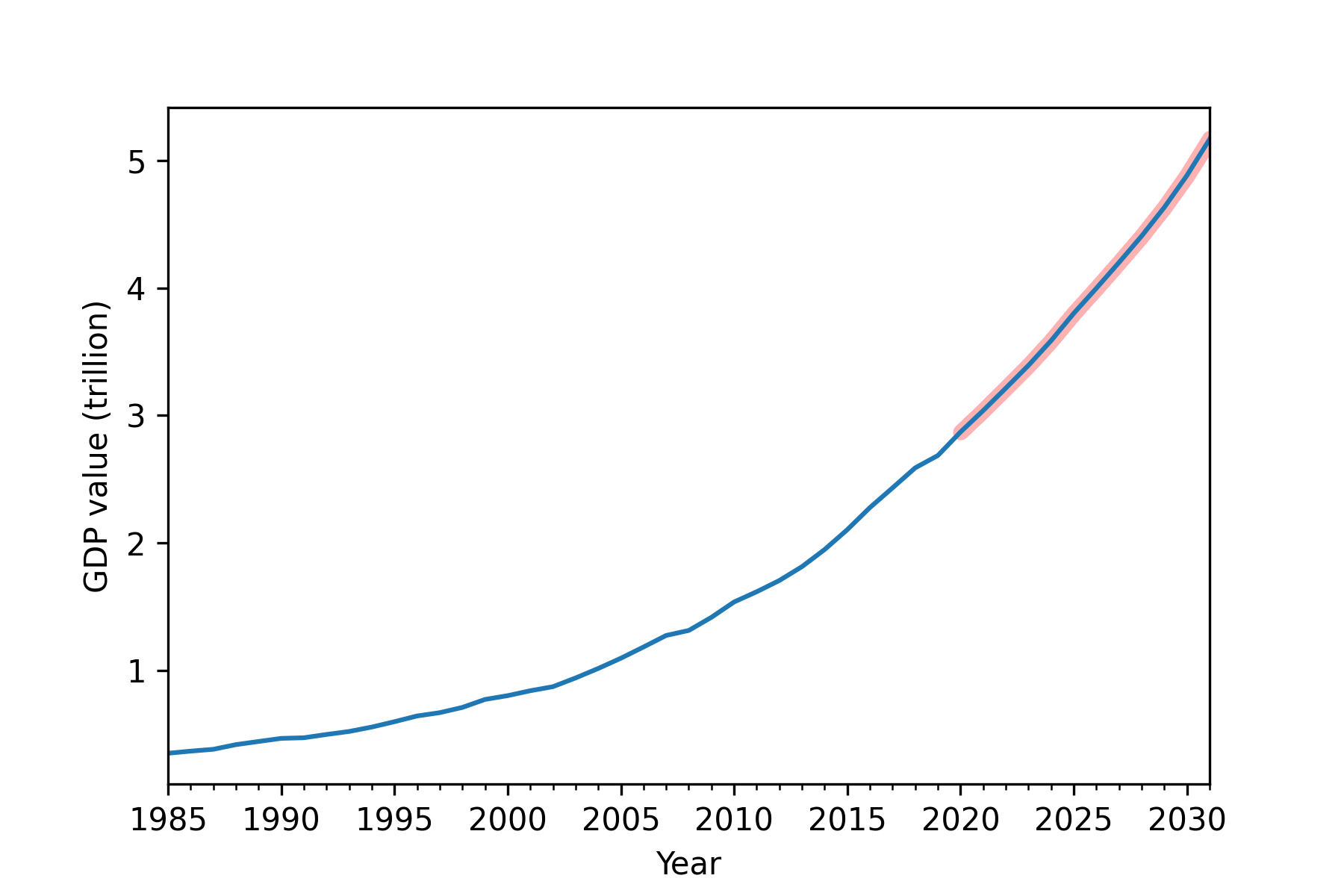}
         \caption{India GDP with the model prediction (mean) and uncertainty (95$\%$ CI).}
     \end{subfigure}
        \caption{India: actual value and recursive multivariate ED-LSTM model prediction for next decade (2020-2031).}
        \label{fig:India_final_forecast}
\end{figure}

\begin{figure}[h]
     \centering
     \begin{subfigure}[h]{0.45\textwidth}
         \centering
         \includegraphics[width=\textwidth]{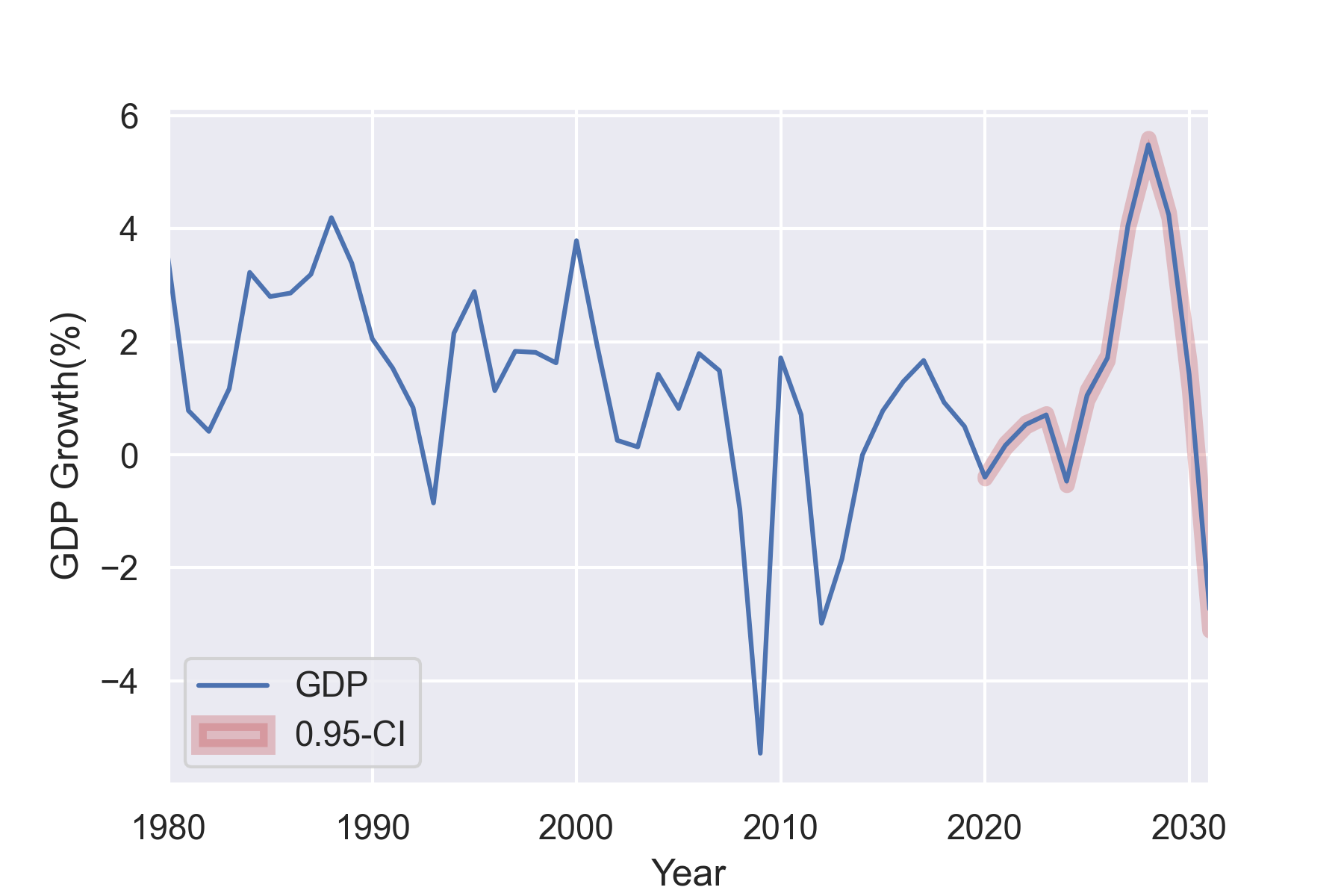}
         \caption{Italy GDP with model prediction (mean) and uncertainty (95$\%$ CI). }
     \end{subfigure}
     \hfill
     \begin{subfigure}[h]{0.45\textwidth}
         \centering
         \includegraphics[width=\textwidth]{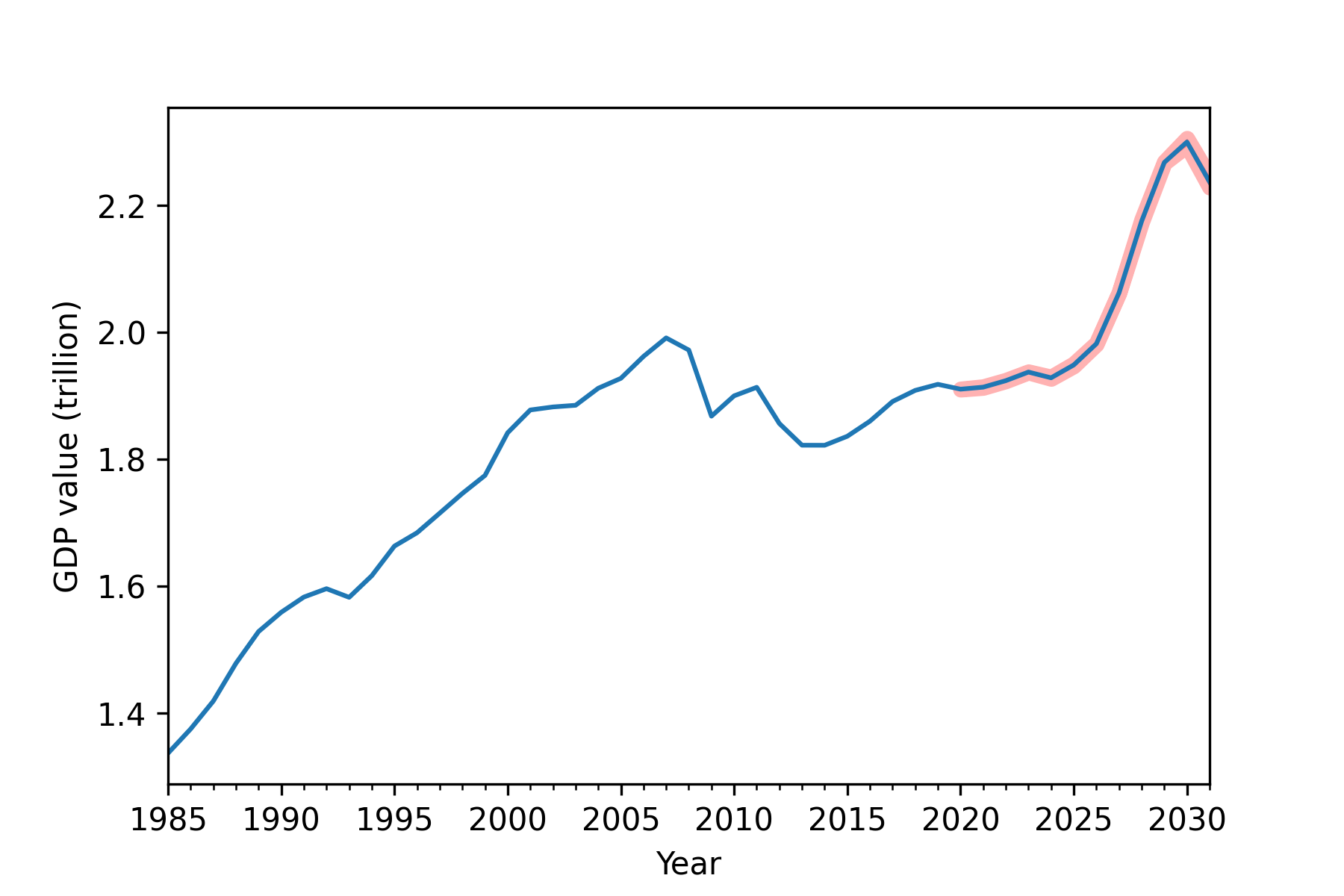}
         \caption{Italy GDP with model prediction (mean) and uncertainty (95$\%$ CI).}
     \end{subfigure}
        \caption{Italy actual value and recursive multivariate ED-LSTM model prediction for next decade (2020-2031).}
        \label{fig:Italy_final_forecast}
\end{figure}

\begin{figure}[htbp!]
     \centering
     \begin{subfigure}[h]{0.45\textwidth}
         \centering
         \includegraphics[width=\textwidth]{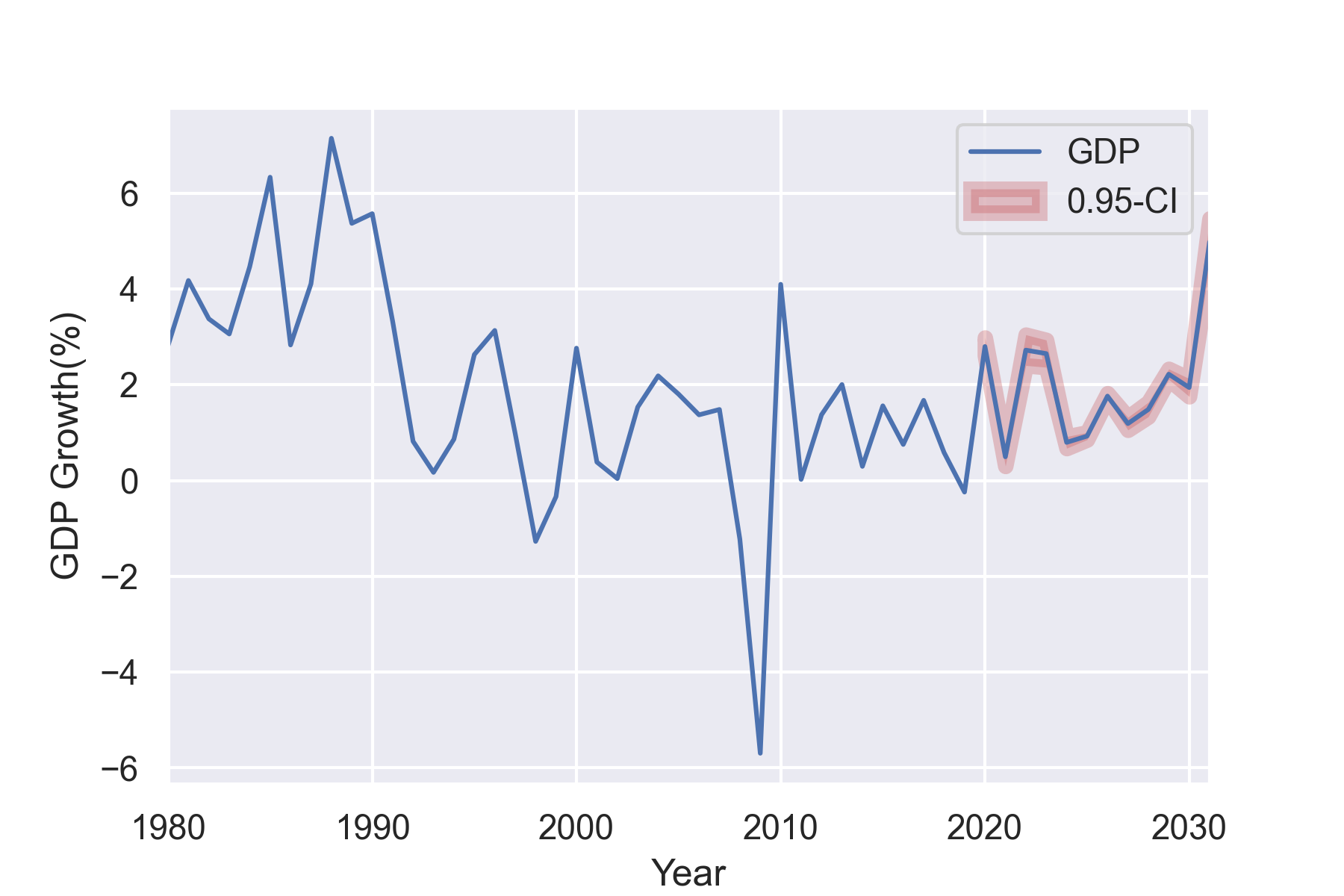}
         \caption{Japan GDP with the model prediction (mean) and uncertainty (95$\%$ CI). }
     \end{subfigure}
     \hfill
     \begin{subfigure}[h]{0.45\textwidth}
         \centering
         \includegraphics[width=\textwidth]{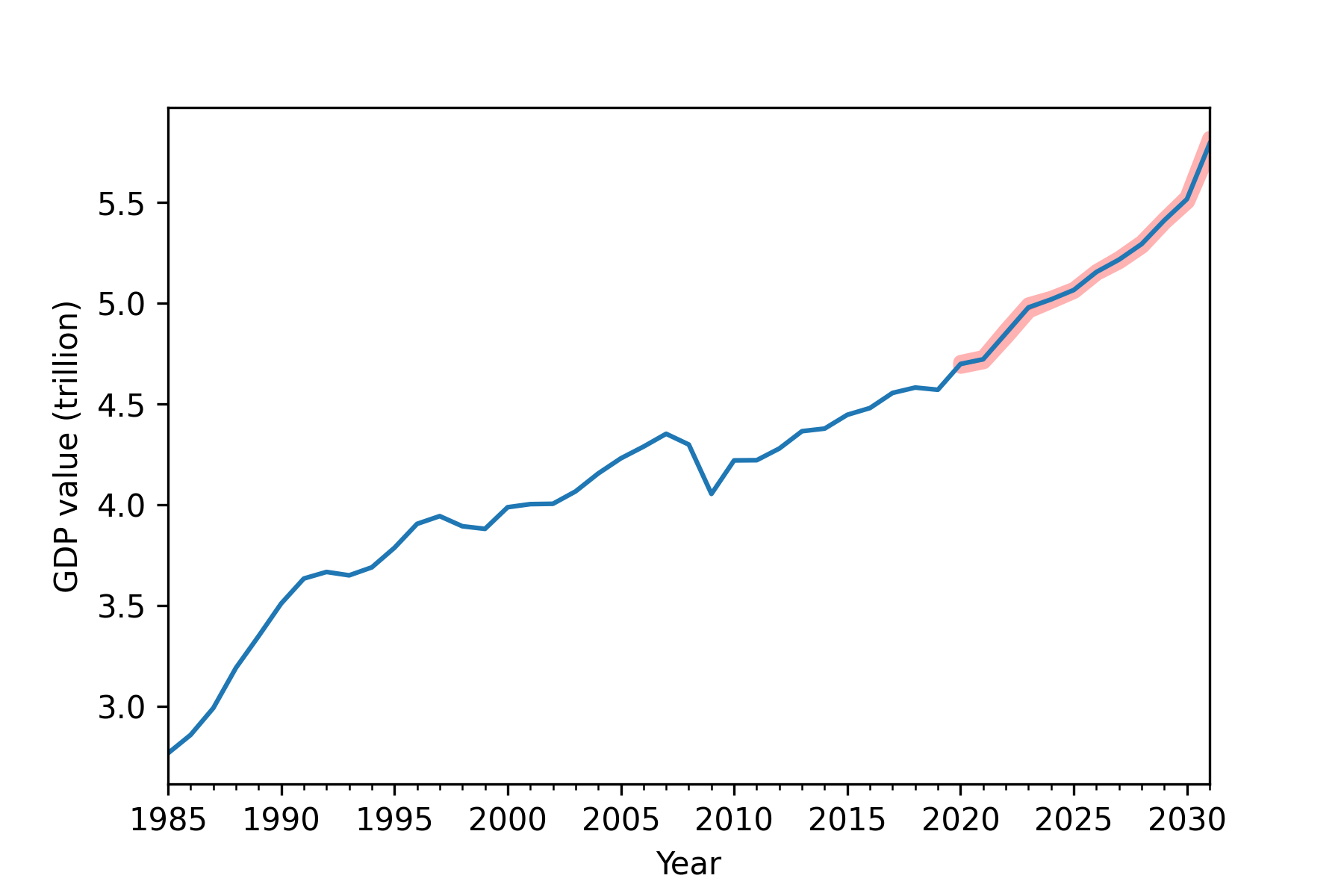}
         \caption{Japan GDP with the model prediction (mean) and uncertainty (95$\%$ CI).}
     \end{subfigure}
        \caption{Japan: actual value and recursive multivariate ED-LSTM model prediction for next decade (2020-2031).}
        \label{fig:Japan_final_forecast}
\end{figure}

\begin{figure}[htbp!]
     \centering
     \begin{subfigure}[h]{0.45\textwidth}
         \centering
         \includegraphics[width=\textwidth]{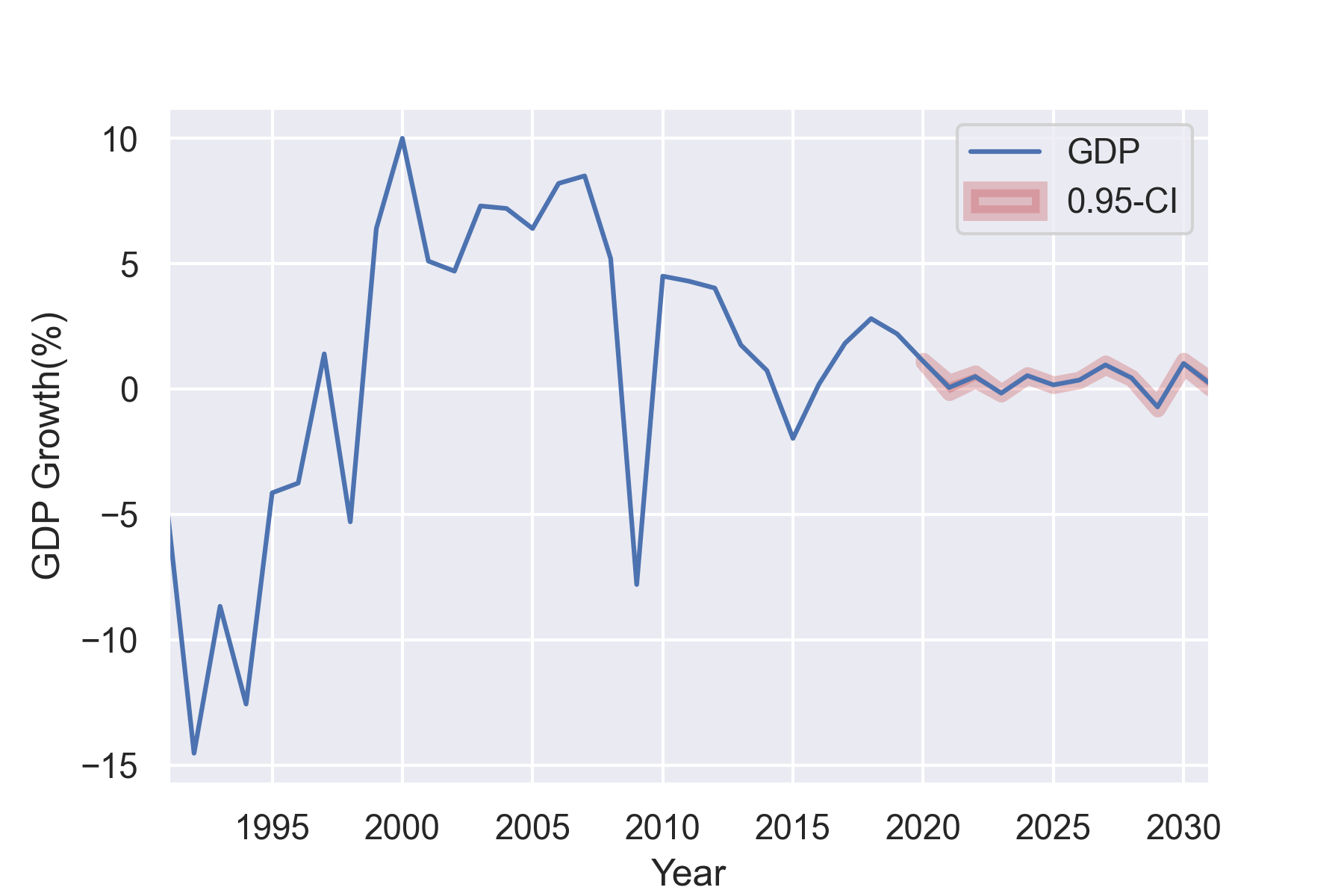}
         \caption{Russia GDP with the model prediction (mean) and uncertainty (95$\%$ CI). }
     \end{subfigure}
     \hfill
     \begin{subfigure}[h]{0.45\textwidth}
         \centering
         \includegraphics[width=\textwidth]{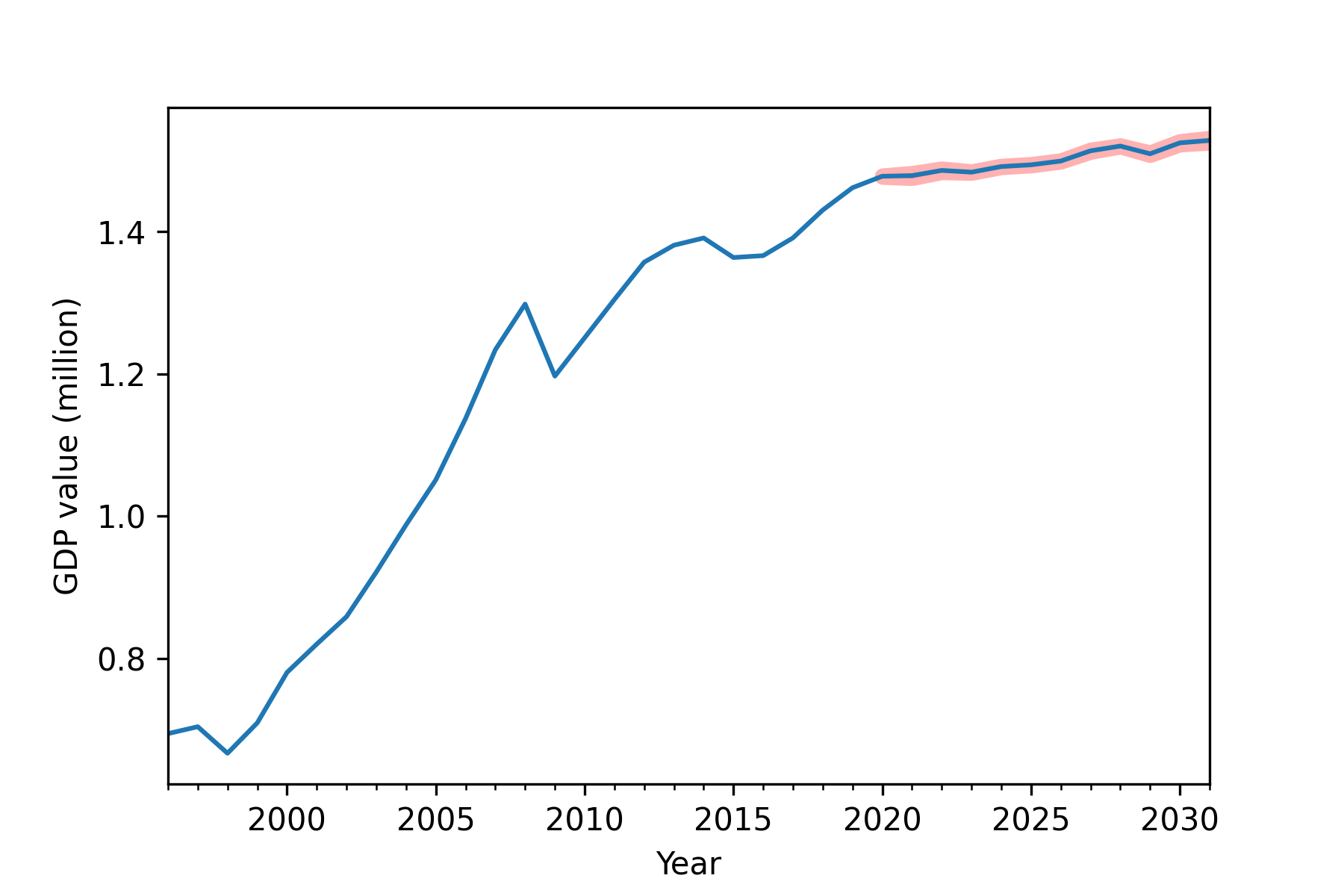}
         \caption{Russia GDP with the model prediction (mean) and uncertainty (95$\%$ CI).}
     \end{subfigure}
        \caption{Russia: actual value and recursive multivariate ED-LSTM model prediction for next decade (2020-2031).}
        \label{fig:Russia_final_forecast}
\end{figure}

\begin{figure}[htbp!]
     \centering
     \begin{subfigure}[h]{0.45\textwidth}
         \centering
         \includegraphics[width=\textwidth]{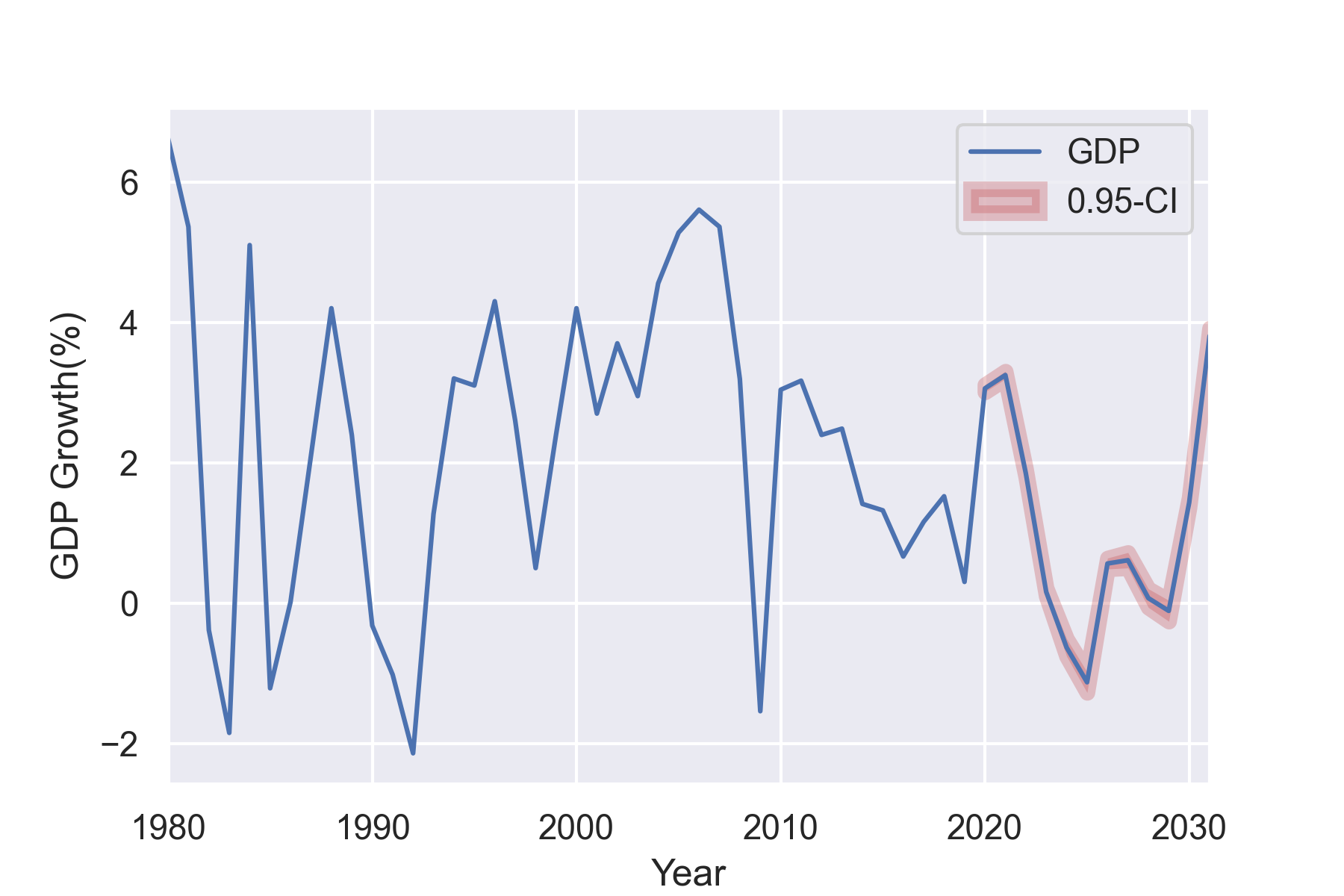}
         \caption{South Africa with the model prediction (mean) and uncertainty (95$\%$ CI). }
     \end{subfigure}
     \hfill
     \begin{subfigure}[h]{0.45\textwidth}
         \centering
         \includegraphics[width=\textwidth]{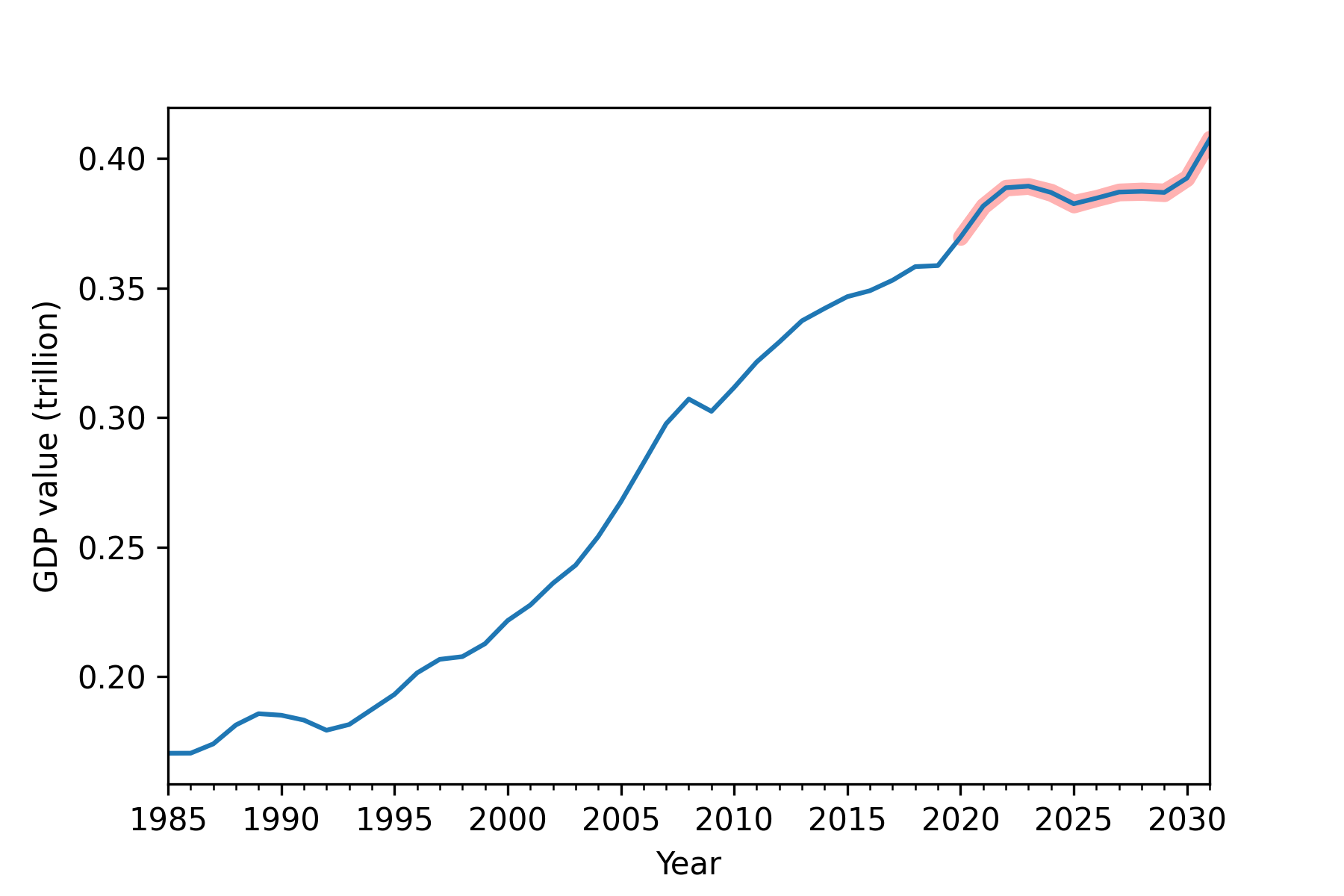}
         \caption{South Africa GDP with the model prediction (mean) and uncertainty (95$\%$ CI).}
     \end{subfigure}
        \caption{South Africa: actual value and recursive multivariate ED-LSTM model prediction for next decade (2020-2031).}
        \label{fig:South_Africa_final_forecast}
\end{figure}

\begin{figure}[htbp!]
     \centering
     \begin{subfigure}[h]{0.45\textwidth}
         \centering
         \includegraphics[width=\textwidth]{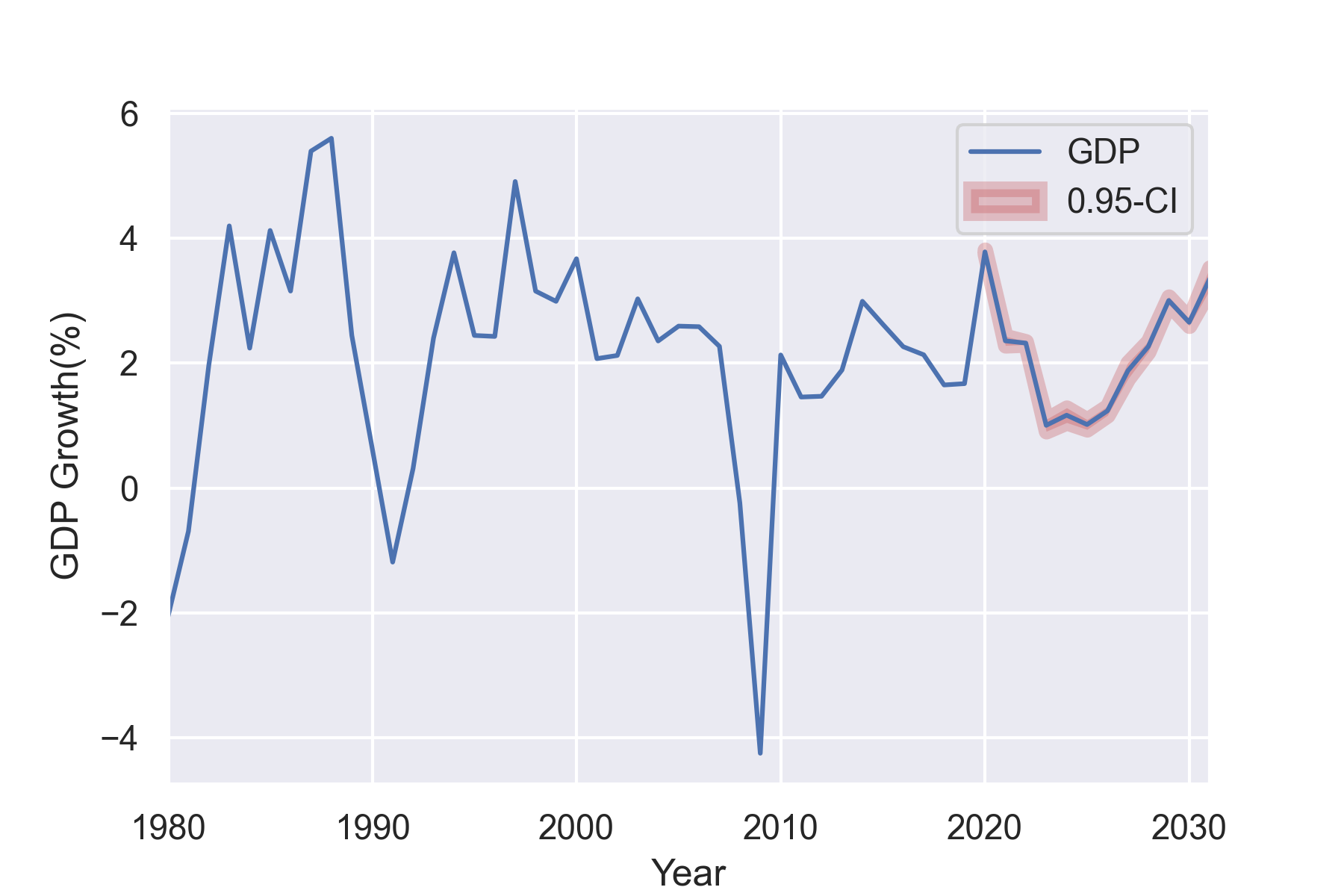}
         \caption{United Kingdom GDP with the model prediction (mean) and uncertainty (95$\%$ CI). }
     \end{subfigure}
     \hfill
     \begin{subfigure}[h]{0.45\textwidth}
         \centering
         \includegraphics[width=\textwidth]{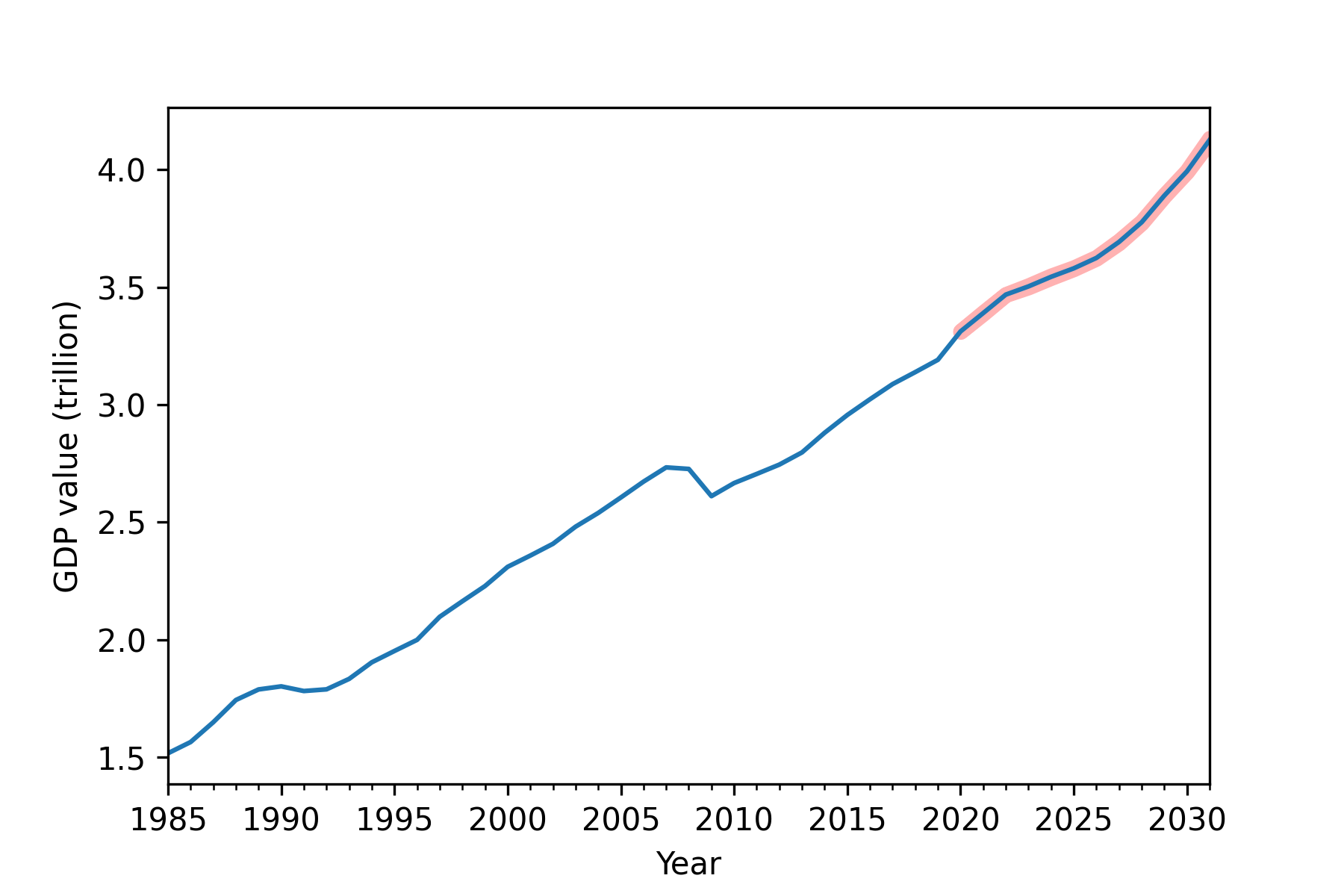}
         \caption{United Kingdom GDP with model prediction (mean) and uncertainty (95$\%$ CI).}
     \end{subfigure}
        \caption{United Kingdom: actual value and recursive multivariate ED-LSTM model prediction for next decade (2020-2031).}
        \label{fig:United_Kingdom_final_forecast}
\end{figure}

\begin{figure}[htbp!]
     \centering
     \begin{subfigure}[h]{0.45\textwidth}
         \centering
         \includegraphics[width=\textwidth]{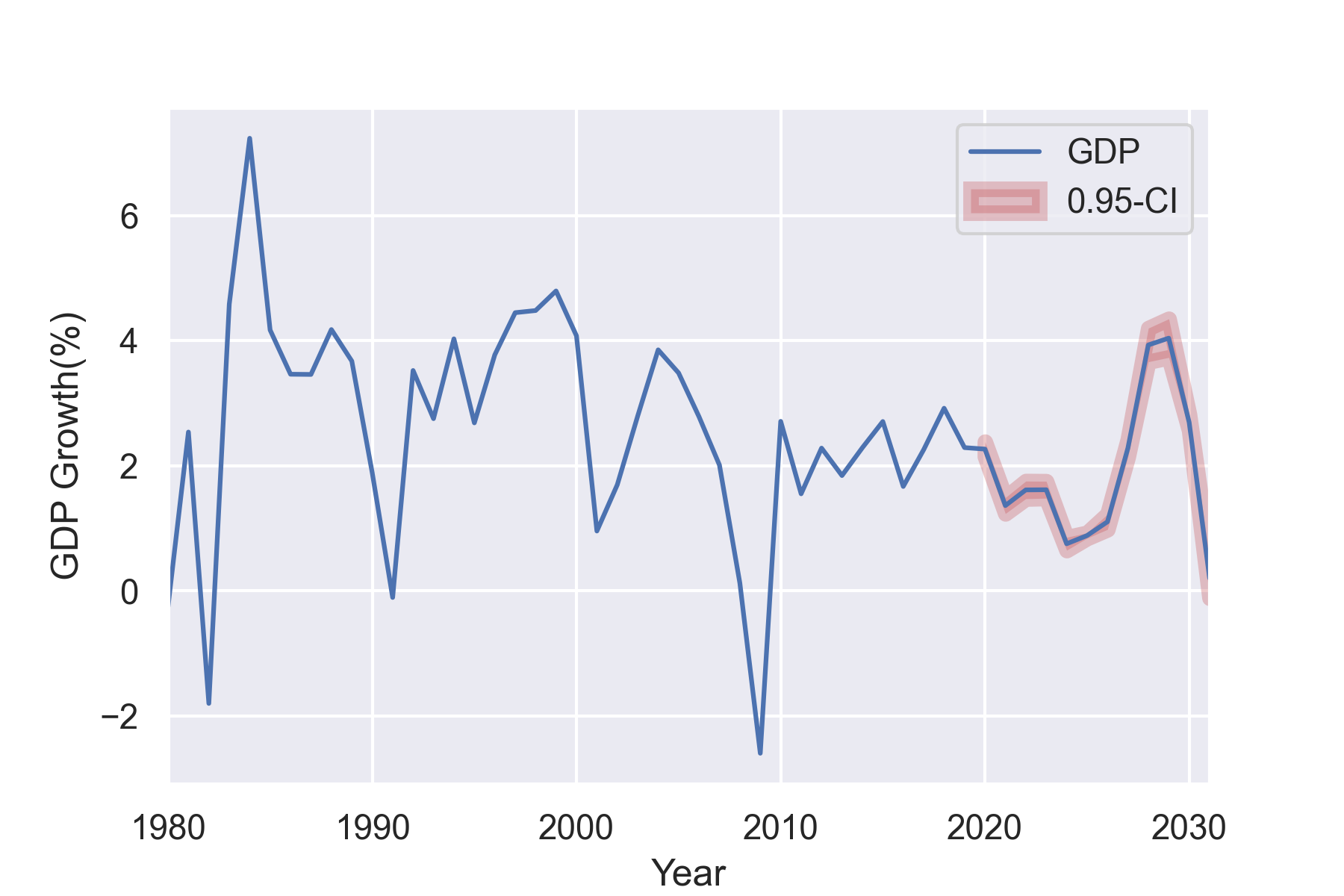}
         \caption{United States GDP with the model prediction (mean) and uncertainty (95$\%$ CI). }
     \end{subfigure}
     \hfill
     \begin{subfigure}[h]{0.45\textwidth}
         \centering
         \includegraphics[width=\textwidth]{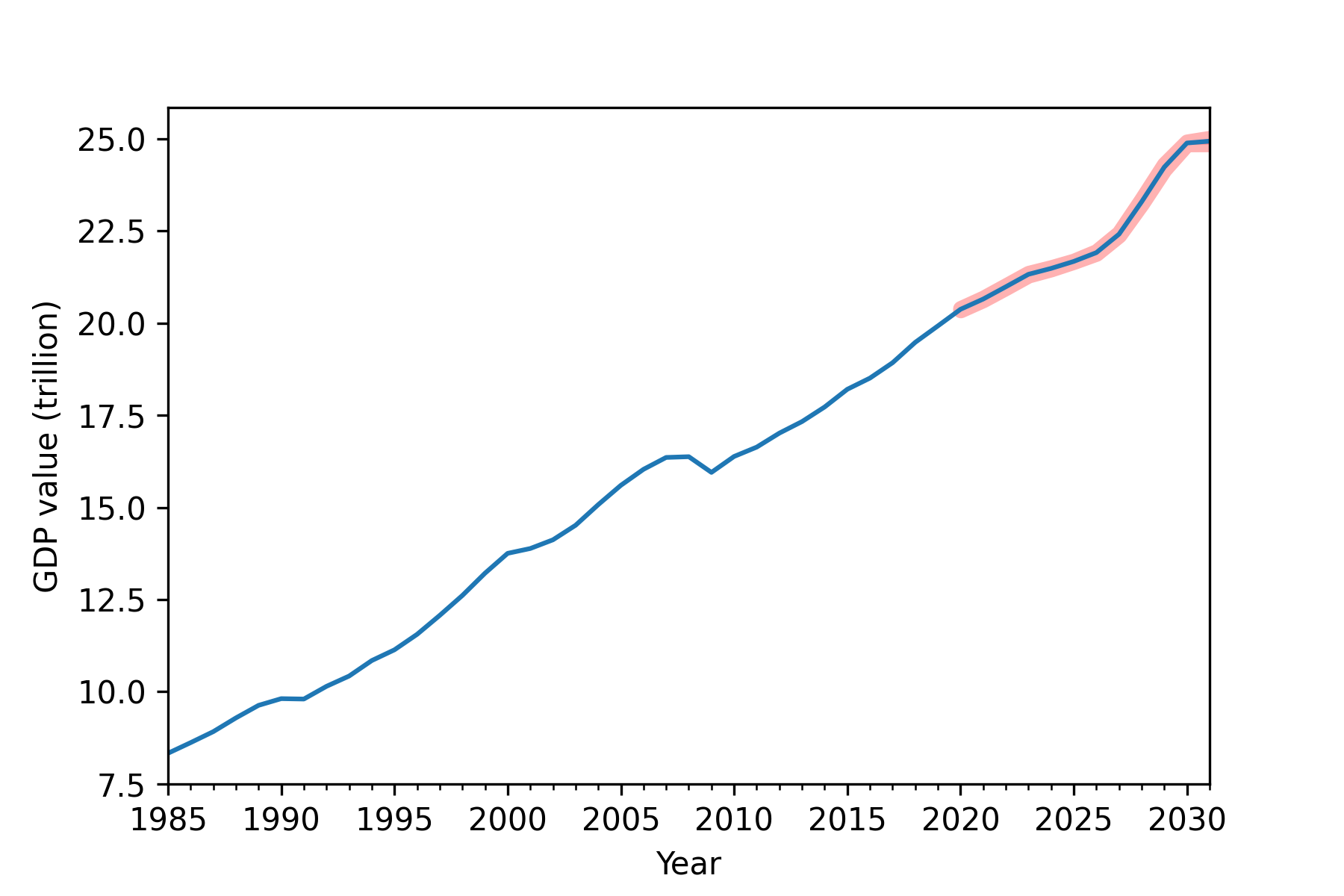}
         \caption{United States GDP with the model prediction (mean) and uncertainty (95$\%$ CI).}
     \end{subfigure}
        \caption{United States: actual value and recursive multivariate ED-LSTM model prediction for next decade (2020-2031).}
        \label{fig:United_States_final_forecast}
\end{figure}

As can be seen from Table \ref{recursive:RMSE}, the prediction results for the features perform well and the RMSE of the test set is equivalent to the training set. Although we used GDP growth rate as the target in the \textit{direct strategy} to train and select the optimal model, the results in Table \ref{recursive:RMSE} tell us that the ED-LSTM model performs well even if the target is replaced with other features in the data set. This also indicates that we can use  \textit{recursive strategy}, i.e., we can predict the features (economic indicators) first and then the GDP growth rate, which demonstrates the feasibility of our recursive strategy. Figures \ref{fig:Australia_final_forecast} \ref{fig:Brazil_final_forecast} \ref{fig:Canada_final_forecast} \ref{fig:China_final_forecast} \ref{fig:France_final_forecast} \ref{fig:Germany_final_forecast} \ref{fig:India_final_forecast} \ref{fig:Italy_final_forecast} \ref{fig:Japan_final_forecast} \ref{fig:Russia_final_forecast} \ref{fig:South_Africa_final_forecast} \ref{fig:United_Kingdom_final_forecast} \ref{fig:United_States_final_forecast} represent the real value from 1980 to 2019 and the prediction from 2020 to 2031, the predictions include the mean of the forecast over 30 experiment runs of model training and the 95$\%$ confidence interval (CI) of the results which are shown in the black line with red shade. We observe that the 95$\%$ confidence interval in each plot is narrow, demonstrating that the results of these 30 experiments did not exhibit large variations in predictions.

\begin{table*}[htbp!]  
    \caption{Ranking of different models for respective datasets (country) according to the RMSE.}
    \label{tab:rk_ark}
\centering 
\begin{tabular}{ c c c c c c c} 
\hline
Country & ARIMA & VAR & LSTM & BD-LSTM & ED-LSTM & CNN  \\ 
\hline
\hline
Australia & 3 & 6 & 5 & 4 & 1 & 2 \\ 
Brazil  & 3 & 2 & 6 & 5 & 1 & 4 \\ 
Canada & 5 & 1 & 6 & 4 & 2 & 3 \\ 
China & 4 & 6 & 5 & 3 & 1 & 2  \\ 
France & 6 & 2 & 4 & 5 & 3 & 1 \\ 
Germany & 5 & 3 & 6 & 4 & 2 & 1  \\ 
India & 6 & 1 & 5 & 3 & 4 & 2 \\ 
Italy  & 3  & 6 & 5 & 4 & 1 & 2\\ 
Japan & 3 & 5 & 4 & 6 & 2 & 1  \\ 
Russia & 4 & 6 & 3 & 5 & 1 & 2  \\ 
South Africa & 6 & 1 & 5 & 4 & 3 & 2  \\ 
United Kingdom & 5 & 4 & 6 & 2 & 1 & 3  \\ 
United States & 6 & 1 & 5 & 4 & 2 & 3  \\ 
\hline
Avg-Rank & 4.53 & 3.38 & 5 & 4.07 & 1.84 & 2  \\ 

Mode-Rank & 3 and 6 & 6  & 5 & 4 & 1 & 2  \\ 
\hline
\hline
\end{tabular}

\end{table*}

\subsection{Comparison with institutional forecasts}

In this section, we compare our model forecast of total GDP given in USD (trillions)  with the forecast based on the International Monetary Fund (IMF) for selected Asian countries, Australia, Russia and the United States in Table\ref{Compare:Lowy} for the year 2027 obtained from the IMF World Economic Outlook Database October 2022 \cite{Lowy2030forecast}.  We can see that aside from China and India, our forecasts are similar to the IMF. The IMF forecast for China's GDP is slightly higher than our expectations, along with the IMF’s forecast for India. We recognise that changing geopolitical circumstances in the case of China imply that it is unlikely to grow at the same rate.  India may benefit from a more stable birthrate and a shift of manufacturing from the  Chinese industry due to US trade disputes. The changing geopolitical circumstances are issues that we need to consider in our future work. Note that our model forecast was based on data before 2019; hence our model did not consider the impact of COVID-19.  This was accounted for by the IMF as the forecasts were released in 2022 \cite{Lowy2030forecast},  which explains the major difference in the forecasts (Table\ref{Compare:Lowy})  in the case of the United States.


\begin{table*}[htbp!]
    \caption{We compare our model forecast with IMF forecasts for 2027 in USD (Trillion). Note that we obtained the IMF forecast from the October 2022 IMF World Economic Outlook \cite{Lowy2030forecast}.}
    \label{Compare:Lowy}
\centering 
\begin{tabular}{ c c c c} 
\hline
Country & Model Forecast & IMF & Difference  \\ 
\hline
\hline
Australia & 1.70& 2.08 &0.38 \\ 
China & 26.00&26.44 &0.44 \\ 
India & 4.30&5.36&1.06 \\ 
Japan & 5.10&5.17&0.07  \\ 
Russia & 1.50&2.24&0.74  \\ 
United States & 22.40 &30.28&7.88  \\ 
\hline
\hline
\end{tabular}
\end{table*}


\begin{figure}[htbp!]
 \centering
 \includegraphics[width=0.45\textwidth]{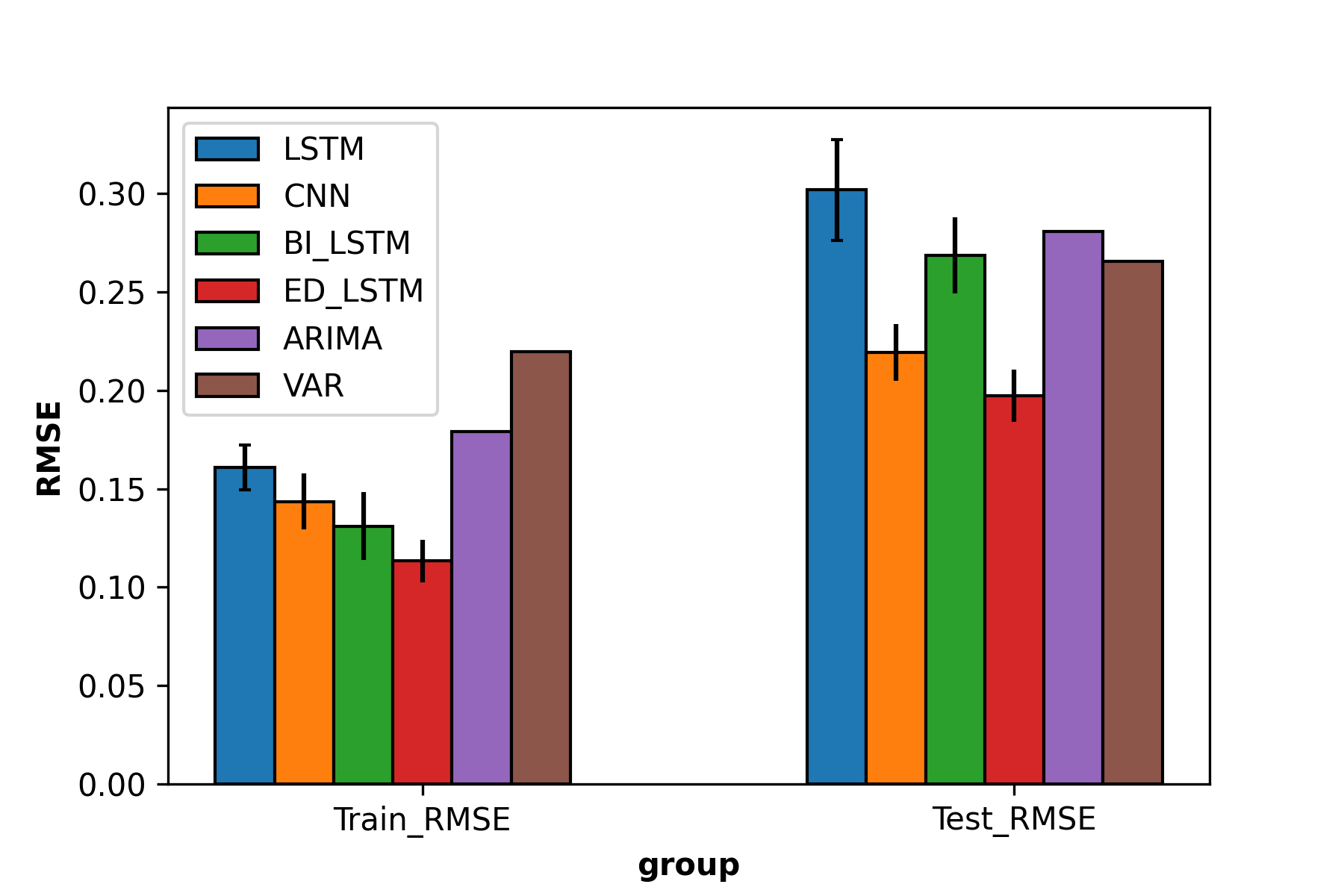}
 \caption{The mean RMSE of 13 countries using 6 models.}
 \label{fig:Mean_RMSE_13}
\end{figure}

\section{Discussion}
 
 In our experiments, we first evaluated our framework using the direct strategy. We selected the data from 13 countries in the Penn World Table from 1980 to 2019  for selected developed and developing countries, respectively. Our results indicated that both for developed countries and developing countries, ED-LSTM provided the overall best results. The CNN and ED-LSTM models also provided the best performance overall with average and mode rank (Table \ref{tab:rk_ark}). The ARIMA model slightly lagged behind the performance of the VAR in the ranking, illustrating the importance of the exogenous variables in this task.  Figure \ref{fig:Mean_RMSE_13} presents a summary showing the mean RMSE of 13 countries using the respective models, where CNN and ED-LSTM provide the best performance on the test datasets.

Based on the model evaluation, we chose ED-LSTM for the recursive strategy as the prediction model for decadal forecasts.  The GDP forecast shows that except for  France, the rest of the developed countries (i.e. Australia, Canada, Germany, Italy, Japan,  United Kingdom, and  the United States) will experience economic growth slowdown or even recession in the first five years of the next decade. These countries will gradually return to normal and improve afterwards. Among developing countries, our model suggests that Brazil, Russia and South Africa will be impacted by the economic crisis in the future, and their economic development will stagnate in the near term. However, China and India which are the two fastest-growing developing countries will maintain a development speed higher than 4$\%$ (Figure \ref{fig:China_final_forecast} \ref{fig:India_final_forecast}). The Centre for Economics and Business Research  \cite{CEBR2035India} forecasted that India will become a 10 trillion dollar GDP by 2035 and Morgan Stanley forecasted that Indian GDP will reach 11 trillion by 2032 \cite{MS2027India}. Due to the impacts of COVID-19 and the Russia-Ukraine war, these forecasts are being rapidly adjusted by the respective organizations, including the  International Monetary Fund  and the World Bank. 
 
There are certain limitations of our study due to the timeline  featured in the datasets, since the data collected during the COVID-19 pandemic has not been included in the modelling process. The timeline considered was due to the Penn World Table dataset that did not feature data after 2019 when retrieved in June 2022 for our experiments. The world economy in recent years has experienced significant turbulence due to COVID-19, international trade disputes, and the Russia-Ukraine war \cite{ozili2023spillover}.  It is well known that a country's economy cannot grow stably and infinitely, it is limited by its own political and social  conditions \cite{modis2013long}. In our GDP forecast of China, the total GDP  reached an incredible value after steady growth. However, in reality, the GDP growth rate of China slowed down in recent years due to the housing crisis \cite{jiang2022china}, trade disputes with the United States\cite{fajgelbaum2022economic}, and COVID-19 \cite{habibi2022potential}. The Chinese policy has also shifted from high-speed growth to high-quality growth \cite{ding2022digital}. These factors are difficult to be captured by our framework given the lack of data; hence, this is also a major limitation. 

Although we defined train and test sets using the time period given by the years, the shuffling strategy can be used to create train/test split from the entire time period, i.e. 1980-2019. If this is done, the true strength of the shuffling strategy can be utilized since both the train/test set would have data from different decades, and this would avoid biases in different trends of growth in different decades for the respective countries. Note that ultimately, we would need to develop a model using a direct strategy approach where such biases are minimized.
  
Future work can consider additional considerations for improving  our framework. Firstly, we can incorporate robust uncertainty quantification in model development using Bayesian deep learning model\cite{neal1992bayesian,chandra2020bayesian}. Uncertainty quantification in model predictions can provide better insights into  extreme situations that will affect the economy, such as COVID-19, and trade disputes. In addition, we need to consider the relationship between economic volume and economic growth rate \cite{lewis2013theory,ccalicskan2015technological}.  \textcolor{black}We need data from  2020 - 2024 that features COVID-19 which needs to be used to train the model so that we can have model predictions for 2025 - 2035. Currently, most institutional forecasts are pointing to a possibility of a recession beginning  in the second half of 2023 for France. 
We provide open-source code for our framework that can be updated as more data is available to make future decadal forecasts.

\section{Conclusion}

We proposed a deep learning framework for predicting the decadal GDP growth rate that included a  direct strategy and a recursive strategy. We presented a novel framework that combined direct and recursive deep learning models for multistep ahead prediction. The goal of the direct strategy was to evaluate which deep learning model was best suited to be used in the recursive strategy.  In the part of the direct strategy, our experiment selected 13 countries and nearly 40 years of data in the macroeconomic data set Penn World Table. We found that shuffled training data effectively avoided the overfitting of the  model. We compared the performance of traditional time series models (ARIMA, and VAR) with deep learning models (CNN, LSTM, BD-LSTM and ED-LSTM) and found that ED-LSTM is the best-performing model. Hence, we later used ED-LSTM as the model for the recursive strategy to predict the decadal GDP growth rate of the selected developed and developing countries. The challenge has been in the prediction of decadal  (long-term) forecasts, since we need to first estimate the economic indicators in order to forecast the GDP growth rate. This is challenging given that we have limited data, i.e. we only considered the last four decades of data on a yearly basis (less than 40 data points). 

Our proposed framework  provided accurate forecasts that compare well with institutional forecasts from the literature. A further challenge that econometric methods face compared with deep learning methods is forecast accuracy. Our results show that in the next ten years, most countries will experience an economic slowdown in terms of the GDP growth rate -- negative growth in the first five years,  and will return to normal in the later years. Our model only predicts China, France and India  to stabilize their GDP growth rate over the next decade.  We note that this is based on data till 2019 and the model needs to be trained with the latest data to get  the latest decadal forecasts.

\section*{Code and Data}



We provide open source code and data for further work \footnote{\url{https://github.com/sydney-machine-learning/deeplearning-decadalworldeconomy}.}


 
\clearpage

 \bibliographystyle{IEEEtran}

\bibliography{refs}

\begin{thebibliography}{100}
\providecommand{\url}[1]{#1}
\csname url@samestyle\endcsname
\providecommand{\newblock}{\relax}
\providecommand{\bibinfo}[2]{#2}
\providecommand{\BIBentrySTDinterwordspacing}{\spaceskip=0pt\relax}
\providecommand{\BIBentryALTinterwordstretchfactor}{4}
\providecommand{\BIBentryALTinterwordspacing}{\spaceskip=\fontdimen2\font plus
\BIBentryALTinterwordstretchfactor\fontdimen3\font minus
  \fontdimen4\font\relax}
\providecommand{\BIBforeignlanguage}[2]{{%
\expandafter\ifx\csname l@#1\endcsname\relax
\typeout{** WARNING: IEEEtran.bst: No hyphenation pattern has been}%
\typeout{** loaded for the language `#1'. Using the pattern for}%
\typeout{** the default language instead.}%
\else
\language=\csname l@#1\endcsname
\fi
#2}}
\providecommand{\BIBdecl}{\relax}
\BIBdecl

\bibitem{jansen2018combining}
W.~J. Jansen and J.~M. de~Winter, ``Combining model-based near-term {GDP}
  forecasts and judgmental forecasts: A real-time exercise for the {G7}
  countries,'' \emph{Oxford Bulletin of Economics and Statistics}, vol.~80,
  no.~6, pp. 1213--1242, 2018.

\bibitem{callen2012gross}
T.~Callen, ``Gross domestic product: An economy’s all,'' \emph{International
  Monetary Fund: Washington, DC, USA}, 2012.

\bibitem{lepenies2016power}
\BIBentryALTinterwordspacing
P.~Lepenies, \emph{The Power of a Single Number}.\hskip 1em plus 0.5em minus
  0.4em\relax New York Chichester, West Sussex: Columbia University Press,
  2016. [Online]. Available: \url{https://doi.org/10.7312/lepe17510}
\BIBentrySTDinterwordspacing

\bibitem{akkemik2007response}
K.~A. Akkemik, ``The response of employment to gdp growth in turkey: an
  econometric estimation,'' \emph{Applied Econometrics and International
  Development}, vol.~7, no.~1, 2007.

\bibitem{vladuvsic2020macroeconomic}
L.~Vladu{\v{s}}i{\'c}, A.~{\v{Z}}ivkovi{\'c}, and N.~Panti{\'c},
  ``Macroeconomic analysis of {GDP} and employment in {EU} countries,''
  \emph{Ekonomika}, vol.~66, no.~1, pp. 65--76, 2020.

\bibitem{bekaert2007global}
G.~Bekaert, C.~R. Harvey, C.~Lundblad, and S.~Siegel, ``Global growth
  opportunities and market integration,'' \emph{The Journal of Finance},
  vol.~62, no.~3, pp. 1081--1137, 2007.

\bibitem{evans2012learning}
G.~W. Evans and S.~Honkapohja, ``Learning and expectations in macroeconomics,''
  in \emph{Learning and Expectations in Macroeconomics}.\hskip 1em plus 0.5em
  minus 0.4em\relax Princeton University Press, 2012.

\bibitem{casey2020macroeconomic}
E.~Casey, ``Do macroeconomic forecasters use macroeconomics to forecast?''
  \emph{International Journal of Forecasting}, vol.~36, no.~4, pp. 1439--1453,
  2020.

\bibitem{giannone2008nowcasting}
D.~Giannone, L.~Reichlin, and D.~Small, ``Nowcasting: The real-time
  informational content of macroeconomic data,'' \emph{Journal of monetary
  economics}, vol.~55, no.~4, pp. 665--676, 2008.

\bibitem{lyhagen2015beating}
J.~Lyhagen, S.~Ekberg, and R.~Eidestedt, ``Beating the var: Improving swedish
  gdp forecasts using error and intercept corrections,'' \emph{Journal of
  Forecasting}, vol.~34, no.~5, pp. 354--363, 2015.

\bibitem{Ewing2005}
\BIBentryALTinterwordspacing
R.~Ewing, D.~Gruen, and J.~Hawkins, ``{Forecasting the macro economy},''
  \emph{Economic Roundup}, no.~2, pp. 11--25, June 2005. [Online]. Available:
  \url{https://ideas.repec.org/a/tsy/journl/journl_tsy_er_2005_2_1.html}
\BIBentrySTDinterwordspacing

\bibitem{imf2022report}
IMF, ``World economic outlook,'' \emph{International Monetary Fund}, 2022.

\bibitem{vreeland2003imf}
J.~R. Vreeland, \emph{The {IMF} and economic development}.\hskip 1em plus 0.5em
  minus 0.4em\relax Cambridge University Press, 2003.

\bibitem{garboden2020sources}
P.~M. Garboden, ``Sources and types of big data for macroeconomic
  forecasting,'' in \emph{Macroeconomic Forecasting in the Era of Big
  Data}.\hskip 1em plus 0.5em minus 0.4em\relax Springer, 2020, pp. 3--23.

\bibitem{nicholson2017varx}
W.~B. Nicholson, D.~S. Matteson, and J.~Bien, ``Varx-l: Structured
  regularization for large vector autoregressions with exogenous variables,''
  \emph{International Journal of Forecasting}, vol.~33, no.~3, pp. 627--651,
  2017.

\bibitem{marcellino2010factor}
M.~Marcellino and C.~Schumacher, ``Factor midas for nowcasting and forecasting
  with ragged-edge data: A model comparison for german gdp,'' \emph{Oxford
  Bulletin of Economics and Statistics}, vol.~72, no.~4, pp. 518--550, 2010.

\bibitem{Feldkircher2015}
M.~Feldkircher, F.~Huber, J.~Schreiner, M.~Tirpák, P.~Tóth, and J.~Wörz,
  ``Bridging the information gap: small-scale nowcasting models of gdp growth
  for selected cesee countries,'' \emph{Focus on European economic
  integration}, pp. 56--75, 06 2015.

\bibitem{Garnitz2019}
\BIBentryALTinterwordspacing
J.~Garnitz, R.~Lehmann, and K.~Wohlrabe, ``Forecasting gdp all over the world
  using leading indicators based on comprehensive survey data,'' \emph{Applied
  Economics}, vol.~51, no.~54, pp. 5802--5816, 2019. [Online]. Available:
  \url{https://doi.org/10.1080/00036846.2019.1624915}
\BIBentrySTDinterwordspacing

\bibitem{Ollivaud2016}
\BIBentryALTinterwordspacing
P.~Ollivaud, P.-A. Pionnier, E.~Rusticelli, C.~Schwellnus, and S.-H. Koh,
  ``Forecasting gdp during and after the great recession,'' no. 1313, 2016.
  [Online]. Available:
  \url{https://www.oecd-ilibrary.org/content/paper/5jlv2jj4mw40-en}
\BIBentrySTDinterwordspacing

\bibitem{montgomery2021introduction}
D.~C. Montgomery, E.~A. Peck, and G.~G. Vining, \emph{Introduction to linear
  regression analysis}.\hskip 1em plus 0.5em minus 0.4em\relax John Wiley \&
  Sons, 2021.

\bibitem{box2015time}
G.~E. Box, G.~M. Jenkins, G.~C. Reinsel, and G.~M. Ljung, \emph{Time series
  analysis: forecasting and control}.\hskip 1em plus 0.5em minus 0.4em\relax
  John Wiley \& Sons, 2015.

\bibitem{sims1980macroeconomics}
C.~A. Sims, ``Macroeconomics and reality,'' \emph{Econometrica: journal of the
  Econometric Society}, pp. 1--48, 1980.

\bibitem{athey2018impact}
\BIBentryALTinterwordspacing
S.~Athey, ``21. the impact of machine learning on economics,'' in \emph{The
  Economics of Artificial Intelligence}, A.~Agrawal, J.~Gans, and A.~Goldfarb,
  Eds.\hskip 1em plus 0.5em minus 0.4em\relax Chicago: University of Chicago
  Press, 2019, pp. 507--552. [Online]. Available:
  \url{https://doi.org/10.7208/9780226613475-023}
\BIBentrySTDinterwordspacing

\bibitem{athey2019machine}
S.~Athey and G.~W. Imbens, ``Machine learning methods that economists should
  know about,'' \emph{Annual Review of Economics}, vol.~11, pp. 685--725, 2019.

\bibitem{yoon2021forecasting}
J.~Yoon, ``Forecasting of real {GDP} growth using machine learning models:
  Gradient boosting and random forest approach,'' \emph{Computational
  Economics}, vol.~57, no.~1, pp. 247--265, 2021.

\bibitem{Deng2018ArtificialII}
L.~Deng, ``Artificial intelligence in the rising wave of deep learning: The
  historical path and future outlook [perspectives],'' \emph{IEEE Signal
  Processing Magazine}, vol.~35, pp. 180--177, 2018.

\bibitem{lecun2015deep}
Y.~LeCun, Y.~Bengio, and G.~Hinton, ``Deep learning,'' \emph{Nature}, vol. 521,
  no. 7553, pp. 436--444, 2015.

\bibitem{Elman1990}
\BIBentryALTinterwordspacing
J.~L. Elman, ``Finding structure in time,'' \emph{Cognitive Science}, vol.~14,
  no.~2, pp. 179--211, 1990. [Online]. Available:
  \url{https://www.sciencedirect.com/science/article/pii/036402139090002E}
\BIBentrySTDinterwordspacing

\bibitem{chandra2021covid}
R.~Chandra and A.~Krishna, ``Covid-19 sentiment analysis via deep learning
  during the rise of novel cases,'' \emph{PLoS One}, vol.~16, no.~8, p.
  e0255615, 2021.

\bibitem{sun2017review}
S.~Sun, C.~Luo, and J.~Chen, ``A review of natural language processing
  techniques for opinion mining systems,'' \emph{Information fusion}, vol.~36,
  pp. 10--25, 2017.

\bibitem{chandra2021biden}
R.~Chandra and R.~Saini, ``Biden vs trump: Modeling us general elections using
  bert language model,'' \emph{IEEE Access}, vol.~9, pp. 128\,494--128\,505,
  2021.

\bibitem{tcbb:Hawkins+Boden:2005}
J.~Hawkins and M.~Bod\'en, ``The applicability of recurrent neural networks for
  biological sequence analysis,'' \emph{IEEE/ACM Transactions on Computational
  Biology and Bioinformatics}, vol.~2, no.~3, pp. 243--253, 2005.

\bibitem{chandra2021evaluation}
R.~Chandra, S.~Goyal, and R.~Gupta, ``Evaluation of deep learning models for
  multi-step ahead time series prediction,'' \emph{IEEE Access}, vol.~9, pp.
  83\,105--83\,123, 2021.

\bibitem{hochreiter1997long}
S.~Hochreiter and J.~Schmidhuber, ``Long short-term memory,'' \emph{Neural
  computation}, vol.~9, no.~8, pp. 1735--1780, 1997.

\bibitem{graves2005bidirectional}
A.~Graves, S.~Fern{\'a}ndez, and J.~Schmidhuber, ``Bidirectional lstm networks
  for improved phoneme classification and recognition,'' in \emph{Artificial
  Neural Networks: Formal Models and Their Applications -- ICANN 2005},
  W.~Duch, J.~Kacprzyk, E.~Oja, and S.~Zadro{\.{z}}ny, Eds.\hskip 1em plus
  0.5em minus 0.4em\relax Berlin, Heidelberg: Springer Berlin Heidelberg, 2005,
  pp. 799--804.

\bibitem{said2021predicting}
A.~B. Said, A.~Erradi, H.~A. Aly, and A.~Mohamed, ``Predicting covid-19 cases
  using bidirectional lstm on multivariate time series,'' \emph{Environmental
  Science and Pollution Research}, vol.~28, no.~40, pp. 56\,043--56\,052, 2021.

\bibitem{cho2014learning}
K.~Cho, B.~Van~Merri{\"e}nboer, C.~Gulcehre, D.~Bahdanau, F.~Bougares,
  H.~Schwenk, and Y.~Bengio, ``Learning phrase representations using rnn
  encoder-decoder for statistical machine translation,'' \emph{arXiv preprint
  arXiv:1406.1078}, 2014.

\bibitem{park2018sequence}
S.~H. Park, B.~Kim, C.~M. Kang, C.~C. Chung, and J.~W. Choi,
  ``Sequence-to-sequence prediction of vehicle trajectory via lstm
  encoder-decoder architecture,'' in \emph{2018 IEEE Intelligent Vehicles
  Symposium (IV)}.\hskip 1em plus 0.5em minus 0.4em\relax IEEE, 2018, pp.
  1672--1678.

\bibitem{alzubaidi2021review}
L.~Alzubaidi, J.~Zhang, A.~J. Humaidi, A.~Al-Dujaili, Y.~Duan, O.~Al-Shamma,
  J.~Santamar{\'\i}a, M.~A. Fadhel, M.~Al-Amidie, and L.~Farhan, ``Review of
  deep learning: Concepts, cnn architectures, challenges, applications, future
  directions,'' \emph{Journal of big Data}, vol.~8, no.~1, pp. 1--74, 2021.

\bibitem{chen2016financial}
J.-F. Chen, W.-L. Chen, C.-P. Huang, S.-H. Huang, and A.-P. Chen, ``Financial
  time-series data analysis using deep convolutional neural networks,'' in
  \emph{2016 7th International conference on cloud computing and big data
  (CCBD)}.\hskip 1em plus 0.5em minus 0.4em\relax IEEE, 2016, pp. 87--92.

\bibitem{maehashi2020macroeconomic}
K.~Maehashi and M.~Shintani, ``Macroeconomic forecasting using factor models
  and machine learning: an application to japan,'' \emph{Journal of the
  Japanese and International Economies}, vol.~58, p. 101104, 2020.

\bibitem{makridakis2018statistical}
S.~Makridakis, E.~Spiliotis, and V.~Assimakopoulos, ``Statistical and machine
  learning forecasting methods: Concerns and ways forward,'' \emph{PloS one},
  vol.~13, no.~3, p. e0194889, 2018.

\bibitem{gilliland2020value}
M.~Gilliland, ``The value added by machine learning approaches in
  forecasting,'' \emph{International Journal of Forecasting}, vol.~36, no.~1,
  pp. 161--166, 2020.

\bibitem{bianchi2022belief}
F.~Bianchi, S.~C. Ludvigson, and S.~Ma, ``Belief distortions and macroeconomic
  fluctuations,'' \emph{American Economic Review}, vol. 112, no.~7, pp.
  2269--2315, 2022.

\bibitem{goulet2022machine}
P.~Goulet~Coulombe, M.~Leroux, D.~Stevanovic, and S.~Surprenant, ``How is
  machine learning useful for macroeconomic forecasting?'' \emph{Journal of
  Applied Econometrics}, vol.~37, no.~5, pp. 920--964, 2022.

\bibitem{sa2020prediction}
S.~Sa'adah and M.~S. Wibowo, ``Prediction of gross domestic product (gdp) in
  indonesia using deep learning algorithm,'' in \emph{2020 3rd International
  Seminar on Research of Information Technology and Intelligent Systems
  (ISRITI)}.\hskip 1em plus 0.5em minus 0.4em\relax IEEE, 2020, pp. 32--36.

\bibitem{zhang2022china}
J.~Zhang, J.~Wen, and Z.~Yang, ``China’s gdp forecasting using long short
  term memory recurrent neural network and hidden markov model,'' \emph{PloS
  one}, vol.~17, no.~6, p. e0269529, 2022.

\bibitem{sokolov2016economic}
S.~Sokolov-Mladenovi{\'c}, M.~Milovan{\v{c}}evi{\'c}, I.~Mladenovi{\'c}, and
  M.~Alizamir, ``Economic growth forecasting by artificial neural network with
  extreme learning machine based on trade, import and export parameters,''
  \emph{Computers in Human Behavior}, vol.~65, pp. 43--45, 2016.

\bibitem{bryan1993consumer}
M.~F. Bryan and S.~G. Cecchetti, ``The consumer price index as a measure of
  inflation,'' 1993.

\bibitem{mourougane2006forecasting}
A.~Mourougane, ``Forecasting monthly gdp for canada,'' 2006.

\bibitem{petrova2015using}
S.~Petrova and D.~Marinova, ``Using ‘soft’and ‘hard’social impact
  indicators to understand societal change caused by mining: a western
  australia case study,'' \emph{Impact Assessment and Project Appraisal},
  vol.~33, no.~1, pp. 16--27, 2015.

\bibitem{longo2022neural}
L.~Longo, M.~Riccaboni, and A.~Rungi, ``A neural network ensemble approach for
  gdp forecasting,'' \emph{Journal of Economic Dynamics and Control}, vol. 134,
  p. 104278, 2022.

\bibitem{kidane2013china}
W.~Kidane and W.~Zhu, ``{China-Africa} investment treaties: Old rules, new
  challenges,'' \emph{Fordham Int'l LJ}, vol.~37, p. 1035, 2013.

\bibitem{paloni2012imf}
A.~Paloni and M.~Zanardi, \emph{The {IMF, World Bank} and policy reform}.\hskip
  1em plus 0.5em minus 0.4em\relax Routledge, 2012.

\bibitem{clements1996multi}
M.~P. Clements and D.~F. Hendry, ``Multi-step estimation for forecasting,''
  \emph{Oxford Bulletin of Economics and Statistics}, vol.~58, no.~4, pp.
  657--684, 1996.

\bibitem{klein1971essay}
L.~R. Klein, \emph{essay on the theory of economic prediction}, 1971.

\bibitem{marcellino2006comparison}
M.~Marcellino, J.~H. Stock, and M.~W. Watson, ``A comparison of direct and
  iterated multistep ar methods for forecasting macroeconomic time series,''
  \emph{Journal of econometrics}, vol. 135, no. 1-2, pp. 499--526, 2006.

\bibitem{abeysinghe1998forecasting}
T.~Abeysinghe, ``Forecasting singapore's quarterly gdp with monthly external
  trade,'' \emph{International Journal of Forecasting}, vol.~14, no.~4, pp.
  505--513, 1998.

\bibitem{meyler1998forecasting}
A.~Meyler, G.~Kenny, and T.~Quinn, ``Forecasting irish inflation using arima
  models,'' 1998.

\bibitem{abonazel2019forecasting}
M.~R. Abonazel and A.~I. Abd-Elftah, ``Forecasting egyptian gdp using arima
  models,'' \emph{Reports on Economics and Finance}, vol.~5, no.~1, pp. 35--47,
  2019.

\bibitem{freeman1989vector}
J.~R. Freeman, J.~T. Williams, and T.-m. Lin, ``Vector autoregression and the
  study of politics,'' \emph{American Journal of Political Science}, pp.
  842--877, 1989.

\bibitem{hox1998introduction}
J.~J. Hox and T.~M. Bechger, ``An introduction to structural equation
  modeling,'' 1998.

\bibitem{robertson1999vector}
J.~C. Robertson and E.~W. Tallman, ``Vector autoregressions: forecasting and
  reality,'' \emph{Economic Review-Federal Reserve Bank of Atlanta}, vol.~84,
  no.~1, p.~4, 1999.

\bibitem{salisu2021effect}
A.~A. Salisu, R.~Gupta, and A.~Olaniran, ``The effect of oil uncertainty shock
  on real gdp of 33 countries: a global var approach,'' \emph{Applied Economics
  Letters}, pp. 1--6, 2021.

\bibitem{iorio2022comparison}
F.~D. Iorio and U.~Triacca, ``A comparison between var processes jointly
  modeling gdp and unemployment rate in france and germany,'' \emph{Statistical
  Methods \& Applications}, vol.~31, no.~3, pp. 617--635, 2022.

\bibitem{hryhorkiv2020forecasting}
V.~Hryhorkiv, L.~Buiak, A.~Verstiak, M.~Hryhorkiv, O.~Verstiak, and
  K.~Tokarieva, ``Forecasting financial time sesries using combined arima-ann
  algorithm,'' in \emph{2020 10th International Conference on Advanced Computer
  Information Technologies (ACIT)}.\hskip 1em plus 0.5em minus 0.4em\relax
  IEEE, 2020, pp. 455--458.

\bibitem{d2018macroeconomic}
H.~d’Albis, E.~Boubtane, and D.~Coulibaly, ``Macroeconomic evidence suggests
  that asylum seekers are not a “burden” for western european countries,''
  \emph{Science advances}, vol.~4, no.~6, p. eaaq0883, 2018.

\bibitem{muller1997predicting}
K.-R. M{\"u}ller, A.~J. Smola, G.~R{\"a}tsch, B.~Sch{\"o}lkopf, J.~Kohlmorgen,
  and V.~Vapnik, ``Predicting time series with support vector machines,'' in
  \emph{International conference on artificial neural networks}.\hskip 1em plus
  0.5em minus 0.4em\relax Springer, 1997, pp. 999--1004.

\bibitem{lau2008local}
K.~Lau and Q.~Wu, ``Local prediction of non-linear time series using support
  vector regression,'' \emph{Pattern recognition}, vol.~41, no.~5, pp.
  1539--1547, 2008.

\bibitem{lu2009financial}
C.-J. Lu, T.-S. Lee, and C.-C. Chiu, ``Financial time series forecasting using
  independent component analysis and support vector regression,''
  \emph{Decision support systems}, vol.~47, no.~2, pp. 115--125, 2009.

\bibitem{sapankevych2009time}
N.~I. Sapankevych and R.~Sankar, ``Time series prediction using support vector
  machines: a survey,'' \emph{IEEE computational intelligence magazine},
  vol.~4, no.~2, pp. 24--38, 2009.

\bibitem{amiri2009svm}
S.~Amiri, D.~Von~Rosen, and S.~Zwanzig, ``The svm approach for box--jenkins
  models,'' \emph{REVSTAT-Statistical journal}, vol.~7, no.~1, pp. 23--36,
  2009.

\bibitem{tseng2002combining}
F.-M. Tseng, H.-C. Yu, and G.-H. Tzeng, ``Combining neural network model with
  seasonal time series arima model,'' \emph{Technological forecasting and
  social change}, vol.~69, no.~1, pp. 71--87, 2002.

\bibitem{li2012brief}
J.~Li, J.-h. Cheng, J.-y. Shi, and F.~Huang, ``Brief introduction of back
  propagation (bp) neural network algorithm and its improvement,'' in
  \emph{Advances in computer science and information engineering}.\hskip 1em
  plus 0.5em minus 0.4em\relax Springer, 2012, pp. 553--558.

\bibitem{choi2018stock}
H.~K. Choi, ``Stock price correlation coefficient prediction with arima-lstm
  hybrid model,'' \emph{arXiv preprint arXiv:1808.01560}, 2018.

\bibitem{ouhame2019multivariate}
S.~Ouhame and Y.~Hadi, ``Multivariate workload prediction using vector
  autoregressive and stacked lstm models,'' in \emph{Proceedings of the New
  Challenges in Data Sciences: Acts of the Second Conference of the Moroccan
  Classification Society}, 2019, pp. 1--7.

\bibitem{goodfellow2016deep}
I.~Goodfellow, Y.~Bengio, and A.~Courville, \emph{Deep learning}.\hskip 1em
  plus 0.5em minus 0.4em\relax MIT press, 2016.

\bibitem{patterson2017deep}
J.~Patterson and A.~Gibson, \emph{Deep learning: A practitioner's
  approach}.\hskip 1em plus 0.5em minus 0.4em\relax " O'Reilly Media, Inc.",
  2017.

\bibitem{jelodar2020deep}
H.~Jelodar, Y.~Wang, R.~Orji, and S.~Huang, ``Deep sentiment classification and
  topic discovery on novel coronavirus or covid-19 online discussions: Nlp
  using lstm recurrent neural network approach,'' \emph{IEEE Journal of
  Biomedical and Health Informatics}, vol.~24, no.~10, pp. 2733--2742, 2020.

\bibitem{cadieu2014deep}
C.~F. Cadieu, H.~Hong, D.~L. Yamins, N.~Pinto, D.~Ardila, E.~A. Solomon, N.~J.
  Majaj, and J.~J. DiCarlo, ``Deep neural networks rival the representation of
  primate it cortex for core visual object recognition,'' \emph{PLoS
  computational biology}, vol.~10, no.~12, p. e1003963, 2014.

\bibitem{eraslan2019deep}
G.~Eraslan, {\v{Z}}.~Avsec, J.~Gagneur, and F.~J. Theis, ``Deep learning: new
  computational modelling techniques for genomics,'' \emph{Nature Reviews
  Genetics}, vol.~20, no.~7, pp. 389--403, 2019.

\bibitem{karevan2020transductive}
Z.~Karevan and J.~A. Suykens, ``Transductive lstm for time-series prediction:
  An application to weather forecasting,'' \emph{Neural Networks}, vol. 125,
  pp. 1--9, 2020.

\bibitem{smalter2017macroeconomic}
A.~Smalter~Hall and T.~R. Cook, ``Macroeconomic indicator forecasting with deep
  neural networks,'' \emph{Federal Reserve Bank of Kansas City Working Paper},
  no. 17-11, 2017.

\bibitem{siami2019performance}
S.~Siami-Namini, N.~Tavakoli, and A.~S. Namin, ``The performance of lstm and
  bilstm in forecasting time series,'' in \emph{2019 IEEE International
  Conference on Big Data (Big Data)}.\hskip 1em plus 0.5em minus 0.4em\relax
  IEEE, 2019, pp. 3285--3292.

\bibitem{selvin2017stock}
S.~Selvin, R.~Vinayakumar, E.~Gopalakrishnan, V.~K. Menon, and K.~Soman,
  ``Stock price prediction using lstm, rnn and cnn-sliding window model,'' in
  \emph{2017 international conference on advances in computing, communications
  and informatics (icacci)}.\hskip 1em plus 0.5em minus 0.4em\relax IEEE, 2017,
  pp. 1643--1647.

\bibitem{piao2019housing}
Y.~Piao, A.~Chen, and Z.~Shang, ``Housing price prediction based on cnn,'' in
  \emph{2019 9th international conference on information science and technology
  (ICIST)}.\hskip 1em plus 0.5em minus 0.4em\relax IEEE, 2019, pp. 491--495.

\bibitem{livieris2020cnn}
I.~E. Livieris, E.~Pintelas, and P.~Pintelas, ``A cnn--lstm model for gold
  price time-series forecasting,'' \emph{Neural computing and applications},
  vol.~32, no.~23, pp. 17\,351--17\,360, 2020.

\bibitem{vu2020sources}
K.~M. Vu, ``Sources of growth in the world economy: a comparison of {G7 and E7}
  economies,'' in \emph{Measuring economic growth and productivity}.\hskip 1em
  plus 0.5em minus 0.4em\relax Elsevier, 2020, pp. 55--74.

\bibitem{jorgenson2013emergence}
D.~W. Jorgenson and K.~M. Vu, ``The emergence of the new economic order: Growth
  in the {G7 and the G20},'' \emph{Journal of Policy Modeling}, vol.~35, no.~3,
  pp. 389--399, 2013.

\bibitem{o2001building}
J.~O'neill \emph{et~al.}, ``Building better global economic brics,'' 2001.

\bibitem{feenstra2015next}
R.~C. Feenstra, R.~Inklaar, and M.~P. Timmer, ``The next generation of the penn
  world table,'' \emph{American economic review}, vol. 105, no.~10, pp.
  3150--82, 2015.

\bibitem{bozdogan1987model}
H.~Bozdogan, ``Model selection and akaike's information criterion (aic): The
  general theory and its analytical extensions,'' \emph{Psychometrika},
  vol.~52, no.~3, pp. 345--370, 1987.

\bibitem{sutskever2014sequence}
I.~Sutskever, O.~Vinyals, and Q.~V. Le, ``Sequence to sequence learning with
  neural networks,'' \emph{Advances in neural information processing systems},
  vol.~27, 2014.

\bibitem{lecun1995convolutional}
Y.~LeCun, Y.~Bengio \emph{et~al.}, ``Convolutional networks for images, speech,
  and time series,'' \emph{The handbook of brain theory and neural networks},
  vol. 3361, no.~10, p. 1995, 1995.

\bibitem{majib2021vgg}
M.~S. Majib, M.~M. Rahman, T.~S. Sazzad, N.~I. Khan, and S.~K. Dey,
  ``Vgg-scnet: A vgg net-based deep learning framework for brain tumor
  detection on mri images,'' \emph{IEEE Access}, vol.~9, pp.
  116\,942--116\,952, 2021.

\bibitem{chen2022alexnet}
H.-C. Chen, A.~M. Widodo, A.~Wisnujati, M.~Rahaman, J.~C.-W. Lin, L.~Chen, and
  C.-E. Weng, ``Alexnet convolutional neural network for disease detection and
  classification of tomato leaf,'' \emph{Electronics}, vol.~11, no.~6, p. 951,
  2022.

\bibitem{kingma2014adam}
D.~P. Kingma and J.~Ba, ``Adam: A method for stochastic optimization,''
  \emph{arXiv preprint arXiv:1412.6980}, 2014.

\bibitem{edey2009global}
M.~Edey, ``The global financial crisis and its effects,'' \emph{Economic
  Papers: A journal of applied economics and policy}, vol.~28, no.~3, pp.
  186--195, 2009.

\bibitem{izzeldin2021impact}
M.~Izzeldin, Y.~G. Murado{\u{g}}lu, V.~Pappas, and S.~Sivaprasad, ``The impact
  of {COVID-19 on G7} stock markets volatility: Evidence from a {ST-HAR}
  model,'' \emph{International Review of Financial Analysis}, vol.~74, p.
  101671, 2021.

\bibitem{hudson2003international}
J.~Hudson and P.~Jones, ``International trade in ‘quality goods’:
  signalling problems for developing countries,'' \emph{Journal of
  international Development}, vol.~15, no.~8, pp. 999--1013, 2003.

\bibitem{bussmann2006trade}
M.~Bussmann, H.~Scheuthle, and G.~Schneider, ``Trade liberalization and
  political instability in developing countries,'' in \emph{Programming for
  Peace}.\hskip 1em plus 0.5em minus 0.4em\relax Springer, 2006, pp. 49--70.

\bibitem{feng2001political}
Y.~Feng, ``Political freedom, political instability, and policy uncertainty: A
  study of political institutions and private investment in developing
  countries,'' \emph{International Studies Quarterly}, vol.~45, no.~2, pp.
  271--294, 2001.

\bibitem{Lowy2030forecast}
``{World Economic Outlook: Countering the Cost-of-Living Crisis. Washington,
  DC. October, International Monetary Fund},''
  \url{https://www.imf.org/en/Publications/WEO/weo-database/2022/October},
  2022.

\bibitem{CEBR2035India}
C.~for Economics and B.~Research, ``India to become 10-trillion economy by
  2035: Cebr,''
  \url{https://economictimes.indiatimes.com/news/economy/policy/india-to-become-10-trillion-economy-by-2035-cebr/articleshow/96526283.cms}.

\bibitem{MS2027India}
M.~Stanley, ``India to be 3rd largest economy by 2027,''
  \url{https://www.fortuneindia.com/macro/india-to-be-3rd-largest-economy-by-2027-morgan-stanley/110332#:~:text=As$\%$20India's$\%$20economy$\%$20grows$\%$20manifold,by$\%$202032$\%$2C$\%$20the$\%$20economist$\%$20said}.

\bibitem{ozili2023spillover}
P.~K. Ozili and T.~Arun, ``Spillover of covid-19: impact on the global
  economy,'' in \emph{Managing Inflation and Supply Chain Disruptions in the
  Global Economy}.\hskip 1em plus 0.5em minus 0.4em\relax IGI Global, 2023, pp.
  41--61.

\bibitem{modis2013long}
T.~Modis, ``Long-term gdp forecasts and the prospects for growth,''
  \emph{Technological Forecasting and Social Change}, vol.~80, no.~8, pp.
  1557--1562, 2013.

\bibitem{jiang2022china}
S.~Jiang, J.~Miao, and Y.~Zhang, ``China's housing bubble, infrastructure
  investment, and economic growth,'' \emph{International Economic Review},
  vol.~63, no.~3, pp. 1189--1237, 2022.

\bibitem{fajgelbaum2022economic}
P.~D. Fajgelbaum and A.~K. Khandelwal, ``The economic impacts of the us--china
  trade war,'' \emph{Annual Review of Economics}, vol.~14, pp. 205--228, 2022.

\bibitem{habibi2022potential}
Z.~Habibi, H.~Habibi, and M.~A. Mohammadi, ``The potential impact of covid-19
  on the chinese gdp, trade, and economy,'' \emph{Economies}, vol.~10, no.~4,
  p.~73, 2022.

\bibitem{ding2022digital}
C.~Ding, C.~Liu, C.~Zheng, and F.~Li, ``Digital economy, technological
  innovation and high-quality economic development: Based on spatial effect and
  mediation effect,'' \emph{Sustainability}, vol.~14, no.~1, p. 216, 2022.

\bibitem{neal1992bayesian}
R.~Neal, ``Bayesian learning via stochastic dynamics,'' \emph{Advances in
  neural information processing systems}, vol.~5, 1992.

\bibitem{chandra2020bayesian}
R.~Chandra and A.~Kapoor, ``Bayesian neural multi-source transfer learning,''
  \emph{Neurocomputing}, vol. 378, pp. 54--64, 2020.

\bibitem{lewis2013theory}
W.~A. Lewis, \emph{Theory of economic growth}.\hskip 1em plus 0.5em minus
  0.4em\relax Routledge, 2013.

\bibitem{ccalicskan2015technological}
H.~K. {\c{C}}al{\i}{\c{s}}kan, ``Technological change and economic growth,''
  \emph{Procedia-Social and Behavioral Sciences}, vol. 195, pp. 649--654, 2015.

\end{thebibliography}

 \begin{IEEEbiography}
 [{\includegraphics[width=1in,height=1.25in,clip,keepaspectratio]{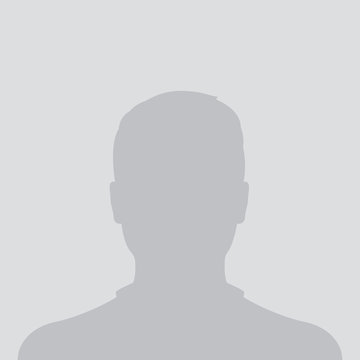}}]{Tianyi Wang} 
 graduated from Masters in Statistics from UNSW Sydney and has strong interests in machine learning, financial forecasting and deep learning.\\  
 \end{IEEEbiography}

\begin{IEEEbiography}
[{\includegraphics[width=1in,height=1.25in,clip,keepaspectratio]{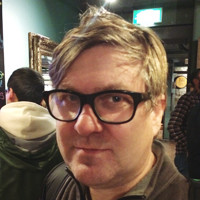}}]{Rodney Beard} 
 graduated with a PhD in 2005 from the University of Queensland. He has a background in economics and is currently a researcher at the Pingala Institute. He has taught a variety of subjects at higher education institutions in Australia, Canada, France, UK, China and Fiji. His research focuses on computational methods in economics and finance,   mathematical modelling, and machine learning.\\
\end{IEEEbiography}

\begin{IEEEbiography}
[{\includegraphics[width=1in,height=1.25in,clip,keepaspectratio]{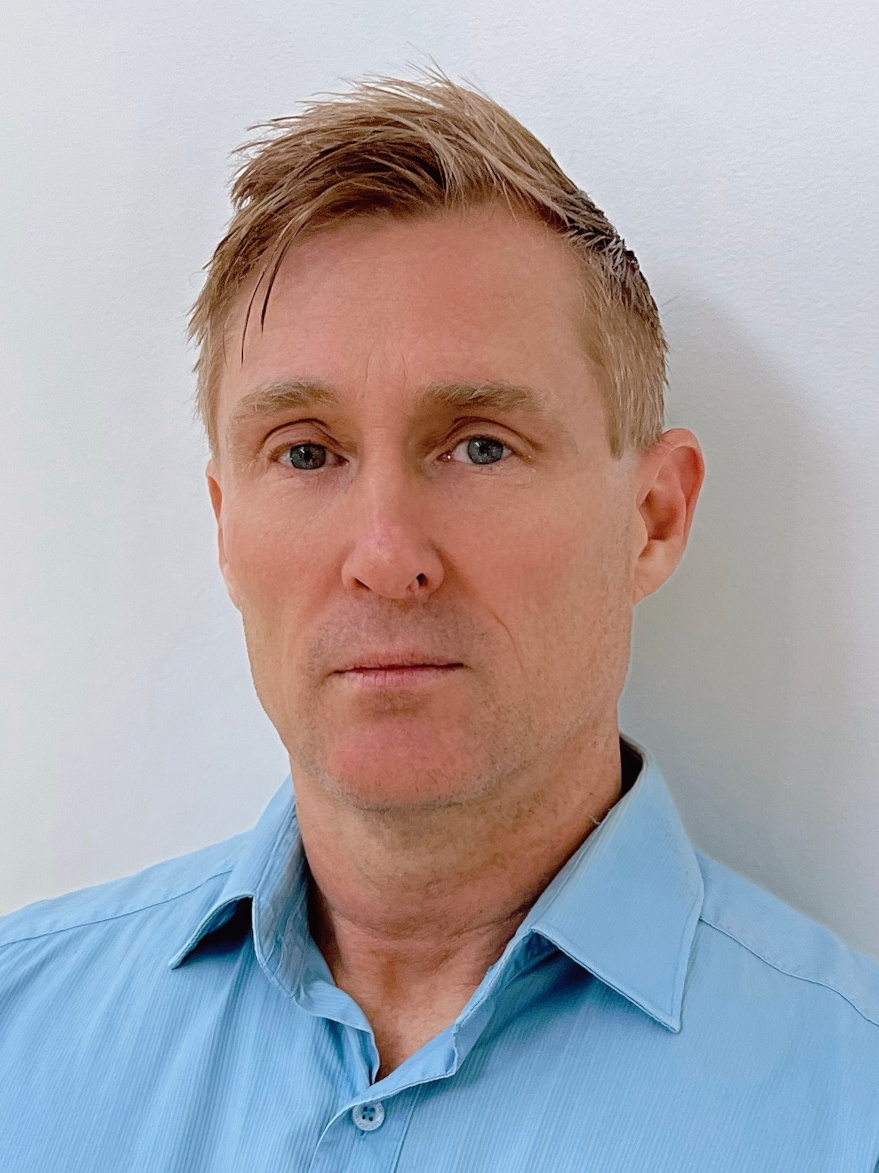}}]{John Hawkins} 

 is the Chief Scientist at Playground XYZ and a researcher at the Pingala Institute. He received his PhD from the University of Queensland for work on bioinformatics applications of neural networks. He has worked for 20 years researching and building machine learning systems in academia and industry. He develops open-source Python packages for data science productivity. He is the author of the book 'Getting Data Science Done' and his research projects span natural language processing, machine reasoning, time series forecasting, and explainability of model predictions. \\
\end{IEEEbiography}

\begin{IEEEbiography}
[{\includegraphics[width=1in,height=1.25in,clip,keepaspectratio]{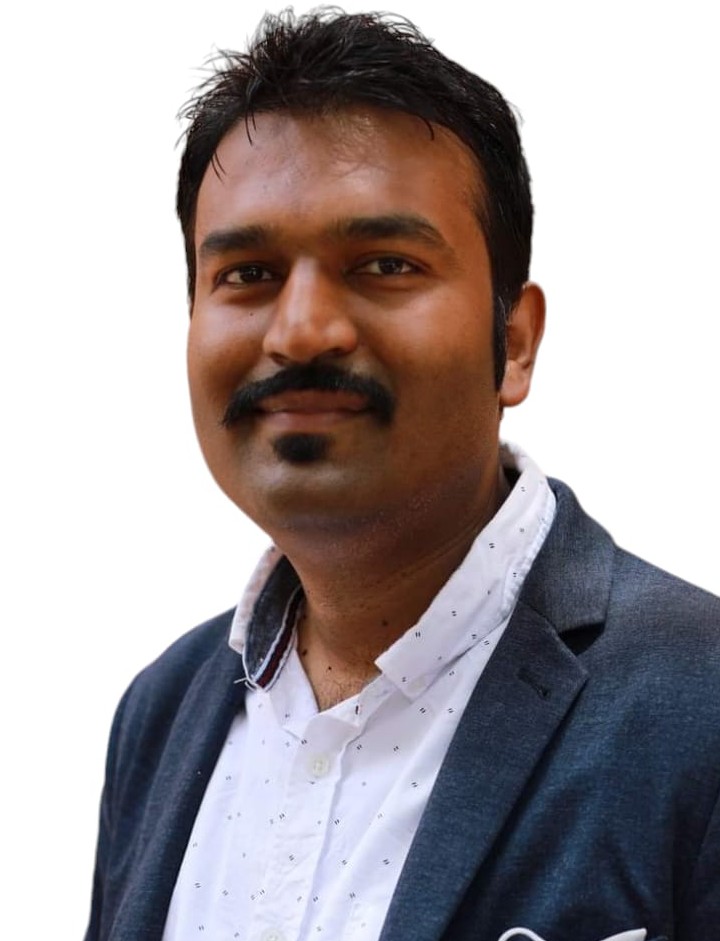}}]
{Rohitash Chandra}  

  is a Senior Lecturer in Data Science at the UNSW School of Mathematics and Statistics and IEEE Senior Member.  He leads a program of research encircling methodologies and applications of artificial intelligence.   Dr Chandra has used  language models for analysis of translations of religious-philosophical texts.   Dr Chandra used machine learning and remote sensing for mineral exploration and is currently focusing on critical metals and climate extremes, such as high category cyclones, stream flow and flooding events. Dr Chandra has been listed in Stanford's list of top 2 \% scientists since 2020. 
\end{IEEEbiography}

\EOD

\end{document}